\documentclass{bmvc2k}

\usepackage{amsmath,amsfonts,amssymb}
\usepackage{algorithmic}
\usepackage{algorithm}
\usepackage{array}
\usepackage{textcomp}
\usepackage{stfloats}
\usepackage{url}
\usepackage{verbatim}
\usepackage{graphicx}

\usepackage[utf8]{inputenc}
\usepackage{cmap}
\usepackage[T1]{fontenc}
\usepackage{hyperref}
\hypersetup{
    colorlinks = true,
    allcolors = [RGB]{255,90,95}
}
\usepackage{url}
\usepackage{booktabs}
\usepackage{nicefrac}
\usepackage{microtype}
\usepackage{fancyhdr}
\usepackage{pgf}
\usepackage{soul}
\usepackage{xcolor}
\usepackage{soul}
\usepackage{rotating}
\usepackage{amssymb}
\usepackage{tcolorbox}
\usepackage{wrapfig}

\definecolor{color0}{HTML}{002147}
\definecolor{color1}{HTML}{DCEDFF}

\newcommand{\newhypbox}[2]{
\begin{tcolorbox}[
    colback=black!5!white,
    colframe=white,
    boxsep=1.1pt,
    width=0.99\textwidth,
    ]
    \vspace{1pt}
    {\textbf{Desideratum #1: }#2}
\end{tcolorbox}
}

\usepackage{tikz}

\begin{document}

\title{Is Single-View Mesh Reconstruction Ready for Robotics?}

\addauthor{Frederik Nolte}{frederik@robots.ox.ac.uk}{1}
\addauthor{Andreas Geiger}{a.geiger@uni-tuebingen.de}{2}
\addauthor{Bernhard Sch\"olkopf}{bs@tue.mpg.de}{3}
\addauthor{Ingmar Posner}{ingmar@robots.ox.ac.uk}{1}

\addinstitution{
 Oxford Robotics Institute,\\
 University of Oxford
}
\addinstitution{
 University of Tübingen,\\
 Tübingen AI Center
}
\addinstitution{
 Max Planck Institute for Intelligent Systems,\\
 ELLIS Institute Tübingen
}

\runninghead{Nolte et al.}{Is Single-View Mesh Reconstruction Ready for Robotics?}

\maketitle

\begin{abstract}
This paper evaluates single-view mesh reconstruction models for their potential in enabling instant digital twin creation for real-time planning and dynamics prediction using physics simulators for robotic manipulation. Recent single-view 3D reconstruction advances offer a promising avenue toward an automated real-to-sim pipeline: directly mapping a single observation of a scene into a simulation instance by reconstructing scene objects as individual, complete, and physically plausible 3D meshes. However, their suitability for physics simulations and robotics applications under immediacy, physical fidelity, and simulation readiness remains underexplored. We establish robotics-specific benchmarking criteria for 3D reconstruction, including handling typical inputs, collision-free and stable geometry, occlusion robustness, and meeting computational constraints. Our empirical evaluation using realistic robotics datasets shows that despite success on computer vision benchmarks, existing approaches fail to meet robotics-specific requirements.
We quantitively examine limitations of single-view reconstruction for practical robotics implementation, in contrast to prior work that focuses on multi-view approaches.
Our findings highlight critical gaps between computer vision advances and robotics needs, guiding future research at this intersection.
\end{abstract}

\section{Introduction}
\label{sec:introduction}
\begin{wrapfigure}{r}{0.49\textwidth}
    \centering
    \vspace{-1pt}
    \includegraphics[width=\linewidth]{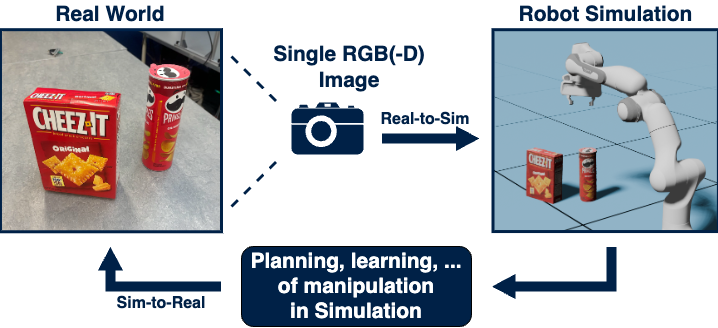}
    \vspace{-15pt}
    \caption{Illustration of a general Real2Sim2Real pipeline for robotic manipulation. A single RGB(-D) view is used for scene reconstruction in a physics simulator. The simulation instance can then be used for planning, model-predictive control, or real-time learning, and the resulting motion is executed in the real world.}
    \label{fig:pipeline}
    \vspace{-10pt}
\end{wrapfigure}

Simulating future world states from perception is central to robot planning and learning \cite{ha_world_2018,kaiser_model_2019}. While learnt world models have shown promise for model-based control and planning \cite{hafner_learning_2019,hafner_dream_2019} they suffer from poor generalisation, opaque failure modes, and limited physical interpretability \cite{wang_benchmarking_2019,moerland_model-based_2023}. An alternative is to bypass learnt dynamics by constructing a \emph{digital twin} in simulation, a geometric reconstruction of the currently observed scene suitable for robotic simulation and interaction, enabling robots to reason about the world using physics-based simulators. Fig. \ref{fig:pipeline} illustrates such an approach.

Creating such digital twins in simulation has traditionally been labour-intensive, requiring specialist training and manually designed assets. This has hindered the integration of simulation into online robotics workflows such as model-predictive control \cite{qin_survey_2003, hansen_temporal_2022}, planning \cite{alterovitz_robot_2016, suomalainen_survey_2022}, online safety evaluation \cite{berkenkamp_safe_2017, o_kelly_scalable_2018}, and real-time learning \cite{smith_demonstrating_2023, bohlinger_gait_2025}. A general, automated reconstruction pipeline for robotic manipulation would directly map RGB(-D) observations of a scene into a simulation instance by reconstructing objects as individual, physically plausible, and complete 3D meshes. If successful, the prospects of scene-agnostic and robust dynamics prediction through simulation are enticing, as these capabilities could address the fundamental data efficiency, generalisation, and robustness limitations of learnt world models \cite{wang_benchmarking_2019, moerland_model-based_2023}.

Recent real-to-sim works leverage multi-view 3D reconstruction to create simulation-compatible environments \cite{barcellona_dream_2024,han_re3sim_2025,li_robogsim_2024,lou_robo-gs_2024,wu_rl-gsbridge_2024,qureshi_splatsim_2024,zhu_vr-robo_2025,jia_discoverse_2024,torne_reconciling_2024,patel_real--sim--real_2024,pfaff_scalable_2025}. While promising, these methods inherently assume complete scene coverage from multiple viewpoints. This progress has been paralleled by emerging computer vision works that produce remarkable improvements in category-agnostic, \textit{single-view}, full-shape 3D mesh reconstruction $-$ models that infer the complete object geometry, including unobserved surfaces, from single-view input \cite{liu_one-2-3-45_2023,xu_instantmesh_2024,zhao_michelangelo_2023}. This raises a compelling question: can the multi-view requirement be relaxed by leveraging single-view reconstruction models out-of-the-box to create accurate digital twin environments for robotic manipulation?

The appeal of single-view approaches extends beyond convenience. In robotics, collecting multiple viewpoints requires physical movement that delays decision-making and introduces additional failure modes through viewpoint planning and collision avoidance under space constraints \cite{meng_two-stage_2017, breyer_closed-loop_2022}. In scenarios such as books on a shelf or objects in cluttered drawers, gaining additional viewpoints is impossible without first interacting with and rearranging the scene \cite{bohg_interactive_2017, pollayil_planning_2021, miao_safe_2023}. We outline four representative scenarios where single-view reconstruction is operationally necessary, and where the reconstructions need not be a perfect replica but must be physically plausible enough to support informative simulation.

\noindent\textbf{Real-time reactive planning.} A robot opening a cluttered drawer or entering a new room must plan immediately $-$ a physically plausible single-view reconstruction enables quick collision checking and coarse forward simulation without geometric perfection.

\noindent\textbf{Viewpoint-inaccessible scenes.} Objects on a shelf, in a narrow cabinet, or at the back of a fridge cannot be observed from additional viewpoints without first manipulating the scene. The robot must reason about geometry before it can interact, making single-view reconstruction a necessity rather than a convenience.

\noindent\textbf{Simulation-in-the-loop safety checking.} A robot forward-simulating in a reconstructed digital twin to check collisions or instability, requiring directional correctness rather than perfect fidelity.

\noindent\textbf{Bootstrapping with refinement.} Single-view reconstruction can initialise active perception: a robot obtains a rough model from one view, identifies high-uncertainty regions, acquires one or two additional views, and refines $-$ more efficiently than exhaustive multi-view scanning.

The existing literature reveals a clear gap. Works leveraging dense viewpoints or scene scans \cite{barcellona_dream_2024,han_re3sim_2025,li_robogsim_2024,lou_robo-gs_2024,wu_rl-gsbridge_2024,qureshi_splatsim_2024,zhu_vr-robo_2025,jia_discoverse_2024,torne_reconciling_2024,patel_real--sim--real_2024,pfaff_scalable_2025} demonstrate promising results but require assumptions about scene coverage that limit practical applicability. Conversely, works that attempt single-view approaches \cite{mu_robotwin_2024,agarwal_scenecomplete_2024,yao_cast_2025,katara_gen2sim_2024} only produce domain-randomised environments, do not perform physics simulation, or lack comprehensive robot experiments. We describe these works in greater detail in Section \ref{sec:taxonomy}.

We are thus led to our central research question: \emph{can recent advancements in single-view mesh reconstruction models be effectively applied to create accurate digital twin environments for online, real-time, and high-throughput robotic manipulation?} To address this systematically, we formulate desiderata particular to robotics and simulation for manipulation, concerning input data, computational resources, and overall scene coherence. We empirically evaluate these desiderata on two representative robotics datasets: the YCB-Video dataset \cite{xiang_posecnn_2018} featuring standardised manipulation objects, and the Aria Digital Twin dataset \cite{pan_aria_2023} containing naturalistic household scenes. Our evaluation focuses on tabletop manipulation scenarios and assesses thirteen state-of-the-art single-view 3D reconstruction models across five key metrics: reconstruction accuracy, collision-free reconstruction, physical stability, occlusion handling, and computational efficiency.

Many robotics workflows require reconstructions that are not perfect replicas but are physically plausible enough to support informative simulation — enabling coarse grasp feasibility assessment, collision checking, or stability prediction. In these settings, the relevant question is not whether the reconstruction is a perfect digital twin, but whether it is sufficiently accurate that downstream simulation-based decisions (e.g., "is this grasp likely to succeed?", "will this object topple if I push that one?") remain reliable. Our desiderata are formulated with this operational notion of sufficiency in mind. This practically important middle ground refutes the apparent dichotomy that single-view reconstruction is either inherently unsuitable (because occluded geometry cannot be recovered with certainty) or unnecessary to evaluate rigorously (because policy training for generalisation does not demand reconstruction fidelity). Neither extreme captures the requirements of the workflows described above.

Our aim is to help robotics practitioners and computer vision researchers select suitable tools by highlighting existing capability gaps. Inspired by robotics requirements, we focus on reconstruction models that (1) work with single-view inputs, since it is impractical to assume multiple meaningful viewpoints are always available, and single-view operation is essential for real-time workflows where viewpoint acquisition directly impacts task performance; (2) produce complete meshes for individual objects, as most robotics simulation engines operate on mesh representations \cite{todorov_mujoco_2012,coumans_pybullet_2016,makoviychuk_isaac_2021}; and (3) are category-agnostic, since real-world robotics encounters diverse objects. We emphasise that our evaluation targets the real-to-sim stage: if reconstructions fail to support basic physics simulation, no meaningful policy can be planned or transferred to the real world, regardless of downstream control strategies.

Our systematic evaluation complements existing surveys that categorise 3D reconstruction models generally \cite{maxim_survey_2021,kantarci_survey_2022,bai_survey_2024,wang_deep_2024} or with a focus on specific representations \cite{remondino_critical_2023, fei_3d_2024, zhu_3d_2024, chen_review_2023, lei_whats_2024}. Unlike \cite{zhu_3d_2024, irshad_neural_2024}, which focus on scanning approaches requiring substantial scene coverage, we focus specifically on single-view reconstruction and evaluate on real-world datasets whether reported computer vision performances transfer to realistic robotics domains.

In Section \ref{sec:desiderata}, we delineate desiderata for single-view 3D reconstruction that robotics domains necessitate. Section \ref{sec:taxonomy} provides a brief overview of existing 3D reconstruction approaches (with a comprehensive taxonomy in the Appendix). In Section \ref{sec:experiments}, we systematically evaluate a selection of models towards these desiderata. Finally, in Section \ref{sec:discussion} we summarise and discuss our findings and position them in the broader context of current research directions.

\section{Desiderata}
\label{sec:desiderata}
\noindent For robotic manipulation tasks such as grasping, placing, and rearranging objects on tabletops, single-view 3D reconstruction models must fulfil several requirements for the reconstructed geometry to be physically meaningful and allow for realistic simulation interactions. Below, we motivate five desiderata grounded in the robotics scenarios described in Section \ref{sec:introduction}. Violating these requirements reliably predicts failure within the downstream task class, while satisfaction may not be sufficient. We position our chosen operating point in relation to the spectrum of acceptable values documented in studies on manipulation. When earlier research establishes a range, our setting aligns with the more lenient boundary where possible, ensuring that any shortcomings we observe are not due to an overly stringent criterion. The appropriate end of the range, strict or lenient, depends on the task: fine-grained prediction of contact during grasping requires tighter precision, whereas assessing general collision or object placement can rely on broader tolerances. Furthermore, deployment in the real world imposes constraints on input data and computational resources. Additional discussion of domain gap considerations is provided in Appendix \ref{app:desiderata_details} alongside an overview of the desiderata in Table \ref{tab:desiderata_tab}.

\noindent\textbf{Reconstruction Accuracy.} Robot manipulation is highly sensitive to surface accuracy. For successful grasping and manipulation, surface estimates should be within 1\,mm of the true object surface \cite{morgan_vision-driven_2021,zha_semantic_2020}, though grasping success also depends on gripper design, control strategy (e.g., closed-loop feedback, pre-grasp motions), and the grasp planning algorithm itself. The accuracy requirement should therefore be understood as one necessary factor among several, rather than a standalone guarantee of task success. Some tasks, such as assembly, require sub-millimetre tolerances \cite{ota_autonomous_2024,niu_tolerance-guided_2021,lim_grasping_2023}. For our setting of household environments, we set a target of 5\,mm which is at the permissive end of the reactive planning and safety-checking scenario range. Failure at 5\,mm thus indicates failure across the entire task-tied range, as well as task-tied ranges of stricter task classes. At this error level, a physics simulator can reliably predict whether a grasp will result in a collision or whether an object placement is stable.

\vspace{-5pt}
\newhypbox{1}{
    Chamfer distances between reconstructed and ground truth meshes should be within 5\,mm. Exceeding this threshold reliably predicts failure in household robot manipulation tasks and physical simulation, though meeting it alone does not guarantee downstream task success (see Section \ref{sec:experiments}).
}
\vspace{-5pt}

\noindent\textbf{Object Collision Constraints.} Physics simulators resolve collisions through penalty- or constraint-based methods \cite{todorov_mujoco_2012,coumans_pybullet_2016,makoviychuk_isaac_2021}. Many 3D reconstruction approaches process objects individually, which can result in collisions when placing reconstructions back into the scene \cite{tochilkin_triposr_2024, liu_syncdreamer_2023, liu_zero-1--3_2023}. Scenes with collisions at initialisation break down during physics simulation due to erratic corrective forces and render a simulation uninformative. This desideratum is thus invariant to task choice as corrective forces are applied in the physics engine itself.

\vspace{-5pt}
\newhypbox{2}{
    No object should violate space occupied by other objects or parts of the environment. We enforce a zero-tolerance policy for collisions during initialisation, providing a conservative estimate of potential failures. Violations cause erratic forces at simulation initialisation, rendering downstream planning uninformative regardless of geometric accuracy.
}
\vspace{-5pt}

\newpage
\noindent\textbf{Object Stability.} Observed scenes are assumed to be in a stable, resting state \cite{ni_phyrecon_2024,guo_physically_2024,chen_atlas3d_2024,yan_phycage_2024}. Object pose and surroundings are necessary conditioning factors, since reconstruction could otherwise ignore parts critical for stability $-$ especially occluded and low-surface-area regions. We adopt a 5$^\circ$ threshold for simulation-in-the-loop safety checking: an object that topples upon advancing physics renders the simulation useless for action prediction. In viewpoint-inaccessible scenarios (e.g. objects on a shelf), stability is critical because the robot must plan contact-rich actions relying on neighbouring objects remaining upright.

\vspace{-5pt}
\newhypbox{3}{
    The reconstructed meshes should offer physically stable poses within 5$^\circ$ tilt of their scene pose. Failure causes reconstructed scenes to collapse under simulation, independently of whether geometric accuracy targets are met. Note that this agreement tolerance between stable and scene pose differs from the settling tolerance in \cite{li_dso_2025} $-$ self-support does not necessarily transfer to scene-pose agreement.
}
\vspace{-5pt}

\noindent\textbf{Partial Occlusion Resolution.} Reconstruction models must handle partial object occlusions that are ubiquitous in the real-world \cite{miao_safe_2023,wang_learning_2025}. We do not expect perfect recovery of occluded geometry $-$ this is fundamentally ill-posed. Rather, we ask whether models can produce plausible completions that do not catastrophically degrade simulation fidelity (e.g., a mug whose occluded handle need not be millimetre-accurate, but should maintain stability and not introduce phantom collisions). This requirement is most directly motivated by the viewpoint-inaccessible scenario: when objects are tightly packed in a shelf or drawer, the robot must reason about full geometry to predict whether pulling one object will dislodge another.

\vspace{-5pt}
\newhypbox{4}{
    For common household objects, meshes should resolve partial occlusions with Chamfer distances no more than 10\% higher than those of visible regions, ensuring accurate simulation of partially-occluded objects. This reflects performance observed in current scene-level approaches, making it a practical benchmark. The acceptable level of degradation may vary depending on the task: more leniency is appropriate when occluded regions are only relevant for free-space estimation. However, precise quantitative bounds for such task-specific variations remain undefined in current studies, which we flag as a direction for future work.
}
\vspace{-5pt}

\noindent\textbf{Computational Efficiency.} Reconstruction must be fast enough for online deployment. Many optimisation-based methods require minutes to hours per object \cite{liu_zero-1--3_2023, qian_magic123_2023, ni_phyrecon_2024, yan_phycage_2024}, while feed-forward approaches produce results within seconds \cite{xu_instantmesh_2024, boss_sf3d_2024, tochilkin_triposr_2024, liu_one-2-3-45_2023,liu_one-2-3-45_2024}. We target a latency range of $<$2s for reconstructing an entire multi-object scene, grounded in the reactive planning scenario where delays beyond a few seconds erode user trust and task fluency in human-robot collaboration settings. This target is lenient relative to real-time control loops operating at much higher frequencies, and the exact requirement depends on whether reconstruction occurs within a closed control loop, at task transitions, or before an episode. We use the reactive planning scenario as our baseline and show in Section \ref{sec:experiments} that current methods surpass this threshold by a significant margin, so our conclusions hold regardless of the exact choice of task-tied latency range. Further discussion of on-device compute constraints and the accuracy/efficiency trade-off is provided in Appendix \ref{app:desiderata_details}.

\vspace{-5pt}
\newhypbox{5}{
    The pipeline should produce a complete scene reconstruction within 2s. This is a deployment constraint determining whether a method is viable in a given operational context, rather than a quality criterion predicting task success.
}

\section{3D Reconstruction $-$ A Brief Overview}
\label{sec:taxonomy}
In this section, we provide a focused overview of 3D reconstruction approaches to position our work within the broader landscape. Given the rapidly evolving nature of this field, we focus on representative methods that illustrate key capabilities and limitations relevant to robotic manipulation. A comprehensive taxonomy covering representation types, input data requirements, optimisation paradigms, scene-level versus object-level approaches, physics-aware reconstruction, and training data considerations is provided in Appendix \ref{app:taxonomy}. Tables \ref{tab:reconstruction_overview_single_1} and \ref{tab:reconstruction_overview_single_2} give an overview of a selection of existing reconstruction models and their properties.

\subsection{Key Distinctions in 3D Reconstruction}
\noindent\textbf{Representations.} Among 3D representations, meshes are the most widely used for robotic simulation, enabling efficient rendering and accurate rigid body collision computation. Simulators are primarily built around meshes \cite{coumans_pybullet_2016,makoviychuk_isaac_2021,mittal_orbit_2023}. While other representations (point clouds, implicit functions \cite{park_deepsdf_2019,mescheder_occupancy_2019}, and 3D Gaussian Splatting (3DGS) \cite{kerbl_3d_2023}) have their merits, meshes are the de-facto standard for rigid body simulation. We note that 3DGS is beginning to be explored for direct simulation \cite{xie_physgaussian_2024}, offering new pathways for combining perception and dynamics. While promising, such approaches remain early-stage and have not yet been demonstrated in robotic manipulation settings. We therefore focus on mesh reconstruction as it aligns with established simulation workflows, but acknowledge this limitation is transient and merits future investigation (see Appendix \ref{app:representation_types} for a full discussion).

\noindent\textbf{Single-view reconstruction.} 3D reconstruction approaches can be categorised by the density of input viewpoints: dense-view, sparse-view, and single-view methods (see Appendix \ref{app:input_data} for a detailed treatment). Single-view reconstruction is arguably the most challenging setting as substantial portions of a scene are unobserved. Due to this ill-posedness, proposed methods are trained on datasets to capture general geometric properties. Single-view reconstruction is particularly central to our work, as it represents the most practical and generalisable scenario for robotic deployment $-$ unlike multi-view methods that require substantial scene coverage or specialised equipment, single-view approaches align with real-world constraints where robots must rapidly assess and interact with novel environments from limited viewpoints.

Broadly, single-view methods fall into two flavours. First, methods that rely solely on the single input view by directly regressing 3D shapes (e.g. \cite{szymanowicz_splatter_2024, fan_point_2017, chen_transformers_2022, li_instant3d_2023}) or via test-time optimisation (e.g. \cite{pavllo_shape_2023, lee_understanding_2022, jain_zero-shot_2022}). Second, methods that leverage multi-view generative models, explicitly generating additional views using diffusion models and then fusing the predicted geometry \cite{tang_lgm_2025, xu_grm_2025, xu_instantmesh_2024, liu_one-2-3-45_2023, liu_one-2-3-45_2024}. Multi-view consistency remains a challenge in the latter approaches as inconsistencies directly translate to reconstruction artifacts.

\noindent\textbf{Feed-forward vs.\ optimisation-based methods.} Feed-forward approaches produce reconstructions in a single forward pass \cite{wang_pixel2mesh_2018, fan_point_2017, tochilkin_triposr_2024, boss_sf3d_2024}, enabling fast inference but are historically limited by small 3D training datasets. The introduction of large-scale datasets such as Objaverse-XL \cite{deitke_objaverse-xl_2023} has enabled category-agnostic feed-forward models. Optimisation-based methods instead iteratively refine a 3D representation using guidance from pre-trained 2D models \cite{poole_dreamfusion_2022, wang_score_2023, liu_zero-1--3_2023, qian_magic123_2023}, often yielding higher fidelity but requiring minutes to hours per object, rendering them impractical for real-time robotics (see Appendix \ref{app:optimisation_types} for details).

\noindent\textbf{Object-level vs.\ scene-level reconstruction.} Object-centric approaches reconstruct objects as distinct entities \cite{wang_pixel2mesh_2018,melas-kyriazi_im-3d_2024,xu_instantmesh_2024,tang_lgm_2025,li_rico_2023}, while wholistic approaches generate unified scene representations \cite{szymanowicz_flash3d_2024, yu_monosdf_2022}. An emerging middle ground reconstructs entire scenes as digital twins while preserving object individuality \cite{yao_cast_2025,dogaru_generalizable_2024,huang_midi_2024,han_reparo_2024}. This hybrid approach is valuable for robotics: it enables scene-level understanding and object-level interaction, and lets algorithms leverage surrounding context to resolve occlusions and spatial relationships. 

\noindent\textbf{Physics-aware reconstruction.} Physical plausibility is essential for simulation. Some approaches formulate loss terms to discourage object collisions using SDF-based representations \cite{wu_object-compositional_2022, zhang_holistic_2021, gao_graphdreamer_2024,li_rico_2023} or bounding boxes \cite{chatterjee_3d-scene-former_2024}. Others use differentiable physics simulation to penalise instability \cite{yan_phycage_2024, ni_phyrecon_2024, chen_atlas3d_2024, mezghanni_physical_2022}. Despite these advances, incorporating physics constraints into reconstruction remains an open challenge, particularly for category-agnostic models operating on novel objects (see Appendix \ref{app:physics} for a detailed discussion).

\subsection{Recent Uses of 3D Reconstruction in Robotic Systems}
3D reconstruction bridges perception and control in robotics $-$ building simulation environments, enabling collision checking, supporting object re-identification, or computing affordances. Here we examine how complete-geometry 3D reconstruction is being integrated into robotic manipulation pipelines. For broader surveys on related topics, we refer to works on SLAM \cite{yue_lidar-based_2024,al-tawil_review_2024}, 3DGS in robotics \cite{zhu_3d_2024}, and neural fields \cite{slapak_neural_2024}.

\textbf{Dense- and multi-view approaches.} 3DGS has recently been adopted for Real2Sim2Real in robotic manipulation \cite{barcellona_dream_2024,han_re3sim_2025,li_robogsim_2024,lou_robo-gs_2024,wu_rl-gsbridge_2024,qureshi_splatsim_2024,jia_discoverse_2024}. Typically, 3D Gaussians are learnt alongside coupled mesh representations from dense viewpoints: 3D Gaussians provide realistic rendering while meshes handle rigid body simulation. Other works forego 3DGS to reconstruct meshes from dense input viewpoints \cite{torne_reconciling_2024, patel_real--sim--real_2024, pfaff_scalable_2025}. While these methods push the state-of-the-art, they all require substantial scene coverage in their input data $-$ a major limitation that restricts their applicability to scenarios where dense observations are available.

\textbf{Single-view approaches.} Recent category-agnostic single-view reconstruction methods have been proposed for robotic manipulation, but with limited scope. \cite{mu_robotwin_2024} generate digital "twins" but rely on language and image generation models to produce domain-randomised environments rather than exact replicas, and do not address object collisions. \cite{agarwal_scenecomplete_2024} demonstrate that pre-trained models can produce digital twins affording parallel grasp poses that transfer to real-world scenes, but do not perform physics simulation. \cite{yao_cast_2025} propose generating physically coherent digital twins using GPT-4v-derived scene graphs for gradient-based physics optimisation, but lack robotics experiments. \cite{katara_gen2sim_2024} mention robotics applications for their approach but provide no experimental evaluation. \cite{lu_manigaussian_2025} produce 3D Gaussians but, as existing simulators do not support 3DGS, they learn a neural dynamics model instead.

Our work addresses important gaps by conducting rigorous empirical evaluations of single-view reconstruction methods specifically in the context of robotic manipulation requirements, providing concrete evidence of current limitations and identifying areas for improvement.

\section{Model Evaluation and Results}
\label{sec:experiments}
\noindent As the field of 3D reconstruction is growing rapidly, we select the following representative set of object- and scene-level reconstruction methods.

\noindent\textbf{Object-Level:} SF3D \cite{boss_sf3d_2024}, InstantMesh \cite{xu_instantmesh_2024}, One2345 \cite{liu_one-2-3-45_2023}, LGM \cite{tang_lgm_2025}, Michelangelo \cite{zhao_michelangelo_2023}, ZeroShape \cite{huang_zeroshape_2024}, Real3D \cite{jiang_real3d_2024}, DSO \cite{li_dso_2025}, TRELLIS.2 \cite{xiang2025trellis2}.

\noindent\textbf{Scene-Level:} MIDI \cite{huang_midi_2024}, Gen3DSR \cite{dogaru_generalizable_2024}, PhysGen3D \cite{chen_physgen3d_2025}, SAM3D \cite{sam3dteam2025sam3d3dfyimages}.

\noindent Appendix \ref{app:model_descriptions} offers a brief description for each model. For an in-depth overview, including the required number of viewpoints, asset and optimisation type, as well as physics considerations of reconstruction methods including those we experimentally evaluate, see Table \ref{tab:reconstruction_overview_single_1}.

\subsection{Evaluation Data}
\noindent We require richly annotated real-world data with complete 3D information. For our experiments, we select two datasets that are representative for robotics applications: parts of the YCB-Video dataset \cite{xiang_posecnn_2018} and the Project Aria Digital Twin dataset \cite{pan_aria_2023}. 
The YCB-Video dataset features standardised objects with varying geometric complexity that have become benchmarks in the robotic manipulation community, while the Aria dataset contains everyday household items arranged in naturalistic configurations that robots would encounter in domestic environments. Together, they span a spectrum from controlled, well-defined object arrangements to complex, cluttered real-world scenes. Both datasets provide ground-truth meshes for all objects, their poses in each camera frame, and additional information such as segmentation masks and camera intrinsics. This rich annotation is crucial for quantitatively evaluating reconstruction accuracy against true object geometry. See Fig. \ref{fig:joint_reconstruction_examples} for example inputs from both datasets. For both datasets, we manually select objects and frames to ensure coverage of different object types, camera view-points, and object poses.

\subsection{Experimental Setup and Common Preprocessing}
\label{sec:methodology}
\noindent Each model receives identical single-view inputs. For each method, we follow the provided data preparation routing, including cropping and resizing of the object to a common size in the image frame.
For all methods that rely on off-the-shelf segmentation models in their pipeline, we provide ground-truth object masks during the reconstruction process. This is done to ensure a fair and consistent comparison independent from off-the-shelf methods.

To evaluate reconstruction error, we need to align the reconstructed mesh to the ground truth. Past works \cite{liu_one-2-3-45_2023,li_dso_2025} have used alignment procedures that we extend as follows. First, we estimate the correct scale of the object. To do this, we compute the standard deviations of the mesh vertices along the top three principle components of each mesh. We then evaluate the ratio of the component-aligned standard deviations of the ground truth mesh w.r.t. the reconstruction and estimate the rescaling parameter as the median of these principal component-aligned ratios.

To align translation and rotation, we sample 512 approximately equidistant unit quaternions from the 3-sphere and use these as initialisations for iterative closest point (ICP) optimisation. The number of initialisations is chosen to avoid local optimisation optima. We then select the transformation with the lowest root mean squared error among all initialisations.

\subsection{Results}
\noindent We now evaluate current single-view 3D reconstruction methods against the robotics-specific desiderata outlined in Section \ref{sec:desiderata}. Violating any one desideratum reliably predicts failure in our downstream task class, while satisfying all of them may not be sufficient for task success, as factors not captured by global geometric metrics (e.g., local surface properties at contact points) also influence outcomes. The following results reveal that current methods violate most of the desiderata, confirming that they are not yet suitable for robotics deployment.

\noindent\textbf{Reconstruction Accuracy.} To evaluate reconstruction accuracy on realistic robotic imagery, we run all models on segmented objects that are entirely visible except for self-occlusion. We select 177 such objects from YCB-Video and 136 objects from the Aria dataset. As there are no guarantees that reconstructed meshes are valid volumes, we evaluate reconstruction accuracy by uniformly sampling 10,000 points from the surfaces of ground truth mesh and reconstruction, and computing the Chamfer distance ($\ell^2$-norm) between both point clouds. Utilisation of the Chamfer distance (CD) as a measure of reconstruction error is in line with recent reconstruction works \cite{liu_zero-1--3_2023,xu_instantmesh_2024,hong_lrm_2023,li_dso_2025}. We follow the implementation of \cite{mescheder_occupancy_2019} for computing the Chamfer distances.

\begin{wrapfigure}{r}{0.49\textwidth}
    \centering
    \vspace{-10pt}
    \resizebox{0.49\textwidth}{!}{\input{plots/ycb_chamfer.pgf}}
        \vspace{-20pt}
        \caption{Chamfer distances on YCB-Video \cite{xiang_posecnn_2018}. CD is averaged across 10,000 sampled surface points from both target and reconstruction. Indicators for min, max, and median are shown. Most achieve $\sim$ 5mm accuracy.}
        \label{fig:ycb_chamfer}
    \vspace{-10pt}
\end{wrapfigure}

Figs. \ref{fig:ycb_chamfer} and \ref{fig:aria_chamfer} depict the resulting Chamfer distances for the single-object models on the two datasets. On YCB-Video \cite{xiang_posecnn_2018}, with the exception of SF3D \cite{boss_sf3d_2024}, DSO \cite{li_dso_2025}, and TRELLIS.2 \cite{xiang2025trellis2}, reconstructed surfaces tend to be about 5mm distant from the closest surface in the ground truth meshes. Reconstruction accuracy on the Aria Digital Twin \cite{pan_aria_2023} dataset is significantly worse, with about double the median reconstruction error. Fig. \ref{fig:joint_reconstruction_examples} depicts a collection of example scenes alongside mesh reconstructions to give the reader an impression of the visual manifestation of such error ranges. Models appear to especially struggle with objects in non-canonical poses, such as view-points towards the top or bottom of the object. Such reconstruction error would result in very different behaviour than the ground truth object, if put into a physics simulator. Further, we evaluate the transfer success rates of grasp poses computed on reconstructions to the ground truth meshes on YCB-Video \cite{xiang_posecnn_2018}. The results in Fig. \ref{fig:ycb_grasps} indicate that the reconstructed meshes are insufficiently accurate for computing grasp poses on the reconstructed meshes and applying them in the real world scene $-$ only about half of all grasp poses can successfully be applied to the target meshes. For grasp sampling, we use the antipodal grasp sampling procedure presented in \cite{gilles_metagraspnet_2022} with a Franka Emika Panda parallel gripper model. Grasp poses computed on the reconstructions are considered to transfer successfully if they are not in collision with the ground truth mesh, the rotations between gripper surface normals and mesh surface normals at the grasp points are not in excess of 22.5$^\circ$, 
and the object is still in contact with the gripper after grasping and shaking it in simulation,
as specified in the implementation of \cite{gilles_metagraspnet_2022}. \cite{gilles_metagraspnet_2022} demonstrate their grasp sampling procedure on common household objects, including YCB objects that also feature in our evaluation. We clarify that end-to-end grasp prediction models, while effective for specific tasks, are not suitable baselines here: they bypass reconstruction and thus cannot be used for forward simulation. Our focus is on whether reconstructed scenes support physics-based reasoning within a digital twin.

To examine the relationship between reconstruction accuracy and downstream task performance, we analyse the per-object correlation between Chamfer distance and grasp transfer success across all evaluated object-level reconstruction models (Fig. \ref{fig:grasp_vs_chamfer}). The overall Pearson correlation is $r = -0.55$ ($r=-0.52$ excluding objects with no valid grasp), indicating a moderate negative association, but the relationship is notably asymmetric. The LOESS fit reveals a threshold effect: Chamfer distance drops sharply from roughly 2cm to 1cm as grasp success increases from 0\% to approximately 5\%, but further improvements up to 100\% are accompanied by only a gradual reduction in Chamfer distance to approximately 0.3cm. The wide confidence interval at zero grasp success reflects heterogeneous failure causes $-$ some objects fail due to poor reconstruction, others are inherently difficult to grasp regardless of reconstruction quality. Two insights follow. First, reconstruction accuracy is \emph{necessary but not sufficient} within the tested task class: high Chamfer distances reliably predict grasp failure, but low distances do not guarantee success. Second, below approximately 1cm Chamfer distance, grasp outcomes become increasingly dominated by factors not captured by global metrics, such as local surface normal accuracy at grasp contact points, mesh topology and watertightness, and the fidelity of fine geometric features like edges and thin structures. These findings strengthen our argument that the proposed desiderata are predictive of the inadequacy of current single-view reconstruction models in robotics.

\begin{figure*}[t]
    \centering
    \begin{minipage}[t]{0.49\textwidth}
        \centering
        \resizebox{\textwidth}{!}{\input{plots/ycb_grasps_sim.pgf}}
        \vspace{-14pt}
        \caption{Grasp transfer success rates from reconstruction to target mesh on YCB-Video \cite{xiang_posecnn_2018}. Indicated are 95\% Wilson intervals and the number of successfully sampled grasps.}
        \label{fig:ycb_grasps}
        \vspace{-10pt}
    \end{minipage}
    \hfill
    \begin{minipage}[t]{0.49\textwidth}
        \centering
        \vspace{-90pt}
        \resizebox{\textwidth}{!}{\input{plots/ycb_grasp_vs_chamfer.pgf}}
        \vspace{-20pt}
        \caption{Chamfer distance against grasp success rate and LOESS regression with 95\% confidence interval on YCB-Video \cite{xiang_posecnn_2018} pooled across all object-level reconstruction models. 
        }
    \label{fig:grasp_vs_chamfer}
        \vspace{-10pt}
    \end{minipage}
\end{figure*}

This threshold effect also explains the seemingly contradictory observation that InstantMesh \cite{xu_instantmesh_2024} achieves similar Chamfer distances to ZeroShape \cite{huang_zeroshape_2024} on YCB-Video \cite{xiang_posecnn_2018} (Fig. \ref{fig:ycb_chamfer}) yet exhibits substantially worse grasp transfer performance (Fig. \ref{fig:ycb_grasps}). Both methods operate in the sub-0.5cm Chamfer distance regime where global geometric accuracy is no longer the dominant predictor of grasp outcomes. Instead, ZeroShape's advantage likely stems from producing more physically plausible local geometry, corroborating the decoupling of global surface error and task performance beyond a certain accuracy threshold.

To evaluate the reconstruction quality of models that process entire scenes, we select 28 scenes from the YCB-Video \cite{xiang_posecnn_2018} dataset, sample 10,000 points each from the reconstructed and ground truth surfaces of the \textit{entire} scene, align the reconstructed scene \textit{as a whole} with the target scene using the ICP process described in Section \ref{sec:methodology}, and compute the Chamfer distances between reconstruction and target, as we have done in the single-object case. Fig. \ref{fig:scene_ycb_chamfer} shows that models that reconstruct entire scenes perform worse on average than single-object reconstruction models. With the median error being between 1cm and 1.5cm, the reconstructed scene and its objects would behave very differently when used in a physics simulation. Such error ranges are too large for robust robotic manipulation and miss our formulated reconstruction targets by a large margin. Fig. \ref{fig:scene_object_distances} separately evaluates the pose estimation of scene reconstruction models by looking at the relative Chamfer distance between two objects in the reconstruction and computing the difference to the relative distance between the same objects in the ground truth scene. The results show that the relative object poses are not preserved in the reconstruction, which is a requirement for both accurate physics simulation and being able to transfer manipulation trajectories from simulation to the real world. Fig. \ref{fig:scene_reconstruction_examples} depicts several reconstruction examples of the scene-level models. It is clear that both reconstruction accuracy and pose estimation are suboptimal.

\begin{figure*}[t]
    \centering
    \begin{minipage}[t]{0.49\textwidth}
    \vspace{-90pt}
    \resizebox{\textwidth}{!}{\input{plots/collision_pie.pgf}}
    \vspace{-20pt}
    \caption{Frequency of reconstructed scenes where reconstructed meshes are in collision. Colliding objects result in unwanted behaviour in physics simulation, such as instantaneous acceleration and collapse of the scene composition.
    }
    \label{fig:collision_pie}
    \vspace{-10pt}
    \end{minipage}
    \hfill
    \begin{minipage}[t]{0.49\textwidth}
    \resizebox{\textwidth}{!}{\input{plots/ycb_stability.pgf}}
    \vspace{-20pt}
    \caption{Total number of objects with physically stable poses within 5$^\circ$ from the ground truth scene pose versus total number of objects without such stable poses. Evaluated on YCB-Video. For results on the Aria Digital Twin \cite{pan_aria_2023} dataset, see Fig. \ref{fig:aria_stability}. 
    }
    \label{fig:ycb_stability}
        \vspace{-10pt}
    \end{minipage}
\end{figure*}

\noindent\textbf{Object Collision Constraints.} To evaluate how often inaccurate reconstructions result in mesh collisions, we select object pairs that are spatially close, reconstruct them individually, place them at their respective scene poses obtained through the ICP procedure described in Section \ref{sec:methodology}, and use FCL \cite{pan_fcl_2012} to compute mesh collisions. The evaluation dataset consists of 96 samples for YCB-Video \cite{xiang_posecnn_2018} and 64 samples for the Aria Digital Twin \cite{pan_aria_2023} dataset. 

Fig. \ref{fig:collision_pie} shows that all single-object reconstruction models produce meshes that are in collision in a majority of the evaluation scenes. If used in a physics simulation, such collisions at initialisation would result in unwanted behaviour, such as instantaneous acceleration and collapse of the scene composition, and are thus unsuitable. Fig. \ref{fig:scene_collision_pie} shows that even though scene reconstruction models operate on entire scenes and thus in principle should be in a good position to avoid mesh collisions in their reconstructions through conditioning one object's reconstructed mesh on other objects' scene occupancy, as well as having control over the positioning of individual objects, their ability to avoid mesh collisions is only on par with the better single-object models. Only PhysGen3D shows vastly reduced mesh collisions. However, this may be an artifact of the model significantly overestimating relative object distances (Fig. \ref{fig:scene_object_distances}) which would automatically result in reduced object collisions.

\noindent\textbf{Object Stability.} To evaluate whether reconstructed objects would be stable when put back into the scene, we first compute their stable orientations, that are all orientations where the centre of mass is within the 2D convex hull of vertices that have minimal elevation. We then filter for those stable orientations where the z-axis is at most 5$^\circ$ tilted from the z-axis of the object in the scene, i.e. where the stable pose is similarly upright as the object in the scene. For each of the resulting candidate poses, we generate 8 equally spaced perturbations that rotate the object 5$^\circ$ away from its stable pose. If in a subsequent physics simulation using PyBullet \cite{coumans_pybullet_2016} at least one of these perturbed poses leads to the object reverting back to the unperturbed pose, the unperturbed pose is considered stable. As a result, we obtain all poses that are physically robust to slight perturbation while also being close to the ground truth pose of the object in the scene. In our evaluation, we only consider objects that are resting on planes, not ones that are leaning against other objects. We use 146 samples from YCB-Video \cite{xiang_posecnn_2018} and 133 samples from the Aria Digital Twin \cite{pan_aria_2023} dataset. 

Fig. \ref{fig:ycb_stability} shows that the vast majority of YCB-Video reconstructions does not have stable poses near the observed scene pose. Michelangelo \cite{zhao_michelangelo_2023} stands out among all other evaluated models with two thirds of the reconstructed meshes having valid stable poses. On the Aria Digital Twin \cite{pan_aria_2023} dataset, however, Michelangelo does not show the same advantage. Instead, all evaluated models produce even fewer stable mesh reconstructions than on YCB-Video. As a result, the reconstructed meshes would be largely unusable for physics simulation without significant, non-trivial postprocessing, as without it the object meshes would topple over as soon as physics steps are advanced. 

Fig. \ref{fig:scene_ycb_stability} shows that the scene reconstruction models do not fare any better. Instead, SAM3D \cite{sam3dteam2025sam3d3dfyimages} shows the least number of stable meshes among all evaluated models. To evaluate whether the reconstructed meshes themselves or the pose estimation of the scene reconstruction models are to blame, we align each reconstructed object mesh with the ground truth mesh separately and compute the number of stable poses again. The results are shown in Fig. \ref{fig:scene_ycb_stability_with_fit}. The number of stable poses increases significantly, especially for SAM3D, corroborating that the scene reconstruction models struggle to estimate the correct object poses. Still, the number of stable poses remains too low for practical use in robotics applications.

\begin{wrapfigure}{r}{0.49\textwidth}
    \vspace{-15pt}
    \centering
    \resizebox{0.49\textwidth}{!}{\input{plots/ycb_occlusion_chamfer.pgf}}
    \vspace{-25pt}
    \caption{Chamfer distances (CD) of unoccluded object parts versus occluded object parts. CD is averaged across occluded/unoccluded surface points from an initial set of 10,000 uniformly sampled surface points. Indicators for min, max, and median are shown. 
    }
    \label{fig:ycb_occlusion_chamfer}
    \vspace{-10pt}
\end{wrapfigure}

\noindent\textbf{Partial Occlusion Resolution.} To evaluate how well reconstruction models are able to resolve occlusions, we select a subset of 80 partially occluded objects from YCB-Video \cite{xiang_posecnn_2018} dataset and a subset of 76 partially occluded objects from the Aria Digital Twin \cite{pan_aria_2023} dataset. During the evaluation, we noticed that many methods struggle with reconstructing occluded object parts and often ignore such parts entirely. This poses a challenge to the ICP procedure we use to align reconstructions to target meshes as we would be trying to align a partial reconstruction to a complete mesh, while relying on $k$-nearest neighbour search. To mitigate this, we use the object masks from the datasets to ignore any points during the ICP step, ground truth or reconstruction, that would be occluded in the given scene if rendered from the true camera view point. This mask is re-computed at every ICP iteration. As a result, we only align the visible parts of the object meshes. Furthermore, we noticed that objects with symmetry axes are sometimes misaligned, resulting in incorrect occlusion error calculations. We flip these objects to their correct rotation as this rotation misalignment is an artifact from the alignment process, not the reconstruction process. This ensures a fair comparison between methods. In cases of severe occlusion, the number of available surface points for alignment is significantly reduced, which inherently weakens the robustness of the ICP procedure. This constraint is unavoidable given the lack of observable geometry and should be considered when interpreting occlusion-specific performance.
We evaluate how well the given reconstruction models estimate occluded structure by comparing the Chamfer distances of unoccluded object parts with the Chamfer distances of occluded object parts. Fig. \ref{fig:ycb_occlusion_chamfer} shows that occlusion does lead to heavily increased reconstruction inaccuracy. The trend is consistent across evaluated models with median Chamfer distances being between 40-105$\%$ higher at occluded regions. This is problematic as occluded object parts are often relevant for manipulation. Furthermore, when using such objects in simulation, the simulation would be based on a mesh that is not representative of the real-world object.

Scene reconstruction models fare much better in comparison. Fig. \ref{fig:scene_ycb_occlusion} shows that across all models, the increase in median Chamfer distance is at most about 10\%. MIDI considers objects jointly during reconstruction, PhysGen3D and Gen3DSR both utilise image inpainting models to complete partial occlusions before mesh reconstruction, and SAM3D uses artificial occlusions during training. All approaches appear well-suited for maintaining reconstruction accuracy at occluded object regions. For this evaluation, we individually aligned each reconstructed object to the target object mesh to avoid compounding error from erroneous pose estimations by the model, as shown in Fig. \ref{fig:scene_object_distances}.

\begin{figure*}[t]
    \centering
    \begin{minipage}[t]{0.49\textwidth}
        \vspace{-80pt}
        \resizebox{\textwidth}{!}{\input{plots/scene_ycb_object_distances.pgf}}
        \vspace{-30pt}
        \caption{Relative object position error of scene-level reconstruction models SAM3D \cite{sam3dteam2025sam3d3dfyimages}, PhysGen3D \cite{chen_physgen3d_2025}, Gen3DSR \cite{dogaru_generalizable_2024}, and MIDI \cite{huang_midi_2024}. Indicators for min, max, median, and 25\textsuperscript{th} and 75\textsuperscript{th} quantile are shown. We evaluate the difference between the Chamfer distances (CD) between all object pairs in the ground truth scene, and the CDs between the corresponding object pairs in the reconstructed scene with relative object poses estimated by the reconstruction models.
        }
        \label{fig:scene_object_distances}
        \vspace{-10pt}
    \end{minipage}
    \hfill
    \begin{minipage}[t]{0.49\textwidth}
        \resizebox{\textwidth}{!}{\input{plots/ycb_computation.pgf}}
        \vspace{-20pt}
        \caption{Computational requirements during inference of all evaluated methods. The VRAM utilisation is in line what current on-device systems such as NVIDIA Jetson devices offer. However, the GPU used in this experiment is significantly faster than the Jetson devices. As such, the reconstruction time poses a major challenge for on-device systems as even on the more powerful evaluation system, reconstruction times are in the order of seconds.
        }
        \label{fig:compute}
        \vspace{-10pt}
    \end{minipage}
\end{figure*}

\noindent\textbf{Computational and Memory Efficiency.} We measure the mean reconstruction time per object and the peak memory allocation during inference on the same data as used in our reconstruction accuracy evaluation. All our experiments are run on a server node with four 2.1GHz CPU cores, 64GB of RAM, and a NVIDIA RTX A6000 GPU with 48GB of VRAM. We note that the VRAM capacity is roughly in line with what current NVIDIA Jetson devices offer for on-robot compute. However, due to the significantly higher speed of our evaluation GPU, the reconstruction times reported here should be understood as a lower-bound projection. Actual inference latency on Jetson hardware would likely be considerably higher. Fig. \ref{fig:compute} depicts that most evaluated models take multiple, sometimes tens of seconds for reconstructing an individual object. SF3D \cite{boss_sf3d_2024} and ZeroShape \cite{huang_zeroshape_2024} stand out by producing reconstruction within roughly 0.5 and 1 second, respectively. Still, reconstruction time scales linearly in the number of objects. We also observe large differences in memory allocation during inference. While SF3D \cite{boss_sf3d_2024}, LGM \cite{tang_lgm_2025}, Michelangelo \cite{zhao_michelangelo_2023}, ZeroShape \cite{huang_zeroshape_2024}, and SAM3D \cite{sam3dteam2025sam3d3dfyimages} all consume less than 10GB of VRAM, InstantMesh \cite{xu_instantmesh_2024} and One2345 \cite{liu_one-2-3-45_2023} require roughly triple and double the memory, respectively.

Comparing reconstruction error (Figs. \ref{fig:ycb_chamfer} and \ref{fig:aria_chamfer}) and computational requirements (Fig. \ref{fig:compute}), we do not observe a clear-cut trend that suggests computationally more expensive models, be it from having more parameters or generating multiple viewpoints, automatically produce more accurate reconstructions.

Scene reconstruction models require substantially more time to reconstruct their inputs than single-object reconstruction models (Fig. \ref{fig:scene_computation}). This appears sensible, as these models reconstruct multiple objects and also come with overhead to align reconstructed meshes. At the same time, VRAM utilisation is not too dissimilar to the single-object reconstruction models. The efficiency evaluation does not depend on the specific latency requirement: none of the methods assessed can reconstruct a multi-object scene within a timeframe suitable for reactive deployment, even when significantly more lenient targets are considered.

\section{Discussion and Avenues for Future Research}
\label{sec:discussion}
\noindent We evaluated single-view 3D reconstruction for creating digital twin environments for robot manipulation tasks, revealing significant limitations despite computer vision advances. Current single-view methods violate the desiderata for even the moderate-fidelity regime required by reactive planning and simulation-based safety checking. This finding is significant precisely because these use cases do not demand sub-millimetre accuracy $-$ they represent the most forgiving realistic deployment scenarios for single-view reconstruction. The fact that even these requirements are unmet underscores the severity of the current gap. Conversely, our evaluation reveals which aspects (e.g., physical stability, collision avoidance) are the most critical bottlenecks, providing targeted guidance for future work.

Our analysis of the relationship between reconstruction accuracy and grasp transfer success provides empirical grounding for reasoning about the relative importance of the proposed desiderata. Reconstruction accuracy (Desideratum 1) functions as a gating criterion: if it is not met, downstream tasks will almost certainly fail. However, meeting the accuracy requirement alone is insufficient, as our data show that objects with low Chamfer distance can still exhibit zero grasp success. Physical stability (Desideratum 3) and collision avoidance (Desideratum 2) capture independent failure modes that persist even when geometric accuracy is adequate $-$ a mesh that is geometrically close to the ground truth but topples upon physics simulation or intersects neighbouring objects is functionally unusable for simulation-based planning. Computational efficiency (Desideratum 5) is best understood as a deployment constraint rather than a quality metric: it determines whether a given reconstruction model can be used in a particular operational context but does not directly affect the fidelity of the reconstruction itself. Occlusion handling (Desideratum 4) modulates all other desiderata, as quality at occluded regions determines whether the effective accuracy and stability of the full object mesh are maintained. In practice, we suggest a hierarchical view of these conditions: practitioners should first verify that a reconstruction method produces stable, collision-free geometry before optimising for geometric accuracy, as instability and collision render even geometrically faithful reconstructions unusable in simulation. Conversely, meeting all desiderata does not guarantee task success $-$ our per-object analysis (Fig. \ref{fig:grasp_vs_chamfer}) demonstrates that factors beyond our metrics influence manipulation outcomes.

The gap between computer vision capabilities and robotics requirements suggests several future research directions. First, models trained on robotics-relevant data $-$ simulated sensor noise, varied lighting, realistic occlusions, lower-resolution imagery, and challenging viewpoints $-$ would better address the domain-specific challenges we identified. Future datasets should reflect real-world capture conditions rather than solely increasing synthetic asset fidelity. The 40-105\% degradation at occluded parts emphasises this need. Scene-level models using inpainting or joint denoising perform adequately in our evaluation, and physics-aware extensions to inpainting are promising.
We advise practitioners against using single-object reconstruction models when objects are physically close, as these models cannot avoid inter-object collisions by design. Scene reconstruction models should be preferred, though our evaluation shows they also struggle with physical plausibility. Object stability remains a major limitation across all evaluated models. Future work should train for scene pose agreement, not only self-support. A detailed discussion of depth integration opportunities and scene-level training data limitations is provided in Appendix \ref{app:discussion_details}.

Our findings emphasise the need for closer collaboration between the computer vision and robotics communities to develop reconstruction techniques that specifically address the unique requirements of robot manipulation tasks.

\clearpage
\newpage

\section*{Acknowledgments}
\noindent The authors thank Joe Watson, Jack Collins, Jun Yamada, and Alex Mitchell for fruitful discussions and valuable assistance in the preparation of this manuscript.
This work is supported by the European Laboratory for Learning and Intelligent Systems and a UKRI/EPSRC Programme Grant [EP/V000748/1]. We would also like to thank SCAN and ARC for use of their GPU acceleration facilities.
Frederik Nolte is supported by the AWS Lighthouse Scholarship. Andreas Geiger and Bernhard Sch\"olkopf are members of the Tübingen AI Center.

\bibliography{references}

\begin{thebibliography}{214}
\providecommand{\natexlab}[1]{#1}
\providecommand{\url}[1]{\texttt{#1}}
\expandafter\ifx\csname urlstyle\endcsname\relax
  \providecommand{\doi}[1]{doi: #1}\else
  \providecommand{\doi}{doi: \begingroup \urlstyle{rm}\Url}\fi

\bibitem[Abdulsalam and Hedabou(2022)]{abdulsalam_security_2022}
Yunusa~Simpa Abdulsalam and Mustapha Hedabou.
\newblock Security and {Privacy} in {Cloud} {Computing}: {Technical} {Review}.
\newblock \emph{Future Internet}, 2022.

\bibitem[Abou-Chakra et~al.(2024)Abou-Chakra, Rana, Dayoub, and Sünderhauf]{abou-chakra_physically_2024}
Jad Abou-Chakra, Krishan Rana, Feras Dayoub, and Niko Sünderhauf.
\newblock Physically {Embodied} {Gaussian} {Splatting}: {A} {Realtime} {Correctable} {World} {Model} for {Robotics}, 2024.

\bibitem[Agarwal et~al.(2024)Agarwal, Singh, Sen, Lozano-Pérez, and Kaelbling]{agarwal_scenecomplete_2024}
Aditya Agarwal, Gaurav Singh, Bipasha Sen, Tomás Lozano-Pérez, and Leslie~Pack Kaelbling.
\newblock {SceneComplete}: {Open}-{World} {3D} {Scene} {Completion} in {Complex} {Real} {World} {Environments} for {Robot} {Manipulation}, 2024.

\bibitem[Al-Tawil et~al.(2024)Al-Tawil, Hempel, Abdelrahman, and Al-Hamadi]{al-tawil_review_2024}
Basheer Al-Tawil, Thorsten Hempel, Ahmed Abdelrahman, and Ayoub Al-Hamadi.
\newblock A review of visual {SLAM} for robotics: evolution, properties, and future applications.
\newblock \emph{Frontiers in Robotics and AI}, 2024.

\bibitem[Alterovitz et~al.(2016)Alterovitz, Koenig, and Likhachev]{alterovitz_robot_2016}
Ron Alterovitz, Sven Koenig, and Maxim Likhachev.
\newblock Robot {Planning} in the {Real} {World}: {Research} {Challenges} and {Opportunities}.
\newblock \emph{AI Magazine}, 2016.

\bibitem[Aulinas et~al.(2008)Aulinas, Petillot, Salvi, Llado, and Xavier]{aulinas_slam_2008}
Josep Aulinas, Yvan Petillot, Joaquim Salvi, Llado, and Xavier.
\newblock The {SLAM} problem: a survey.
\newblock In \emph{Artificial {Intelligence} {Research} and {Development}}. Frontiers in Artificial Intelligence and Applications, 2008.

\bibitem[Bai et~al.(2024)Bai, Wong, and Twan]{bai_survey_2024}
Yonge Bai, LikHang Wong, and TszYin Twan.
\newblock Survey on {Fundamental} {Deep} {Learning} {3D} {Reconstruction} {Techniques}, 2024.

\bibitem[Barcellona et~al.(2024)Barcellona, Zadaianchuk, Allegro, Papa, Ghidoni, and Gavves]{barcellona_dream_2024}
Leonardo Barcellona, Andrii Zadaianchuk, Davide Allegro, Samuele Papa, Stefano Ghidoni, and Efstratios Gavves.
\newblock Dream to {Manipulate}: {Compositional} {World} {Models} {Empowering} {Robot} {Imitation} {Learning} with {Imagination}, 2024.

\bibitem[Baruch et~al.(2021)Baruch, Chen, Dehghan, Feigin, Fu, Gebauer, Kurz, Dimry, Joffe, Schwartz, and Shulman]{baruch_arkitscenes_2021}
Gilad Baruch, Zhuoyuan Chen, Afshin Dehghan, Yuri Feigin, Peter Fu, Thomas Gebauer, Daniel Kurz, Tal Dimry, Brandon Joffe, Arik Schwartz, and Elad Shulman.
\newblock {ARKitScenes}: {A} {Diverse} {Real}-{World} {Dataset} {For} {3D} {Indoor} {Scene} {Understanding} {Using} {Mobile} {RGB}-{D} {Data}.
\newblock In \emph{NeurIPS 2021 Datasets and Benchmarks Track}, 2021.

\bibitem[Berkenkamp et~al.(2017)Berkenkamp, Turchetta, Schoellig, and Krause]{berkenkamp_safe_2017}
Felix Berkenkamp, Matteo Turchetta, Angela Schoellig, and Andreas Krause.
\newblock Safe {Model}-based {Reinforcement} {Learning} with {Stability} {Guarantees}.
\newblock In \emph{Advances in {Neural} {Information} {Processing} {Systems}}, 2017.

\bibitem[Bohg et~al.(2017)Bohg, Hausman, Sankaran, Brock, Kragic, Schaal, and Sukhatme]{bohg_interactive_2017}
Jeannette Bohg, Karol Hausman, Bharath Sankaran, Oliver Brock, Danica Kragic, Stefan Schaal, and Gaurav~S. Sukhatme.
\newblock Interactive {Perception}: {Leveraging} {Action} in {Perception} and {Perception} in {Action}.
\newblock \emph{IEEE Transactions on Robotics}, 2017.

\bibitem[Bohlinger et~al.(2025)Bohlinger, Kinzel, Palenicek, Antczak, and Peters]{bohlinger_gait_2025}
Nico Bohlinger, Jonathan Kinzel, Daniel Palenicek, Lukasz Antczak, and Jan Peters.
\newblock Gait in {Eight}: {Efficient} {On}-{Robot} {Learning} for {Omnidirectional} {Quadruped} {Locomotion}, 2025.

\bibitem[Boss et~al.(2024)Boss, Huang, Vasishta, and Jampani]{boss_sf3d_2024}
Mark Boss, Zixuan Huang, Aaryaman Vasishta, and Varun Jampani.
\newblock {SF3D}: {Stable} {Fast} {3D} {Mesh} {Reconstruction} with {UV}-unwrapping and {Illumination} {Disentanglement}, 2024.

\bibitem[Breyer et~al.(2022)Breyer, Ott, Siegwart, and Chung]{breyer_closed-loop_2022}
Michel Breyer, Lionel Ott, Roland Siegwart, and Jen~Jen Chung.
\newblock Closed-{Loop} {Next}-{Best}-{View} {Planning} for {Target}-{Driven} {Grasping}.
\newblock In \emph{2022 {IEEE}/{RSJ} {International} {Conference} on {Intelligent} {Robots} and {Systems} ({IROS})}, 2022.

\bibitem[Calli et~al.(2015)Calli, Singh, Walsman, Srinivasa, Abbeel, and Dollar]{calli_ycb_2015}
Berk Calli, Arjun Singh, Aaron Walsman, Siddhartha Srinivasa, Pieter Abbeel, and Aaron~M. Dollar.
\newblock The {YCB} object and {Model} set: {Towards} common benchmarks for manipulation research.
\newblock In \emph{2015 {International} {Conference} on {Advanced} {Robotics} ({ICAR})}, 2015.

\bibitem[Chabal et~al.(2025)Chabal, Chen, Ponce, and Schmid]{chabal_online_2025}
Thomas Chabal, Shizhe Chen, Jean Ponce, and Cordelia Schmid.
\newblock Online {3D} {Scene} {Reconstruction} {Using} {Neural} {Object} {Priors}.
\newblock In \emph{{3DV} 2025 - 12th {International} {Conference} on {3D} {Vision} 2025}, 2025.

\bibitem[Chan et~al.(2022)Chan, Lin, Chan, Nagano, Pan, De~Mello, Gallo, Guibas, Tremblay, Khamis, Karras, and Wetzstein]{chan_efficient_2022}
Eric~R. Chan, Connor~Z. Lin, Matthew~A. Chan, Koki Nagano, Boxiao Pan, Shalini De~Mello, Orazio Gallo, Leonidas~J. Guibas, Jonathan Tremblay, Sameh Khamis, Tero Karras, and Gordon Wetzstein.
\newblock Efficient {Geometry}-{Aware} {3D} {Generative} {Adversarial} {Networks}.
\newblock In \emph{Proceedings of the IEEE/CVF Conference on Computer Vision and Pattern Recognition (CVPR)}, 2022.

\bibitem[Chang et~al.(2017)Chang, Dai, Funkhouser, Halber, Niebner, Savva, Song, Zeng, and Zhang]{chang_matterport3d_2017}
Angel Chang, Angela Dai, Thomas Funkhouser, Maciej Halber, Matthias Niebner, Manolis Savva, Shuran Song, Andy Zeng, and Yinda Zhang.
\newblock {Matterport3D}: {Learning} from {RGB}-{D} {Data} in {Indoor} {Environments}.
\newblock In \emph{2017 International Conference on 3D Vision (3DV)}, 2017.

\bibitem[Chang et~al.(2015)Chang, Funkhouser, Guibas, Hanrahan, Huang, Li, Savarese, Savva, Song, Su, Xiao, Yi, and Yu]{chang_shapenet_2015}
Angel~X. Chang, Thomas Funkhouser, Leonidas Guibas, Pat Hanrahan, Qixing Huang, Zimo Li, Silvio Savarese, Manolis Savva, Shuran Song, Hao Su, Jianxiong Xiao, Li~Yi, and Fisher Yu.
\newblock {ShapeNet}: {An} {Information}-{Rich} {3D} {Model} {Repository}, 2015.

\bibitem[Charatan et~al.(2024)Charatan, Li, Tagliasacchi, and Sitzmann]{charatan_pixelsplat_2024}
David Charatan, Sizhe~Lester Li, Andrea Tagliasacchi, and Vincent Sitzmann.
\newblock {pixelSplat}: {3D} {Gaussian} {Splats} from {Image} {Pairs} for {Scalable} {Generalizable} {3D} {Reconstruction}.
\newblock In \emph{Proceedings of the IEEE/CVF Conference on Computer Vision and Pattern Recognition (CVPR)}, 2024.

\bibitem[Chatterjee and Torres Vega(2024)]{chatterjee_3d-scene-former_2024}
Jit Chatterjee and Maria Torres Vega.
\newblock {3D}-{Scene}-{Former}: {3D} scene generation from a single {RGB} image using {Transformers}.
\newblock \emph{The Visual Computer}, 2024.

\bibitem[Chen et~al.(2025)Chen, Jiang, Liu, Gupta, Li, Zhao, and Wang]{chen_physgen3d_2025}
Boyuan Chen, Hanxiao Jiang, Shaowei Liu, Saurabh Gupta, Yunzhu Li, Hao Zhao, and Shenlong Wang.
\newblock {PhysGen3D}: {Crafting} a {Miniature} {Interactive} {World} from a {Single} {Image}, 2025.

\bibitem[Chen et~al.(2023{\natexlab{a}})Chen, Gu, Chen, Tian, Tu, Liu, and Su]{chen_single-stage_2023}
Hansheng Chen, Jiatao Gu, Anpei Chen, Wei Tian, Zhuowen Tu, Lingjie Liu, and Hao Su.
\newblock Single-{Stage} {Diffusion} {NeRF}: {A} {Unified} {Approach} to {3D} {Generation} and {Reconstruction}.
\newblock In \emph{Proceedings of the IEEE/CVF International Conference on Computer Vision (ICCV)}, 2023{\natexlab{a}}.

\bibitem[Chen et~al.(2023{\natexlab{b}})Chen, Chen, Jiao, and Jia]{chen_fantasia3d_2023}
Rui Chen, Yongwei Chen, Ningxin Jiao, and Kui Jia.
\newblock {Fantasia3D}: {Disentangling} {Geometry} and {Appearance} for {High}-quality {Text}-to-{3D} {Content} {Creation}.
\newblock In \emph{Proceedings of the IEEE/CVF International Conference on Computer Vision (ICCV)}, 2023{\natexlab{b}}.

\bibitem[Chen and Wang(2022)]{chen_transformers_2022}
Yinbo Chen and Xiaolong Wang.
\newblock Transformers as {Meta}-learners for {Implicit} {Neural} {Representations}.
\newblock In \emph{Computer {Vision} – {ECCV} 2022}, 2022.

\bibitem[Chen et~al.(2024{\natexlab{a}})Chen, Ni, Jiang, Zhang, Zhu, and Huang]{chen_single-view_2024}
Yixin Chen, Junfeng Ni, Nan Jiang, Yaowei Zhang, Yixin Zhu, and Siyuan Huang.
\newblock Single-view {3D} {Scene} {Reconstruction} with {High}-fidelity {Shape} and {Texture}.
\newblock In \emph{2024 {International} {Conference} on {3D} {Vision} ({3DV})}, 2024{\natexlab{a}}.

\bibitem[Chen et~al.(2024{\natexlab{b}})Chen, Xie, Ye, Yang, Xie, Chen, Li, and Huo]{chen_2l3_2024}
Yizheng Chen, Rengan Xie, Qi~Ye, Sen Yang, Zixuan Xie, Tianxiao Chen, Rong Li, and Yuchi Huo.
\newblock {2L3}: {Lifting} {Imperfect} {Generated} {2D} {Images} into {Accurate} {3D}, 2024{\natexlab{b}}.

\bibitem[Chen et~al.(2024{\natexlab{c}})Chen, Xie, Zong, Li, Gao, Yang, Wu, and Jiang]{chen_atlas3d_2024}
Yunuo Chen, Tianyi Xie, Zeshun Zong, Xuan Li, Feng Gao, Yin Yang, Ying~Nian Wu, and Chenfanfu Jiang.
\newblock {Atlas3D}: {Physically} {Constrained} {Self}-{Supporting} {Text}-to-{3D} for {Simulation} and {Fabrication}.
\newblock In \emph{Advances in Neural Information Processing Systems 37 (NeurIPS 2024)}, 2024{\natexlab{c}}.

\bibitem[Chen(2023)]{chen_review_2023}
Zhiqin Chen.
\newblock A {Review} of {Deep} {Learning}-{Powered} {Mesh} {Reconstruction} {Methods}, 2023.

\bibitem[Chen et~al.(2024{\natexlab{d}})Chen, Walsman, Memmel, Mo, Fang, Vemuri, Wu, Fox, and Gupta]{chen_urdformer_2024}
Zoey Chen, Aaron Walsman, Marius Memmel, Kaichun Mo, Alex Fang, Karthikeya Vemuri, Alan Wu, Dieter Fox, and Abhishek Gupta.
\newblock {URDFormer}: {A} {Pipeline} for {Constructing} {Articulated} {Simulation} {Environments} from {Real}-{World} {Images}, 2024{\natexlab{d}}.

\bibitem[Choy et~al.(2016)Choy, Xu, Gwak, Chen, and Savarese]{choy_3d-r2n2_2016}
Christopher~B. Choy, Danfei Xu, JunYoung Gwak, Kevin Chen, and Silvio Savarese.
\newblock {3D}-{R2N2}: {A} {Unified} {Approach} for {Single} and {Multi}-view {3D} {Object} {Reconstruction}.
\newblock In \emph{Computer {Vision} – {ECCV} 2016}, 2016.

\bibitem[Chu et~al.(2023)Chu, Zhang, Liu, and Wang]{chu_buol_2023}
Tao Chu, Pan Zhang, Qiong Liu, and Jiaqi Wang.
\newblock {BUOL}: {A} {Bottom}-{Up} {Framework} {With} {Occupancy}-{Aware} {Lifting} for {Panoptic} {3D} {Scene} {Reconstruction} {From} a {Single} {Image}.
\newblock In \emph{Proceedings of the IEEE/CVF Conference on Computer Vision and Pattern Recognition (CVPR)}, 2023.

\bibitem[Coumans and Bai(2016)]{coumans_pybullet_2016}
Erwin Coumans and Yunfei Bai.
\newblock Pybullet, a python module for physics simulation for games, robotics and machine learning, 2016.

\bibitem[Dai et~al.(2017)Dai, Chang, Savva, Halber, Funkhouser, and Niessner]{dai_scannet_2017}
Angela Dai, Angel~X. Chang, Manolis Savva, Maciej Halber, Thomas Funkhouser, and Matthias Niessner.
\newblock {ScanNet}: {Richly}-{Annotated} {3D} {Reconstructions} of {Indoor} {Scenes}.
\newblock In \emph{Proceedings of the IEEE Conference on Computer Vision and Pattern Recognition (CVPR)}, 2017.

\bibitem[Dalal et~al.(2024)Dalal, Hagen, Robbersmyr, and Knausgård]{dalal_gaussian_2024}
Anurag Dalal, Daniel Hagen, Kjell~G. Robbersmyr, and Kristian~Muri Knausgård.
\newblock Gaussian {Splatting}: {3D} {Reconstruction} and {Novel} {View} {Synthesis}: {A} {Review}.
\newblock \emph{IEEE Access}, 2024.

\bibitem[Deitke et~al.(2022)Deitke, VanderBilt, Herrasti, Weihs, Ehsani, Salvador, Han, Kolve, Kembhavi, and Mottaghi]{deitke_procthor_2022}
Matt Deitke, Eli VanderBilt, Alvaro Herrasti, Luca Weihs, Kiana Ehsani, Jordi Salvador, Winson Han, Eric Kolve, Aniruddha Kembhavi, and Roozbeh Mottaghi.
\newblock {ProcTHOR}: {Large}-{Scale} {Embodied} {AI} {Using} {Procedural} {Generation}.
\newblock \emph{Advances in Neural Information Processing Systems}, 2022.

\bibitem[Deitke et~al.(2023)Deitke, Liu, Wallingford, Ngo, Michel, Kusupati, Fan, Laforte, Voleti, Gadre, VanderBilt, Kembhavi, Vondrick, Gkioxari, Ehsani, Schmidt, and Farhadi]{deitke_objaverse-xl_2023}
Matt Deitke, Ruoshi Liu, Matthew Wallingford, Huong Ngo, Oscar Michel, Aditya Kusupati, Alan Fan, Christian Laforte, Vikram Voleti, Samir~Yitzhak Gadre, Eli VanderBilt, Aniruddha Kembhavi, Carl Vondrick, Georgia Gkioxari, Kiana Ehsani, Ludwig Schmidt, and Ali Farhadi.
\newblock Objaverse-{XL}: {A} {Universe} of {10M}+ {3D} {Objects}.
\newblock In \emph{Advances in {Neural} {Information} {Processing} {Systems}}, 2023.

\bibitem[Deng et~al.(2020)Deng, Xiang, Mousavian, Eppner, Bretl, and Fox]{deng_self-supervised_2020}
Xinke Deng, Yu~Xiang, Arsalan Mousavian, Clemens Eppner, Timothy Bretl, and Dieter Fox.
\newblock Self-supervised {6D} {Object} {Pose} {Estimation} for {Robot} {Manipulation}.
\newblock In \emph{2020 {IEEE} {International} {Conference} on {Robotics} and {Automation} ({ICRA})}, 2020.

\bibitem[Dogaru et~al.(2024)Dogaru, Özer, and Egger]{dogaru_generalizable_2024}
Andreea Dogaru, Mert Özer, and Bernhard Egger.
\newblock Generalizable {3D} {Scene} {Reconstruction} via {Divide} and {Conquer} from a {Single} {View}, 2024.

\bibitem[Downs et~al.(2022)Downs, Francis, Koenig, Kinman, Hickman, Reymann, McHugh, and Vanhoucke]{downs_google_2022}
Laura Downs, Anthony Francis, Nate Koenig, Brandon Kinman, Ryan Hickman, Krista Reymann, Thomas~B. McHugh, and Vincent Vanhoucke.
\newblock Google {Scanned} {Objects}: {A} {High}-{Quality} {Dataset} of {3D} {Scanned} {Household} {Items}.
\newblock In \emph{2022 {International} {Conference} on {Robotics} and {Automation} ({ICRA})}, 2022.

\bibitem[Fan et~al.(2017)Fan, Su, and Guibas]{fan_point_2017}
Haoqiang Fan, Hao Su, and Leonidas~J. Guibas.
\newblock A {Point} {Set} {Generation} {Network} for {3D} {Object} {Reconstruction} {From} a {Single} {Image}.
\newblock In \emph{Proceedings of the IEEE Conference on Computer Vision and Pattern Recognition (CVPR)}, 2017.

\bibitem[Fei et~al.(2024)Fei, Xu, Zhang, Zhou, Yang, and He]{fei_3d_2024}
Ben Fei, Jingyi Xu, Rui Zhang, Qingyuan Zhou, Weidong Yang, and Ying He.
\newblock {3D} {Gaussian} as a {New} {Era}: {A} {Survey}.
\newblock \emph{IEEE Transactions on Visualization and Computer Graphics}, 2024.

\bibitem[Feng et~al.(2024)Feng, Shang, Li, Shao, Jiang, and Yang]{feng_pie-nerf_2024}
Yutao Feng, Yintong Shang, Xuan Li, Tianjia Shao, Chenfanfu Jiang, and Yin Yang.
\newblock {PIE}-{NeRF}: {Physics}-based {Interactive} {Elastodynamics} with {NeRF}.
\newblock In \emph{Proceedings of the IEEE/CVF Conference on Computer Vision and Pattern Recognition (CVPR)}, 2024.

\bibitem[FISCHLER~AND(1981)]{fischler_and_random_1981}
MA~FISCHLER~AND.
\newblock Random sample consensus: a paradigm for model fitting with applications to image analysis and automated cartography.
\newblock \emph{Commun. ACM}, 1981.

\bibitem[Gao et~al.(2024)Gao, Liu, Chen, Geiger, and Schölkopf]{gao_graphdreamer_2024}
Gege Gao, Weiyang Liu, Anpei Chen, Andreas Geiger, and Bernhard Schölkopf.
\newblock {GraphDreamer}: {Compositional} {3D} {Scene} {Synthesis} from {Scene} {Graphs}.
\newblock In \emph{Proceedings of the IEEE/CVF Conference on Computer Vision and Pattern Recognition (CVPR)}, 2024.

\bibitem[Gilles et~al.(2022)Gilles, Chen, Robin~Winter, Zhixuan~Zeng, and Wong]{gilles_metagraspnet_2022}
Maximilian Gilles, Yuhao Chen, Tim Robin~Winter, E.~Zhixuan~Zeng, and Alexander Wong.
\newblock {MetaGraspNet}: {A} {Large}-{Scale} {Benchmark} {Dataset} for {Scene}-{Aware} {Ambidextrous} {Bin} {Picking} via {Physics}-based {Metaverse} {Synthesis}.
\newblock In \emph{2022 {IEEE} 18th {International} {Conference} on {Automation} {Science} and {Engineering} ({CASE})}, 2022.

\bibitem[Goodfellow et~al.(2014)Goodfellow, Pouget-Abadie, Mirza, Xu, Warde-Farley, Ozair, Courville, and Bengio]{goodfellow_generative_2014}
Ian~J. Goodfellow, Jean Pouget-Abadie, Mehdi Mirza, Bing Xu, David Warde-Farley, Sherjil Ozair, Aaron Courville, and Yoshua Bengio.
\newblock Generative {Adversarial} {Nets}.
\newblock In \emph{Advances in {Neural} {Information} {Processing} {Systems}}, 2014.

\bibitem[Groueix et~al.(2018)Groueix, Fisher, Kim, Russell, and Aubry]{groueix_papier-mache_2018}
Thibault Groueix, Matthew Fisher, Vladimir~G. Kim, Bryan~C. Russell, and Mathieu Aubry.
\newblock A {Papier}-{Mâché} {Approach} to {Learning} {3D} {Surface} {Generation}.
\newblock In \emph{Proceedings of the IEEE Conference on Computer Vision and Pattern Recognition (CVPR)}, 2018.

\bibitem[Guo et~al.(2024)Guo, Wang, Ma, Zhang, Owens, Gan, Tenenbaum, He, and Matusik]{guo_physically_2024}
Minghao Guo, Bohan Wang, Pingchuan Ma, Tianyuan Zhang, Crystal~E. Owens, Chuang Gan, Joshua~B. Tenenbaum, Kaiming He, and Wojciech Matusik.
\newblock Physically {Compatible} {3D} {Object} {Modeling} from a {Single} {Image}.
\newblock In \emph{Advances in {Neural} {Information} {Processing} {Systems}}, 2024.

\bibitem[Ha and Schmidhuber(2018)]{ha_world_2018}
David Ha and Jürgen Schmidhuber.
\newblock World models.
\newblock \emph{arXiv preprint arXiv:1803.10122}, 2\penalty0 (3), 2018.

\bibitem[Hafner et~al.(2019{\natexlab{a}})Hafner, Lillicrap, Ba, and Norouzi]{hafner_dream_2019}
Danijar Hafner, Timothy Lillicrap, Jimmy Ba, and Mohammad Norouzi.
\newblock Dream to {Control}: {Learning} {Behaviors} by {Latent} {Imagination}.
\newblock In \emph{International Conference on Learning Representations}, 2019{\natexlab{a}}.

\bibitem[Hafner et~al.(2019{\natexlab{b}})Hafner, Lillicrap, Fischer, Villegas, Ha, Lee, and Davidson]{hafner_learning_2019}
Danijar Hafner, Timothy Lillicrap, Ian Fischer, Ruben Villegas, David Ha, Honglak Lee, and James Davidson.
\newblock Learning {Latent} {Dynamics} for {Planning} from {Pixels}.
\newblock In \emph{Proceedings of the 36th {International} {Conference} on {Machine} {Learning}}. PMLR, 2019{\natexlab{b}}.

\bibitem[Han et~al.(2024)Han, Yang, Liao, Xing, Xu, Yu, Zha, Li, and Li]{han_reparo_2024}
Haonan Han, Rui Yang, Huan Liao, Jiankai Xing, Zunnan Xu, Xiaoming Yu, Junwei Zha, Xiu Li, and Wanhua Li.
\newblock {REPARO}: {Compositional} {3D} {Assets} {Generation} with {Differentiable} {3D} {Layout} {Alignment}, 2024.

\bibitem[Han et~al.(2021)Han, Laga, and Bennamoun]{han_image-based_2021}
Xian-Feng Han, Hamid Laga, and Mohammed Bennamoun.
\newblock Image-{Based} {3D} {Object} {Reconstruction}: {State}-of-the-{Art} and {Trends} in the {Deep} {Learning} {Era}.
\newblock \emph{IEEE Transactions on Pattern Analysis and Machine Intelligence}, 2021.

\bibitem[Han et~al.(2025)Han, Liu, Chen, Yu, Lyu, Tian, Wang, Zhang, and Pang]{han_re3sim_2025}
Xiaoshen Han, Minghuan Liu, Yilun Chen, Junqiu Yu, Xiaoyang Lyu, Yang Tian, Bolun Wang, Weinan Zhang, and Jiangmiao Pang.
\newblock Re\${\textasciicircum}3\${Sim}: {Generating} {High}-{Fidelity} {Simulation} {Data} via {3D}-{Photorealistic} {Real}-to-{Sim} for {Robotic} {Manipulation}, 2025.

\bibitem[Hansen et~al.(2022)Hansen, Su, and Wang]{hansen_temporal_2022}
Nicklas~A. Hansen, Hao Su, and Xiaolong Wang.
\newblock Temporal {Difference} {Learning} for {Model} {Predictive} {Control}.
\newblock In \emph{Proceedings of the 39th {International} {Conference} on {Machine} {Learning}}, 2022.

\bibitem[Hassena et~al.(2024)Hassena, Moon, Fujii, Yuen, Snavely, Marschner, and Hariharan]{hassena_objectcarver_2024}
Gemmechu Hassena, Jonathan Moon, Ryan Fujii, Andrew Yuen, Noah Snavely, Steve Marschner, and Bharath Hariharan.
\newblock {ObjectCarver}: {Semi}-automatic segmentation, reconstruction and separation of {3D} objects, 2024.

\bibitem[He et~al.(2016)He, Zhang, Ren, and Sun]{he_deep_2016}
Kaiming He, Xiangyu Zhang, Shaoqing Ren, and Jian Sun.
\newblock Deep {Residual} {Learning} for {Image} {Recognition}.
\newblock In \emph{Proceedings of the IEEE Conference on Computer Vision and Pattern Recognition (CVPR)}, 2016.

\bibitem[Hong et~al.(2022)Hong, Zhang, Pan, Cai, Yang, and Liu]{hong_avatarclip_2022}
Fangzhou Hong, Mingyuan Zhang, Liang Pan, Zhongang Cai, Lei Yang, and Ziwei Liu.
\newblock {AvatarCLIP}: zero-shot text-driven generation and animation of {3D} avatars.
\newblock \emph{ACM Trans. Graph.}, 2022.

\bibitem[Hong et~al.(2023)Hong, Zhang, Gu, Bi, Zhou, Liu, Liu, Sunkavalli, Bui, and Tan]{hong_lrm_2023}
Yicong Hong, Kai Zhang, Jiuxiang Gu, Sai Bi, Yang Zhou, Difan Liu, Feng Liu, Kalyan Sunkavalli, Trung Bui, and Hao Tan.
\newblock {LRM}: {Large} {Reconstruction} {Model} for {Single} {Image} to {3D}.
\newblock In \emph{International Conference on Learning Representations}, 2023.

\bibitem[Huang et~al.(2024{\natexlab{a}})Huang, Yu, Chen, Geiger, and Gao]{huang_2d_2024}
Binbin Huang, Zehao Yu, Anpei Chen, Andreas Geiger, and Shenghua Gao.
\newblock {2D} {Gaussian} {Splatting} for {Geometrically} {Accurate} {Radiance} {Fields}.
\newblock In \emph{{ACM} {SIGGRAPH} 2024 {Conference} {Papers}}, {SIGGRAPH} '24, 2024{\natexlab{a}}.

\bibitem[Huang et~al.(2024{\natexlab{b}})Huang, Guo, An, Yang, Li, Zou, Liang, Liu, Cao, and Sheng]{huang_midi_2024}
Zehuan Huang, Yuan-Chen Guo, Xingqiao An, Yunhan Yang, Yangguang Li, Zi-Xin Zou, Ding Liang, Xihui Liu, Yan-Pei Cao, and Lu~Sheng.
\newblock {MIDI}: {Multi}-{Instance} {Diffusion} for {Single} {Image} to {3D} {Scene} {Generation}, 2024{\natexlab{b}}.

\bibitem[Huang et~al.(2024{\natexlab{c}})Huang, Wen, Wang, Ren, and Jia]{huang_surface_2024}
ZhangJin Huang, Yuxin Wen, ZiHao Wang, Jinjuan Ren, and Kui Jia.
\newblock Surface {Reconstruction} {From} {Point} {Clouds}: {A} {Survey} and a {Benchmark}.
\newblock \emph{IEEE Transactions on Pattern Analysis and Machine Intelligence}, 2024{\natexlab{c}}.

\bibitem[Huang et~al.(2024{\natexlab{d}})Huang, Stojanov, Thai, Jampani, and Rehg]{huang_zeroshape_2024}
Zixuan Huang, Stefan Stojanov, Anh Thai, Varun Jampani, and James~M. Rehg.
\newblock {ZeroShape}: {Regression}-based {Zero}-shot {Shape} {Reconstruction}.
\newblock In \emph{Proceedings of the IEEE/CVF Conference on Computer Vision and Pattern Recognition (CVPR)}, 2024{\natexlab{d}}.

\bibitem[Irshad et~al.(2024)Irshad, Comi, Lin, Heppert, Valada, Ambrus, Kira, and Tremblay]{irshad_neural_2024}
Muhammad~Zubair Irshad, Mauro Comi, Yen-Chen Lin, Nick Heppert, Abhinav Valada, Rares Ambrus, Zsolt Kira, and Jonathan Tremblay.
\newblock Neural {Fields} in {Robotics}: {A} {Survey}, 2024.

\bibitem[Jack et~al.(2019)Jack, Pontes, Sridharan, Fookes, Shirazi, Maire, and Eriksson]{jack_learning_2019}
Dominic Jack, Jhony~K. Pontes, Sridha Sridharan, Clinton Fookes, Sareh Shirazi, Frederic Maire, and Anders Eriksson.
\newblock Learning {Free}-{Form} {Deformations} for {3D} {Object} {Reconstruction}.
\newblock In \emph{Computer {Vision} – {ACCV} 2018}, 2019.

\bibitem[Jain et~al.(2022)Jain, Mildenhall, Barron, Abbeel, and Poole]{jain_zero-shot_2022}
Ajay Jain, Ben Mildenhall, Jonathan~T. Barron, Pieter Abbeel, and Ben Poole.
\newblock Zero-{Shot} {Text}-{Guided} {Object} {Generation} {With} {Dream} {Fields}.
\newblock In \emph{Proceedings of the IEEE/CVF Conference on Computer Vision and Pattern Recognition (CVPR)}, 2022.

\bibitem[Jaunet et~al.(2021)Jaunet, Bono, Vuillemot, and Wolf]{jaunet_sim2realviz_2021}
Theo Jaunet, Guillaume Bono, Romain Vuillemot, and Christian Wolf.
\newblock {SIM2REALVIZ}: {Visualizing} the {Sim2Real} {Gap} in {Robot} {Ego}-{Pose} {Estimation}, 2021.

\bibitem[Jia et~al.(2024)Jia, Wang, Dong, Wu, Zeng, Ge, Ding, Yan, Gu, Li, Wang, Cheng, Sui, Huang, and Zhou]{jia_discoverse_2024}
Yufei Jia, Guangyu Wang, Yuhang Dong, Junzhe Wu, Yupei Zeng, Haizhou Ge, Kairui Ding, Zike Yan, Weibin Gu, Chuxuan Li, Ziming Wang, Yunjie Cheng, Wei Sui, Ruqi Huang, and Guyue Zhou.
\newblock {DISCOVERSE}: {Efficient} {Robot} {Simulation} in {Complex} {High}-{Fidelity} {Environments}, 2024.

\bibitem[Jiang et~al.(2024{\natexlab{a}})Jiang, Huang, and Pavlakos]{jiang_real3d_2024}
Hanwen Jiang, Qixing Huang, and Georgios Pavlakos.
\newblock {Real3D}: {Scaling} {Up} {Large} {Reconstruction} {Models} with {Real}-{World} {Images}, 2024{\natexlab{a}}.

\bibitem[Jiang et~al.(2024{\natexlab{b}})Jiang, Yu, Xie, Li, Feng, Wang, Li, Lau, Gao, Yang, and Jiang]{jiang_vr-gs_2024}
Ying Jiang, Chang Yu, Tianyi Xie, Xuan Li, Yutao Feng, Huamin Wang, Minchen Li, Henry Lau, Feng Gao, Yin Yang, and Chenfanfu Jiang.
\newblock {VR}-{GS}: {A} {Physical} {Dynamics}-{Aware} {Interactive} {Gaussian} {Splatting} {System} in {Virtual} {Reality}.
\newblock In \emph{{ACM} {SIGGRAPH} 2024 {Conference} {Papers}}, {SIGGRAPH} '24, 2024{\natexlab{b}}.

\bibitem[Jun and Nichol(2023)]{jun_shap-e_2023}
Heewoo Jun and Alex Nichol.
\newblock Shap-{E}: {Generating} {Conditional} {3D} {Implicit} {Functions}, 2023.

\bibitem[Kaiser et~al.(2019)Kaiser, Babaeizadeh, Miłos, Osiński, Campbell, Czechowski, Erhan, Finn, Kozakowski, Levine, Mohiuddin, Sepassi, Tucker, and Michalewski]{kaiser_model_2019}
Łukasz Kaiser, Mohammad Babaeizadeh, Piotr Miłos, Błażej Osiński, Roy~H. Campbell, Konrad Czechowski, Dumitru Erhan, Chelsea Finn, Piotr Kozakowski, Sergey Levine, Afroz Mohiuddin, Ryan Sepassi, George Tucker, and Henryk Michalewski.
\newblock Model {Based} {Reinforcement} {Learning} for {Atari}.
\newblock In \emph{International Conference on Learning Representations}, 2019.

\bibitem[Kantarci et~al.(2022)Kantarci, Gökberk, and Akarun]{kantarci_survey_2022}
Merve~Gül Kantarci, Berk Gökberk, and Lale Akarun.
\newblock A {Survey} of {3D} {Object} {Reconstruction} {Methods}.
\newblock In \emph{2022 30th {Signal} {Processing} and {Communications} {Applications} {Conference} ({SIU})}, 2022.

\bibitem[Kar et~al.(2017)Kar, Häne, and Malik]{kar_learning_2017}
Abhishek Kar, Christian Häne, and Jitendra Malik.
\newblock Learning a {Multi}-{View} {Stereo} {Machine}.
\newblock In \emph{Advances in {Neural} {Information} {Processing} {Systems}}, 2017.

\bibitem[Katara et~al.(2024)Katara, Xian, and Fragkiadaki]{katara_gen2sim_2024}
Pushkal Katara, Zhou Xian, and Katerina Fragkiadaki.
\newblock {Gen2Sim}: {Scaling} up {Robot} {Learning} in {Simulation} with {Generative} {Models}.
\newblock In \emph{2024 {IEEE} {International} {Conference} on {Robotics} and {Automation} ({ICRA})}, 2024.

\bibitem[Keetha et~al.(2024)Keetha, Karhade, Jatavallabhula, Yang, Scherer, Ramanan, and Luiten]{keetha_splatam_2024}
Nikhil Keetha, Jay Karhade, Krishna~Murthy Jatavallabhula, Gengshan Yang, Sebastian Scherer, Deva Ramanan, and Jonathon Luiten.
\newblock {SplaTAM}: {Splat} {Track} \& {Map} {3D} {Gaussians} for {Dense} {RGB}-{D} {SLAM}.
\newblock In \emph{Proceedings of the IEEE/CVF Conference on Computer Vision and Pattern Recognition (CVPR)}, 2024.

\bibitem[Kent et~al.(2020)Kent, Saldanha, and Chernova]{kent_leveraging_2020}
David Kent, Carl Saldanha, and Sonia Chernova.
\newblock Leveraging depth data in remote robot teleoperation interfaces for general object manipulation.
\newblock \emph{The International Journal of Robotics Research}, 2020.

\bibitem[Kerbl et~al.(2023)Kerbl, Kopanas, Leimkuehler, and Drettakis]{kerbl_3d_2023}
Bernhard Kerbl, Georgios Kopanas, Thomas Leimkuehler, and George Drettakis.
\newblock {3D} {Gaussian} {Splatting} for {Real}-{Time} {Radiance} {Field} {Rendering}.
\newblock \emph{ACM Trans. Graph.}, 2023.

\bibitem[Khalid et~al.(2022)Khalid, Xie, Belilovsky, and Popa]{khalid_clip-mesh_2022}
Nasir~Mohammad Khalid, Tianhao Xie, Eugene Belilovsky, and Tiberiu Popa.
\newblock {CLIP}-{Mesh}: {Generating} textured meshes from text using pretrained image-text models.
\newblock In \emph{{SIGGRAPH} {Asia} 2022 {Conference} {Papers}}, 2022.

\bibitem[Khan et~al.(2022)Khan, Nazir, Pagani, Mokayed, Liwicki, Stricker, and Afzal]{khan_comprehensive_2022}
Muhammad Ahmed~Ullah Khan, Danish Nazir, Alain Pagani, Hamam Mokayed, Marcus Liwicki, Didier Stricker, and Muhammad~Zeshan Afzal.
\newblock A comprehensive survey of depth completion approaches.
\newblock \emph{Sensors}, 2022.

\bibitem[Khanna et~al.(2024)Khanna, Mao, Jiang, Haresh, Shacklett, Batra, Clegg, Undersander, Chang, and Savva]{khanna_habitat_2024}
Mukul Khanna, Yongsen Mao, Hanxiao Jiang, Sanjay Haresh, Brennan Shacklett, Dhruv Batra, Alexander Clegg, Eric Undersander, Angel~X. Chang, and Manolis Savva.
\newblock Habitat {Synthetic} {Scenes} {Dataset} ({HSSD}-200): {An} {Analysis} of {3D} {Scene} {Scale} and {Realism} {Tradeoffs} for {ObjectGoal} {Navigation}.
\newblock In \emph{Proceedings of the IEEE/CVF Conference on Computer Vision and Pattern Recognition (CVPR)}, 2024.

\bibitem[Kim et~al.(2025)Kim, Park, and Cho]{kim_complete_2025}
Wongyeom Kim, Jisun Park, and Kyungeun Cho.
\newblock Complete {Object}-{Compositional} {Neural} {Implicit} {Surfaces} {With} {3D} {Pseudo} {Supervision}.
\newblock \emph{IEEE Access}, 2025.

\bibitem[Kingma and Welling(2014)]{kingma_auto-encoding_2014}
Diederik~P. Kingma and Max Welling.
\newblock Auto-{Encoding} {Variational} {Bayes}.
\newblock In \emph{Proceedings of the {International} {Conference} on {Learning} {Representations}}, 2014.

\bibitem[Kolve et~al.(2022)Kolve, Mottaghi, Han, VanderBilt, Weihs, Herrasti, Deitke, Ehsani, Gordon, Zhu, Kembhavi, Gupta, and Farhadi]{kolve_ai2-thor_2022}
Eric Kolve, Roozbeh Mottaghi, Winson Han, Eli VanderBilt, Luca Weihs, Alvaro Herrasti, Matt Deitke, Kiana Ehsani, Daniel Gordon, Yuke Zhu, Aniruddha Kembhavi, Abhinav Gupta, and Ali Farhadi.
\newblock {AI2}-{THOR}: {An} {Interactive} {3D} {Environment} for {Visual} {AI}, 2022.

\bibitem[Kong et~al.(2023)Kong, Liu, Taher, and Davison]{kong_vmap_2023}
Xin Kong, Shikun Liu, Marwan Taher, and Andrew~J. Davison.
\newblock {vMAP}: {Vectorised} {Object} {Mapping} for {Neural} {Field} {SLAM}.
\newblock In \emph{Proceedings of the IEEE/CVF Conference on Computer Vision and Pattern Recognition (CVPR)}, 2023.

\bibitem[Le et~al.(2024)Le, Xie, Liang, Wang, Yang, Ma, Vedder, Krishna, Jayaraman, and Eaton]{le_articulate-anything_2024}
Long Le, Jason Xie, William Liang, Hung-Ju Wang, Yue Yang, Yecheng~Jason Ma, Kyle Vedder, Arjun Krishna, Dinesh Jayaraman, and Eric Eaton.
\newblock Articulate-{Anything}: {Automatic} {Modeling} of {Articulated} {Objects} via a {Vision}-{Language} {Foundation} {Model}, 2024.

\bibitem[Lee and Chang(2022)]{lee_understanding_2022}
Han-Hung Lee and Angel~X. Chang.
\newblock Understanding {Pure} {CLIP} {Guidance} for {Voxel} {Grid} {NeRF} {Models}, 2022.

\bibitem[Lei et~al.(2024)Lei, Li, Xu, Li, and Gu]{lei_whats_2024}
Na~Lei, Zezeng Li, Zebin Xu, Ying Li, and Xianfeng Gu.
\newblock What's the {Situation} {With} {Intelligent} {Mesh} {Generation}: {A} {Survey} and {Perspectives}.
\newblock \emph{IEEE Transactions on Visualization and Computer Graphics}, 2024.

\bibitem[Lerher et~al.(2023)Lerher, Bencak, Hercog, Jerman, and Bizjak]{lerher_robotic_2023}
Tone Lerher, Primož Bencak, Darko Hercog, Boris Jerman, and Luka Bizjak.
\newblock Robotic bin-picking: {Benchmarking} robotics grippers with modified {YCB} object and model set.
\newblock \emph{Progress in Material Handling Research}, 2023.

\bibitem[Li et~al.(2023{\natexlab{a}})Li, Tan, Zhang, Xu, Luan, Xu, Hong, Sunkavalli, Shakhnarovich, and Bi]{li_instant3d_2023}
Jiahao Li, Hao Tan, Kai Zhang, Zexiang Xu, Fujun Luan, Yinghao Xu, Yicong Hong, Kalyan Sunkavalli, Greg Shakhnarovich, and Sai Bi.
\newblock {Instant3D}: {Fast} {Text}-to-{3D} with {Sparse}-view {Generation} and {Large} {Reconstruction} {Model}.
\newblock In \emph{International Conference on Learning Representations}, 2023{\natexlab{a}}.

\bibitem[Li et~al.(2024{\natexlab{a}})Li, Wang, Peng, Shen, Zhu, and Xu]{li_visual_2024}
Jinming Li, Wanying Wang, Yaxin Peng, Chaomin Shen, Yichen Zhu, and Zhiyuan Xu.
\newblock Visual {Robotic} {Manipulation} with {Depth}-{Aware} {Pretraining}.
\newblock In \emph{2024 {IEEE} {International} {Conference} on {Robotics} and {Biomimetics} ({ROBIO})}, 2024{\natexlab{a}}.

\bibitem[Li et~al.(2025{\natexlab{a}})Li, Zheng, Rupprecht, and Vedaldi]{li_dragapart_2025}
Ruining Li, Chuanxia Zheng, Christian Rupprecht, and Andrea Vedaldi.
\newblock {DragAPart}: {Learning} a {Part}-{Level} {Motion} {Prior} for {Articulated} {Objects}.
\newblock In \emph{Computer {Vision} – {ECCV} 2024}, 2025{\natexlab{a}}.

\bibitem[Li et~al.(2025{\natexlab{b}})Li, Zheng, Rupprecht, and Vedaldi]{li_dso_2025}
Ruining Li, Chuanxia Zheng, Christian Rupprecht, and Andrea Vedaldi.
\newblock {DSO}: {Aligning} {3D} {Generators} with {Simulation} {Feedback} for {Physical} {Soundness}, 2025{\natexlab{b}}.

\bibitem[Li et~al.(2024{\natexlab{b}})Li, Li, Zhang, Zhang, Jia, Wang, Fan, Tseng, and Wang]{li_robogsim_2024}
Xinhai Li, Jialin Li, Ziheng Zhang, Rui Zhang, Fan Jia, Tiancai Wang, Haoqiang Fan, Kuo-Kun Tseng, and Ruiping Wang.
\newblock {RoboGSim}: {A} {Real2Sim2Real} {Robotic} {Gaussian} {Splatting} {Simulator}, 2024{\natexlab{b}}.

\bibitem[Li et~al.(2023{\natexlab{b}})Li, Lyu, Ding, Wang, Liao, and Liu]{li_rico_2023}
Zizhang Li, Xiaoyang Lyu, Yuanyuan Ding, Mengmeng Wang, Yiyi Liao, and Yong Liu.
\newblock {RICO}: {Regularizing} the {Unobservable} for {Indoor} {Compositional} {Reconstruction}.
\newblock In \emph{Proceedings of the IEEE/CVF International Conference on Computer Vision (ICCV)}, 2023{\natexlab{b}}.

\bibitem[Lim and Pham(2023)]{lim_grasping_2023}
Joyce Xin-Yan Lim and Quang-Cuong Pham.
\newblock Grasping, {Part} {Identification}, and {Pose} {Refinement} in {One} {Shot} with a {Tactile} {Gripper}, 2023.

\bibitem[Liu et~al.(2024{\natexlab{a}})Liu, Wang, Qiu, Lin, Zhou, and Su]{liu_gaussian_2024}
Liu Liu, Xinjie Wang, Jiaxiong Qiu, Tianwei Lin, Xiaolin Zhou, and Zhizhong Su.
\newblock Gaussian {Object} {Carver}: {Object}-{Compositional} {Gaussian} {Splatting} with surfaces completion, 2024{\natexlab{a}}.

\bibitem[Liu et~al.(2023{\natexlab{a}})Liu, Xu, Jin, Chen, Varma~T, Xu, and Su]{liu_one-2-3-45_2023}
Minghua Liu, Chao Xu, Haian Jin, Linghao Chen, Mukund Varma~T, Zexiang Xu, and Hao Su.
\newblock One-2-3-45: {Any} {Single} {Image} to {3D} {Mesh} in 45 {Seconds} without {Per}-{Shape} {Optimization}.
\newblock \emph{Advances in Neural Information Processing Systems}, 2023{\natexlab{a}}.

\bibitem[Liu et~al.(2024{\natexlab{b}})Liu, Shi, Chen, Zhang, Xu, Wei, Chen, Zeng, Gu, and Su]{liu_one-2-3-45_2024}
Minghua Liu, Ruoxi Shi, Linghao Chen, Zhuoyang Zhang, Chao Xu, Xinyue Wei, Hansheng Chen, Chong Zeng, Jiayuan Gu, and Hao Su.
\newblock One-2-3-45++: {Fast} {Single} {Image} to {3D} {Objects} with {Consistent} {Multi}-{View} {Generation} and {3D} {Diffusion}.
\newblock In \emph{Proceedings of the IEEE/CVF Conference on Computer Vision and Pattern Recognition (CVPR)}, 2024{\natexlab{b}}.

\bibitem[Liu et~al.(2023{\natexlab{b}})Liu, Wu, Van~Hoorick, Tokmakov, Zakharov, and Vondrick]{liu_zero-1--3_2023}
Ruoshi Liu, Rundi Wu, Basile Van~Hoorick, Pavel Tokmakov, Sergey Zakharov, and Carl Vondrick.
\newblock Zero-1-to-3: {Zero}-shot {One} {Image} to {3D} {Object}.
\newblock In \emph{Proceedings of the IEEE/CVF International Conference on Computer Vision (ICCV)}, 2023{\natexlab{b}}.

\bibitem[Liu et~al.(2023{\natexlab{c}})Liu, Lin, Zeng, Long, Liu, Komura, and Wang]{liu_syncdreamer_2023}
Yuan Liu, Cheng Lin, Zijiao Zeng, Xiaoxiao Long, Lingjie Liu, Taku Komura, and Wenping Wang.
\newblock {SyncDreamer}: {Generating} {Multiview}-consistent {Images} from a {Single}-view {Image}.
\newblock In \emph{International Conference on Learning Representations}, 2023{\natexlab{c}}.

\bibitem[Long et~al.(2024)Long, Guo, Lin, Liu, Dou, Liu, Ma, Zhang, Habermann, Theobalt, and Wang]{long_wonder3d_2024}
Xiaoxiao Long, Yuan-Chen Guo, Cheng Lin, Yuan Liu, Zhiyang Dou, Lingjie Liu, Yuexin Ma, Song-Hai Zhang, Marc Habermann, Christian Theobalt, and Wenping Wang.
\newblock {Wonder3D}: {Single} {Image} to {3D} using {Cross}-{Domain} {Diffusion}.
\newblock In \emph{Proceedings of the IEEE/CVF Conference on Computer Vision and Pattern Recognition (CVPR)}, 2024.

\bibitem[Lou et~al.(2024)Lou, Liu, Pan, Geng, Chen, Ma, Li, Wang, Feng, Shi, Luo, and Shi]{lou_robo-gs_2024}
Haozhe Lou, Yurong Liu, Yike Pan, Yiran Geng, Jianteng Chen, Wenlong Ma, Chenglong Li, Lin Wang, Hengzhen Feng, Lu~Shi, Liyi Luo, and Yongliang Shi.
\newblock Robo-{GS}: {A} {Physics} {Consistent} {Spatial}-{Temporal} {Model} for {Robotic} {Arm} with {Hybrid} {Representation}, 2024.

\bibitem[Lu et~al.(2025)Lu, Zhang, Wang, Liu, Lu, and Tang]{lu_manigaussian_2025}
Guanxing Lu, Shiyi Zhang, Ziwei Wang, Changliu Liu, Jiwen Lu, and Yansong Tang.
\newblock {ManiGaussian}: {Dynamic} {Gaussian} {Splatting} for {Multi}-task {Robotic} {Manipulation}.
\newblock In \emph{Computer {Vision} – {ECCV} 2024}, 2025.

\bibitem[Lu et~al.(2024)Lu, Gao, Dai, Zha, Hou, Wu, and Xia]{lu_large_2024}
Longfei Lu, Huachen Gao, Tao Dai, Yaohua Zha, Zhi Hou, Junta Wu, and Shu-Tao Xia.
\newblock Large {Point}-to-{Gaussian} {Model} for {Image}-to-{3D} {Generation}.
\newblock In \emph{Proceedings of the 32nd {ACM} {International} {Conference} on {Multimedia}}, {MM} '24, 2024.

\bibitem[Lu et~al.(2021)Lu, Baron, Clark, and Rojas]{lu_systematic_2021}
Qiujie Lu, Nicholas Baron, Angus~B. Clark, and Nicolas Rojas.
\newblock Systematic object-invariant in-hand manipulation via reconfigurable underactuation: {Introducing} the {RUTH} gripper.
\newblock \emph{The International Journal of Robotics Research}, 2021.

\bibitem[Makoviychuk et~al.(2021)Makoviychuk, Wawrzyniak, Guo, Lu, Storey, Macklin, Hoeller, Rudin, Allshire, Handa, and State]{makoviychuk_isaac_2021}
Viktor Makoviychuk, Lukasz Wawrzyniak, Yunrong Guo, Michelle Lu, Kier Storey, Miles Macklin, David Hoeller, Nikita Rudin, Arthur Allshire, Ankur Handa, and Gavriel State.
\newblock Isaac {Gym}: {High} {Performance} {GPU} {Based} {Physics} {Simulation} {For} {Robot} {Learning}.
\newblock In \emph{Thirty-fifth Conference on Neural Information Processing Systems Datasets and Benchmarks Track}, 2021.

\bibitem[Matsuki et~al.(2024)Matsuki, Murai, Kelly, and Davison]{matsuki_gaussian_2024}
Hidenobu Matsuki, Riku Murai, Paul H.~J. Kelly, and Andrew~J. Davison.
\newblock Gaussian {Splatting} {SLAM}.
\newblock In \emph{Proceedings of the IEEE/CVF Conference on Computer Vision and Pattern Recognition (CVPR)}, 2024.

\bibitem[Maxim and Nedevschi(2021)]{maxim_survey_2021}
Bogdan Maxim and Sergiu Nedevschi.
\newblock A survey on the current state of the art on deep learning {3D} reconstruction.
\newblock In \emph{2021 {IEEE} 17th {International} {Conference} on {Intelligent} {Computer} {Communication} and {Processing} ({ICCP})}, 2021.

\bibitem[Melas-Kyriazi et~al.(2024)Melas-Kyriazi, Laina, Rupprecht, Neverova, Vedaldi, Gafni, and Kokkinos]{melas-kyriazi_im-3d_2024}
Luke Melas-Kyriazi, Iro Laina, Christian Rupprecht, Natalia Neverova, Andrea Vedaldi, Oran Gafni, and Filippos Kokkinos.
\newblock {IM}-{3D}: {Iterative} {Multiview} {Diffusion} and {Reconstruction} for {High}-{Quality} {3D} {Generation}.
\newblock In \emph{Proceedings of the 41st {International} {Conference} on {Machine} {Learning}}, 2024.

\bibitem[Meng et~al.(2017)Meng, Qin, Chen, Chen, Sun, Lin, and Ang]{meng_two-stage_2017}
Zehui Meng, Hailong Qin, Ziyue Chen, Xudong Chen, Hao Sun, Feng Lin, and Marcelo~H. Ang.
\newblock A {Two}-{Stage} {Optimized} {Next}-{View} {Planning} {Framework} for 3-{D} {Unknown} {Environment} {Exploration}, and {Structural} {Reconstruction}.
\newblock \emph{IEEE Robotics and Automation Letters}, 2017.

\bibitem[Mescheder et~al.(2019)Mescheder, Oechsle, Niemeyer, Nowozin, and Geiger]{mescheder_occupancy_2019}
Lars Mescheder, Michael Oechsle, Michael Niemeyer, Sebastian Nowozin, and Andreas Geiger.
\newblock Occupancy {Networks}: {Learning} {3D} {Reconstruction} in {Function} {Space}.
\newblock In \emph{Proceedings of the IEEE/CVF Conference on Computer Vision and Pattern Recognition (CVPR)}, 2019.

\bibitem[Mezghanni et~al.(2021)Mezghanni, Boulkenafed, Lieutier, and Ovsjanikov]{mezghanni_physically-aware_2021}
Mariem Mezghanni, Malika Boulkenafed, André Lieutier, and Maks Ovsjanikov.
\newblock Physically-aware {Generative} {Network} for {3D} {Shape} {Modeling}.
\newblock In \emph{2021 {IEEE}/{CVF} {Conference} on {Computer} {Vision} and {Pattern} {Recognition} ({CVPR})}, 2021.

\bibitem[Mezghanni et~al.(2022)Mezghanni, Bodrito, Boulkenafed, and Ovsjanikov]{mezghanni_physical_2022}
Mariem Mezghanni, Théo Bodrito, Malika Boulkenafed, and Maks Ovsjanikov.
\newblock Physical {Simulation} {Layer} for {Accurate} {3D} {Modeling}.
\newblock In \emph{2022 {IEEE}/{CVF} {Conference} on {Computer} {Vision} and {Pattern} {Recognition} ({CVPR})}, 2022.

\bibitem[Miao et~al.(2023)Miao, Wang, and Bekris]{miao_safe_2023}
Yinglong Miao, Rui Wang, and Kostas Bekris.
\newblock Safe, {Occlusion}-{Aware} {Manipulation} for {Online} {Object} {Reconstruction} in {Confined} {Spaces}.
\newblock In \emph{Robotics {Research}}, 2023.

\bibitem[Mildenhall et~al.(2022)Mildenhall, Srinivasan, Tancik, Barron, Ramamoorthi, and Ng]{mildenhall_nerf_2022}
Ben Mildenhall, Pratul~P. Srinivasan, Matthew Tancik, Jonathan~T. Barron, Ravi Ramamoorthi, and Ren Ng.
\newblock {NeRF}: representing scenes as neural radiance fields for view synthesis.
\newblock \emph{Communications of the ACM}, 2022.

\bibitem[Mittal et~al.(2023)Mittal, Yu, Yu, Liu, Rudin, Hoeller, Yuan, Singh, Guo, Mazhar, Mandlekar, Babich, State, Hutter, and Garg]{mittal_orbit_2023}
Mayank Mittal, Calvin Yu, Qinxi Yu, Jingzhou Liu, Nikita Rudin, David Hoeller, Jia~Lin Yuan, Ritvik Singh, Yunrong Guo, Hammad Mazhar, Ajay Mandlekar, Buck Babich, Gavriel State, Marco Hutter, and Animesh Garg.
\newblock Orbit: {A} {Unified} {Simulation} {Framework} for {Interactive} {Robot} {Learning} {Environments}.
\newblock \emph{IEEE Robotics and Automation Letters}, 2023.

\bibitem[Moerland et~al.(2023)Moerland, Broekens, Plaat, and Jonker]{moerland_model-based_2023}
Thomas~M. Moerland, Joost Broekens, Aske Plaat, and Catholijn~M. Jonker.
\newblock Model-based {Reinforcement} {Learning}: {A} {Survey}.
\newblock \emph{Foundations and Trends in Machine Learning}, 2023.

\bibitem[Morgan et~al.(2021)Morgan, Wen, Liang, Boularias, Dollar, and Bekris]{morgan_vision-driven_2021}
Andrew~S Morgan, Bowen Wen, Junchi Liang, Abdeslam Boularias, Aaron~M Dollar, and Kostas Bekris.
\newblock Vision-driven {Compliant} {Manipulation} for {Reliable}, {High}-{Precision} {Assembly} {Tasks}.
\newblock In \emph{17th {Robotics}: {Science} and {Systems}, {RSS} 2021}, 2021.

\bibitem[Mu et~al.(2024)Mu, Chen, Peng, Chen, Gao, Zou, Lin, Xie, and Luo]{mu_robotwin_2024}
Yao Mu, Tianxing Chen, Shijia Peng, Zanxin Chen, Zeyu Gao, Yude Zou, Lunkai Lin, Zhiqiang Xie, and Ping Luo.
\newblock {RoboTwin}: {Dual}-{Arm} {Robot} {Benchmark} with {Generative} {Digital} {Twins} (early version), 2024.

\bibitem[Nguyen-Phuoc et~al.(2019)Nguyen-Phuoc, Li, Theis, Richardt, and Yang]{nguyen-phuoc_hologan_2019}
Thu Nguyen-Phuoc, Chuan Li, Lucas Theis, Christian Richardt, and Yong-Liang Yang.
\newblock {HoloGAN}: {Unsupervised} {Learning} of {3D} {Representations} {From} {Natural} {Images}.
\newblock In \emph{Proceedings of the IEEE/CVF International Conference on Computer Vision (ICCV)}, 2019.

\bibitem[Ni et~al.(2024)Ni, Chen, Jing, Jiang, Wang, Dai, Li, Zhu, Zhu, and Huang]{ni_phyrecon_2024}
Junfeng Ni, Yixin Chen, Bohan Jing, Nan Jiang, Bin Wang, Bo~Dai, Puhao Li, Yixin Zhu, Song-Chun Zhu, and Siyuan Huang.
\newblock {PhyRecon}: {Physically} {Plausible} {Neural} {Scene} {Reconstruction}.
\newblock In \emph{Advances in Neural Information Processing Systems 37}, 2024.

\bibitem[Niu et~al.(2021)Niu, Wang, and Liu]{niu_tolerance-guided_2021}
Boshen Niu, Chenxi Wang, and Changliu Liu.
\newblock Tolerance-{Guided} {Policy} {Learning} for {Adaptable} and {Transferrable} {Delicate} {Industrial} {Insertion}.
\newblock In \emph{Proceedings of the 2020 {Conference} on {Robot} {Learning}}, 2021.

\bibitem[O'~Kelly et~al.(2018)O'~Kelly, Sinha, Namkoong, Tedrake, and Duchi]{o_kelly_scalable_2018}
Matthew O'~Kelly, Aman Sinha, Hongseok Namkoong, Russ Tedrake, and John~C Duchi.
\newblock Scalable {End}-to-{End} {Autonomous} {Vehicle} {Testing} via {Rare}-event {Simulation}.
\newblock In \emph{Advances in {Neural} {Information} {Processing} {Systems}}, 2018.

\bibitem[Oquab et~al.(2023)Oquab, Darcet, Moutakanni, Vo, Szafraniec, Khalidov, Fernandez, Haziza, Massa, El-Nouby, Assran, Ballas, Galuba, Howes, Huang, Li, Misra, Rabbat, Sharma, Synnaeve, Xu, Jegou, Mairal, Labatut, Joulin, and Bojanowski]{oquab_dinov2_2023}
Maxime Oquab, Timothée Darcet, Théo Moutakanni, Huy~V. Vo, Marc Szafraniec, Vasil Khalidov, Pierre Fernandez, Daniel Haziza, Francisco Massa, Alaaeldin El-Nouby, Mido Assran, Nicolas Ballas, Wojciech Galuba, Russell Howes, Po-Yao Huang, Shang-Wen Li, Ishan Misra, Michael Rabbat, Vasu Sharma, Gabriel Synnaeve, Hu~Xu, Herve Jegou, Julien Mairal, Patrick Labatut, Armand Joulin, and Piotr Bojanowski.
\newblock {DINOv2}: {Learning} {Robust} {Visual} {Features} without {Supervision}.
\newblock \emph{Transactions on Machine Learning Research}, 2023.

\bibitem[Ota et~al.(2024)Ota, Jha, Jain, Yerazunis, Corcodel, Shukla, Bronars, and Romeres]{ota_autonomous_2024}
Kei Ota, Devesh~K. Jha, Siddarth Jain, Bill Yerazunis, Radu Corcodel, Yash Shukla, Antonia Bronars, and Diego Romeres.
\newblock Autonomous {Robotic} {Assembly}: {From} {Part} {Singulation} to {Precise} {Assembly}.
\newblock In \emph{2024 {IEEE}/{RSJ} {International} {Conference} on {Intelligent} {Robots} and {Systems} ({IROS})}, 2024.

\bibitem[Pan et~al.(2012)Pan, Chitta, and Manocha]{pan_fcl_2012}
Jia Pan, Sachin Chitta, and Dinesh Manocha.
\newblock {FCL}: {A} general purpose library for collision and proximity queries.
\newblock In \emph{2012 {IEEE} {International} {Conference} on {Robotics} and {Automation}}, 2012.

\bibitem[Pan et~al.(2023)Pan, Charron, Yang, Peters, Whelan, Kong, Parkhi, Newcombe, and Ren]{pan_aria_2023}
Xiaqing Pan, Nicholas Charron, Yongqian Yang, Scott Peters, Thomas Whelan, Chen Kong, Omkar Parkhi, Richard Newcombe, and Yuheng~(Carl) Ren.
\newblock Aria {Digital} {Twin}: {A} {New} {Benchmark} {Dataset} for {Egocentric} {3D} {Machine} {Perception}.
\newblock In \emph{Proceedings of the IEEE/CVF International Conference on Computer Vision (ICCV)}, 2023.

\bibitem[Park et~al.(2019)Park, Florence, Straub, Newcombe, and Lovegrove]{park_deepsdf_2019}
Jeong~Joon Park, Peter Florence, Julian Straub, Richard Newcombe, and Steven Lovegrove.
\newblock {DeepSDF}: {Learning} {Continuous} {Signed} {Distance} {Functions} for {Shape} {Representation}.
\newblock In \emph{Proceedings of the IEEE/CVF Conference on Computer Vision and Pattern Recognition (CVPR)}, 2019.

\bibitem[Patel et~al.(2024)Patel, Yin, Huang, Garg, Nayyeri, Fei-Fei, Lazebnik, and Li]{patel_real--sim--real_2024}
Shivansh Patel, Xinchen Yin, Wenlong Huang, Shubham Garg, Hooshang Nayyeri, Li~Fei-Fei, Svetlana Lazebnik, and Yunzhu Li.
\newblock A {Real}-to-{Sim}-to-{Real} {Approach} to {Robotic} {Manipulation} with {VLM}-{Generated} {Iterative} {Keypoint} {Rewards}.
\newblock In \emph{2025 IEEE International Conference on Robotics and Automation (ICRA)}, 2024.

\bibitem[Pavllo et~al.(2023)Pavllo, Tan, Rakotosaona, and Tombari]{pavllo_shape_2023}
Dario Pavllo, David~Joseph Tan, Marie-Julie Rakotosaona, and Federico Tombari.
\newblock Shape, {Pose}, and {Appearance} {From} a {Single} {Image} via {Bootstrapped} {Radiance} {Field} {Inversion}.
\newblock In \emph{Proceedings of the IEEE/CVF Conference on Computer Vision and Pattern Recognition (CVPR)}, 2023.

\bibitem[Pfaff et~al.(2025)Pfaff, Fu, Binagia, Isola, and Tedrake]{pfaff_scalable_2025}
Nicholas Pfaff, Evelyn Fu, Jeremy Binagia, Phillip Isola, and Russ Tedrake.
\newblock Scalable {Real2Sim}: {Physics}-{Aware} {Asset} {Generation} {Via} {Robotic} {Pick}-and-{Place} {Setups}, 2025.

\bibitem[Pollayil et~al.(2021)Pollayil, Grioli, Bonilla, and Bicchi]{pollayil_planning_2021}
George~Jose Pollayil, Giorgio Grioli, M.~Bonilla, and Antonio Bicchi.
\newblock Planning {Robotic} {Manipulation} with {Tight} {Environment} {Constraints}.
\newblock In \emph{2021 {IEEE}/{RSJ} {International} {Conference} on {Intelligent} {Robots} and {Systems} ({IROS})}, 2021.

\bibitem[Pontes et~al.(2019)Pontes, Kong, Sridharan, Lucey, Eriksson, and Fookes]{pontes_image2mesh_2019}
Jhony~K. Pontes, Chen Kong, Sridha Sridharan, Simon Lucey, Anders Eriksson, and Clinton Fookes.
\newblock {Image2Mesh}: {A} {Learning} {Framework} for {Single} {Image} {3D} {Reconstruction}.
\newblock In \emph{Computer {Vision} – {ACCV} 2018}, 2019.

\bibitem[Poole et~al.(2022)Poole, Jain, Barron, and Mildenhall]{poole_dreamfusion_2022}
Ben Poole, Ajay Jain, Jonathan~T. Barron, and Ben Mildenhall.
\newblock {DreamFusion}: {Text}-to-{3D} using {2D} {Diffusion}.
\newblock In \emph{The Eleventh International Conference on Learning Representations}, 2022.

\bibitem[Qi et~al.(2023)Qi, Yu, Dong, and Ma]{qi_vpp_2023}
Zekun Qi, Muzhou Yu, Runpei Dong, and Kaisheng Ma.
\newblock {VPP}: {Efficient} {Conditional} {3D} {Generation} via {Voxel}-{Point} {Progressive} {Representation}.
\newblock In \emph{Advances in {Neural} {Information} {Processing} {Systems}}, 2023.

\bibitem[Qian et~al.(2023)Qian, Mai, Hamdi, Ren, Siarohin, Li, Lee, Skorokhodov, Wonka, Tulyakov, and Ghanem]{qian_magic123_2023}
Guocheng Qian, Jinjie Mai, Abdullah Hamdi, Jian Ren, Aliaksandr Siarohin, Bing Li, Hsin-Ying Lee, Ivan Skorokhodov, Peter Wonka, Sergey Tulyakov, and Bernard Ghanem.
\newblock Magic123: {One} {Image} to {High}-{Quality} {3D} {Object} {Generation} {Using} {Both} {2D} and {3D} {Diffusion} {Priors}.
\newblock In \emph{International Conference on Learning Representations 2024}, 2023.

\bibitem[Qin and Badgwell(2003)]{qin_survey_2003}
S.~Joe Qin and Thomas~A. Badgwell.
\newblock A survey of industrial model predictive control technology.
\newblock \emph{Control Engineering Practice}, 2003.

\bibitem[Qureshi et~al.(2024)Qureshi, Garg, Yandun, Held, Kantor, and Silwal]{qureshi_splatsim_2024}
Mohammad~Nomaan Qureshi, Sparsh Garg, Francisco Yandun, David Held, George Kantor, and Abhisesh Silwal.
\newblock {SplatSim}: {Zero}-{Shot} {Sim2Real} {Transfer} of {RGB} {Manipulation} {Policies} {Using} {Gaussian} {Splatting}, 2024.

\bibitem[Radford et~al.(2021)Radford, Kim, Hallacy, Ramesh, Goh, Agarwal, Sastry, Askell, Mishkin, Clark, Krueger, and Sutskever]{radford_learning_2021}
Alec Radford, Jong~Wook Kim, Chris Hallacy, Aditya Ramesh, Gabriel Goh, Sandhini Agarwal, Girish Sastry, Amanda Askell, Pamela Mishkin, Jack Clark, Gretchen Krueger, and Ilya Sutskever.
\newblock Learning {Transferable} {Visual} {Models} {From} {Natural} {Language} {Supervision}.
\newblock In \emph{Proceedings of the 38th {International} {Conference} on {Machine} {Learning}}, 2021.

\bibitem[Rajeswar et~al.(2018)Rajeswar, Mannan, Golemo, Vazquez, Nowrouzezahrai, and Courville]{rajeswar_pix2scene_2018}
Sai Rajeswar, Fahim Mannan, Florian Golemo, David Vazquez, Derek Nowrouzezahrai, and Aaron Courville.
\newblock {Pix2Scene}: {Learning} {Implicit} {3D} {Representations} from {Images}.
\newblock 2018.

\bibitem[Rematas et~al.(2021)Rematas, Martin-Brualla, and Ferrari]{rematas_sharf_2021}
Konstantinos Rematas, Ricardo Martin-Brualla, and Vittorio Ferrari.
\newblock Sharf: {Shape}-conditioned {Radiance} {Fields} from a {Single} {View}.
\newblock In \emph{Proceedings of the 38th {International} {Conference} on {Machine} {Learning}}, 2021.

\bibitem[Remondino et~al.(2023)Remondino, Karami, Yan, Mazzacca, Rigon, and Qin]{remondino_critical_2023}
Fabio Remondino, Ali Karami, Ziyang Yan, Gabriele Mazzacca, Simone Rigon, and Rongjun Qin.
\newblock A {Critical} {Analysis} of {NeRF}-{Based} {3D} {Reconstruction}.
\newblock \emph{Remote Sensing}, 2023.

\bibitem[Ronneberger et~al.(2015)Ronneberger, Fischer, and Brox]{ronneberger_u-net_2015}
Olaf Ronneberger, Philipp Fischer, and Thomas Brox.
\newblock U-{Net}: {Convolutional} {Networks} for {Biomedical} {Image} {Segmentation}.
\newblock In \emph{Medical {Image} {Computing} and {Computer}-{Assisted} {Intervention} – {MICCAI} 2015}, 2015.

\bibitem[Schonberger and Frahm(2016)]{schonberger_structure--motion_2016}
Johannes~L. Schonberger and Jan-Michael Frahm.
\newblock Structure-{From}-{Motion} {Revisited}.
\newblock In \emph{Proceedings of the IEEE Conference on Computer Vision and Pattern Recognition (CVPR)}, 2016.

\bibitem[Shen et~al.(2024)Shen, Wu, Yi, Zhou, Zhang, Yan, and Wang]{shen_gamba_2024}
Qiuhong Shen, Zike Wu, Xuanyu Yi, Pan Zhou, Hanwang Zhang, Shuicheng Yan, and Xinchao Wang.
\newblock Gamba: {Marry} {Gaussian} {Splatting} with {Mamba} for single view {3D} reconstruction, 2024.

\bibitem[Shen et~al.(2021)Shen, Gao, Yin, Liu, and Fidler]{shen_deep_2021}
Tianchang Shen, Jun Gao, Kangxue Yin, Ming-Yu Liu, and Sanja Fidler.
\newblock Deep {Marching} {Tetrahedra}: a {Hybrid} {Representation} for {High}-{Resolution} {3D} {Shape} {Synthesis}.
\newblock In \emph{Advances in {Neural} {Information} {Processing} {Systems}}, 2021.

\bibitem[Shi et~al.(2023)Shi, Wang, Ye, Mai, Li, and Yang]{shi_mvdream_2023}
Yichun Shi, Peng Wang, Jianglong Ye, Long Mai, Kejie Li, and Xiao Yang.
\newblock {MVDream}: {Multi}-view {Diffusion} for {3D} {Generation}.
\newblock In \emph{International Conference on Learning Representations 2024}, 2023.

\bibitem[Siddiqui et~al.(2024)Siddiqui, Alliegro, Artemov, Tommasi, Sirigatti, Rosov, Dai, and Nießner]{siddiqui_meshgpt_2024}
Yawar Siddiqui, Antonio Alliegro, Alexey Artemov, Tatiana Tommasi, Daniele Sirigatti, Vladislav Rosov, Angela Dai, and Matthias Nießner.
\newblock {MeshGPT}: {Generating} {Triangle} {Meshes} with {Decoder}-{Only} {Transformers}.
\newblock In \emph{Proceedings of the IEEE/CVF Conference on Computer Vision and Pattern Recognition (CVPR)}, 2024.

\bibitem[Silberman et~al.(2012)Silberman, Hoiem, Kohli, and Fergus]{silberman_indoor_2012}
Nathan Silberman, Derek Hoiem, Pushmeet Kohli, and Rob Fergus.
\newblock Indoor {Segmentation} and {Support} {Inference} from {RGBD} {Images}.
\newblock In \emph{Computer {Vision} – {ECCV} 2012}, 2012.

\bibitem[Smith et~al.(2023)Smith, Kostrikov, and Levine]{smith_demonstrating_2023}
Laura Smith, Ilya Kostrikov, and Sergey Levine.
\newblock Demonstrating a walk in the park: {Learning} to walk in 20 minutes with model-free reinforcement learning.
\newblock \emph{Robotics: Science and Systems (RSS) Demo}, 2023.

\bibitem[Sun et~al.(2023)Sun, Zhang, Shao, Wang, Liu, Xie, and Liu]{sun_dreamcraft3d_2023}
Jingxiang Sun, Bo~Zhang, Ruizhi Shao, Lizhen Wang, Wen Liu, Zhenda Xie, and Yebin Liu.
\newblock {DreamCraft3D}: {Hierarchical} {3D} {Generation} with {Bootstrapped} {Diffusion} {Prior}.
\newblock In \emph{International Conference on Learning Representations 2024}, 2023.

\bibitem[Suomalainen et~al.(2022)Suomalainen, Karayiannidis, and Kyrki]{suomalainen_survey_2022}
Markku Suomalainen, Yiannis Karayiannidis, and Ville Kyrki.
\newblock A survey of robot manipulation in contact.
\newblock \emph{Robotics and Autonomous Systems}, 2022.

\bibitem[Swaminathan et~al.(2024)Swaminathan, Silver, and Akilan]{swaminathan_benchmarking_2024}
Tushar~Prasanna Swaminathan, Christopher Silver, and Thangarajah Akilan.
\newblock Benchmarking {Deep} {Learning} {Models} on {NVIDIA} {Jetson} {Nano} for {Real}-{Time} {Systems}: {An} {Empirical} {Investigation}, 2024.

\bibitem[Szymanowicz et~al.(2024{\natexlab{a}})Szymanowicz, Insafutdinov, Zheng, Campbell, Henriques, Rupprecht, and Vedaldi]{szymanowicz_flash3d_2024}
Stanislaw Szymanowicz, Eldar Insafutdinov, Chuanxia Zheng, Dylan Campbell, João~F. Henriques, Christian Rupprecht, and Andrea Vedaldi.
\newblock {Flash3D}: {Feed}-{Forward} {Generalisable} {3D} {Scene} {Reconstruction} from a {Single} {Image}, 2024{\natexlab{a}}.

\bibitem[Szymanowicz et~al.(2024{\natexlab{b}})Szymanowicz, Rupprecht, and Vedaldi]{szymanowicz_splatter_2024}
Stanislaw Szymanowicz, Chrisitian Rupprecht, and Andrea Vedaldi.
\newblock Splatter {Image}: {Ultra}-{Fast} {Single}-{View} {3D} {Reconstruction}.
\newblock In \emph{Proceedings of the IEEE/CVF Conference on Computer Vision and Pattern Recognition (CVPR)}, 2024{\natexlab{b}}.

\bibitem[Szymanowicz et~al.(2025)Szymanowicz, Zhang, Srinivasan, Gao, Brussee, Holynski, Martin-Brualla, Barron, and Henzler]{szymanowicz_bolt3d_2025}
Stanislaw Szymanowicz, Jason~Y. Zhang, Pratul Srinivasan, Ruiqi Gao, Arthur Brussee, Aleksander Holynski, Ricardo Martin-Brualla, Jonathan~T. Barron, and Philipp Henzler.
\newblock {Bolt3D}: {Generating} {3D} {Scenes} in {Seconds}, 2025.

\bibitem[Tadic et~al.(2022)Tadic, Toth, Vizvari, Klincsik, Sari, Sarcevic, Sarosi, and Biro]{tadic_perspectives_2022}
Vladimir Tadic, Attila Toth, Zoltan Vizvari, Mihaly Klincsik, Zoltan Sari, Peter Sarcevic, Jozsef Sarosi, and Istvan Biro.
\newblock Perspectives of {RealSense} and {ZED} {Depth} {Sensors} for {Robotic} {Vision} {Applications}.
\newblock \emph{Machines}, 2022.

\bibitem[Tang et~al.(2023)Tang, Ren, Zhou, Liu, and Zeng]{tang_dreamgaussian_2023}
Jiaxiang Tang, Jiawei Ren, Hang Zhou, Ziwei Liu, and Gang Zeng.
\newblock {DreamGaussian}: {Generative} {Gaussian} {Splatting} for {Efficient} {3D} {Content} {Creation}.
\newblock In \emph{International Conference on Learning Representations 2024}, 2023.

\bibitem[Tang et~al.(2025)Tang, Chen, Chen, Wang, Zeng, and Liu]{tang_lgm_2025}
Jiaxiang Tang, Zhaoxi Chen, Xiaokang Chen, Tengfei Wang, Gang Zeng, and Ziwei Liu.
\newblock {LGM}: {Large} {Multi}-view {Gaussian} {Model} for {High}-{Resolution} {3D} {Content} {Creation}.
\newblock In \emph{Computer {Vision} – {ECCV} 2024}, 2025.

\bibitem[Team et~al.(2025)Team, Chen, Chu, Gleize, Liang, Sax, Tang, Wang, Guo, Hardin, Li, Lin, Liu, Ma, Sagar, Song, Wang, Yang, Zhang, Dollár, Gkioxari, Feiszli, and Malik]{sam3dteam2025sam3d3dfyimages}
SAM~3D Team, Xingyu Chen, Fu-Jen Chu, Pierre Gleize, Kevin~J Liang, Alexander Sax, Hao Tang, Weiyao Wang, Michelle Guo, Thibaut Hardin, Xiang Li, Aohan Lin, Jiawei Liu, Ziqi Ma, Anushka Sagar, Bowen Song, Xiaodong Wang, Jianing Yang, Bowen Zhang, Piotr Dollár, Georgia Gkioxari, Matt Feiszli, and Jitendra Malik.
\newblock Sam 3d: 3dfy anything in images.
\newblock 2025.
\newblock URL \url{https://arxiv.org/abs/2511.16624}.

\bibitem[Tochilkin et~al.(2024)Tochilkin, Pankratz, Liu, Huang, Letts, Li, Liang, Laforte, Jampani, and Cao]{tochilkin_triposr_2024}
Dmitry Tochilkin, David Pankratz, Zexiang Liu, Zixuan Huang, Adam Letts, Yangguang Li, Ding Liang, Christian Laforte, Varun Jampani, and Yan-Pei Cao.
\newblock {TripoSR}: {Fast} {3D} {Object} {Reconstruction} from a {Single} {Image}, 2024.

\bibitem[Todorov et~al.(2012)Todorov, Erez, and Tassa]{todorov_mujoco_2012}
Emanuel Todorov, Tom Erez, and Yuval Tassa.
\newblock {MuJoCo}: {A} physics engine for model-based control.
\newblock In \emph{2012 {IEEE}/{RSJ} {International} {Conference} on {Intelligent} {Robots} and {Systems}}, 2012.

\bibitem[Torne et~al.(2024)Torne, Simeonov, Li, Chan, Chen, Gupta, and Agrawal]{torne_reconciling_2024}
Marcel Torne, Anthony Simeonov, Zechu Li, April Chan, Tao Chen, Abhishek Gupta, and Pulkit Agrawal.
\newblock Reconciling {Reality} through {Simulation}: {A} {Real}-to-{Sim}-to-{Real} {Approach} for {Robust} {Manipulation}, 2024.

\bibitem[Turkulainen et~al.(2025)Turkulainen, Ren, Melekhov, Seiskari, Rahtu, and Kannala]{turkulainen_dn-splatter_2025}
Matias Turkulainen, Xuqian Ren, Iaroslav Melekhov, Otto Seiskari, Esa Rahtu, and Juho Kannala.
\newblock {DN}-{Splatter}: {Depth} and {Normal} {Priors} for {Gaussian} {Splatting} and {Meshing}.
\newblock In \emph{2025 {IEEE}/{CVF} {Winter} {Conference} on {Applications} of {Computer} {Vision} ({WACV})}, 2025.

\bibitem[Vaswani et~al.(2017)Vaswani, Shazeer, Parmar, Uszkoreit, Jones, Gomez, Kaiser, and Polosukhin]{vaswani_attention_2017}
Ashish Vaswani, Noam Shazeer, Niki Parmar, Jakob Uszkoreit, Llion Jones, Aidan~N Gomez, Łukasz Kaiser, and Illia Polosukhin.
\newblock Attention is {All} you {Need}.
\newblock In \emph{Advances in {Neural} {Information} {Processing} {Systems}}, 2017.

\bibitem[Voleti et~al.(2025)Voleti, Yao, Boss, Letts, Pankratz, Tochilkin, Laforte, Rombach, and Jampani]{voleti_sv3d_2025}
Vikram Voleti, Chun-Han Yao, Mark Boss, Adam Letts, David Pankratz, Dmitry Tochilkin, Christian Laforte, Robin Rombach, and Varun Jampani.
\newblock {SV3D}: {Novel} {Multi}-view {Synthesis} and {3D} {Generation} from a {Single} {Image} {Using} {Latent} {Video} {Diffusion}.
\newblock In \emph{Computer {Vision} – {ECCV} 2024}, 2025.

\bibitem[Wang et~al.(2024)Wang, Reza, Vats, Ju, Thakurdesai, Wang, Crandall, Jung, and Seo]{wang_deep_2024}
Chuhua Wang, Md~Alimoor Reza, Vibhas Vats, Yingnan Ju, Nikhil Thakurdesai, Yuchen Wang, David~J. Crandall, Soon-heung Jung, and Jeongil Seo.
\newblock Deep learning-based {3D} reconstruction from multiple images: {A} survey.
\newblock \emph{Neurocomputing}, 2024.

\bibitem[Wang et~al.(2023{\natexlab{a}})Wang, Du, Li, Yeh, and Shakhnarovich]{wang_score_2023}
Haochen Wang, Xiaodan Du, Jiahao Li, Raymond~A. Yeh, and Greg Shakhnarovich.
\newblock Score {Jacobian} {Chaining}: {Lifting} {Pretrained} {2D} {Diffusion} {Models} for {3D} {Generation}.
\newblock In \emph{Proceedings of the IEEE/CVF Conference on Computer Vision and Pattern Recognition (CVPR)}, 2023{\natexlab{a}}.

\bibitem[Wang et~al.(2022)Wang, Wang, Long, Theobalt, Komura, Liu, and Wang]{wang_neuris_2022}
Jiepeng Wang, Peng Wang, Xiaoxiao Long, Christian Theobalt, Taku Komura, Lingjie Liu, and Wenping Wang.
\newblock {NeuRIS}: {Neural} {Reconstruction} of {Indoor} {Scenes} {Using} {Normal} {Priors}.
\newblock In \emph{Computer {Vision} – {ECCV} 2022}, 2022.

\bibitem[Wang et~al.(2018)Wang, Zhang, Li, Fu, Liu, and Jiang]{wang_pixel2mesh_2018}
Nanyang Wang, Yinda Zhang, Zhuwen Li, Yanwei Fu, Wei Liu, and Yu-Gang Jiang.
\newblock {Pixel2Mesh}: {Generating} {3D} {Mesh} {Models} from {Single} {RGB} {Images}.
\newblock In \emph{Proceedings of the European Conference on Computer Vision (ECCV)}, 2018.

\bibitem[Wang et~al.(2023{\natexlab{b}})Wang, Tan, Bi, Xu, Luan, Sunkavalli, Wang, Xu, and Zhang]{wang_pf-lrm_2023}
Peng Wang, Hao Tan, Sai Bi, Yinghao Xu, Fujun Luan, Kalyan Sunkavalli, Wenping Wang, Zexiang Xu, and Kai Zhang.
\newblock {PF}-{LRM}: {Pose}-{Free} {Large} {Reconstruction} {Model} for {Joint} {Pose} and {Shape} {Prediction}.
\newblock In \emph{International Conference on Learning Representations 2024}, 2023{\natexlab{b}}.

\bibitem[Wang et~al.(2019)Wang, Bao, Clavera, Hoang, Wen, Langlois, Zhang, Zhang, Abbeel, and Ba]{wang_benchmarking_2019}
Tingwu Wang, Xuchan Bao, Ignasi Clavera, Jerrick Hoang, Yeming Wen, Eric Langlois, Shunshi Zhang, Guodong Zhang, Pieter Abbeel, and Jimmy Ba.
\newblock Benchmarking {Model}-{Based} {Reinforcement} {Learning}, 2019.

\bibitem[Wang and Kasaei(2025)]{wang_learning_2025}
Yongliang Wang and Hamidreza Kasaei.
\newblock Learning {Dual}-{Arm} {Push} and {Grasp} {Synergy} in {Dense} {Clutter}.
\newblock \emph{IEEE Robotics and Automation Letters}, 2025.

\bibitem[Wei et~al.(2024)Wei, Geng, Chen, Deng, Wenbo, Zhao, Fang, Guibas, and Wang]{wei_d3roma_2024}
Songlin Wei, Haoran Geng, Jiayi Chen, Congyue Deng, Cui Wenbo, Chengyang Zhao, Xiaomeng Fang, Leonidas Guibas, and He~Wang.
\newblock D\${\textasciicircum}3\${RoMa}: {Disparity} {Diffusion}-based {Depth} {Sensing} for {Material}-{Agnostic} {Robotic} {Manipulation}.
\newblock In \emph{ECCV 2024 Workshop on Wild 3D: 3D Modeling, Reconstruction, and Generation in the Wild}, 2024.

\bibitem[Wei et~al.(2025)Wei, Zhang, Bi, Tan, Luan, Deschaintre, Sunkavalli, Su, and Xu]{wei_meshlrm_2025}
Xinyue Wei, Kai Zhang, Sai Bi, Hao Tan, Fujun Luan, Valentin Deschaintre, Kalyan Sunkavalli, Hao Su, and Zexiang Xu.
\newblock {MeshLRM}: {Large} {Reconstruction} {Model} for {High}-{Quality} {Meshes}, 2025.

\bibitem[Wen and Cho(2023)]{wen_object-aware_2023}
Mingyun Wen and Kyungeun Cho.
\newblock Object-{Aware} {3D} {Scene} {Reconstruction} from {Single} {2D} {Images} of {Indoor} {Scenes}.
\newblock \emph{Mathematics}, 2023.

\bibitem[Wu et~al.(2023{\natexlab{a}})Wu, Johnson, Malik, Feichtenhofer, and Gkioxari]{wu_multiview_2023}
Chao-Yuan Wu, Justin Johnson, Jitendra Malik, Christoph Feichtenhofer, and Georgia Gkioxari.
\newblock Multiview {Compressive} {Coding} for {3D} {Reconstruction}.
\newblock In \emph{Proceedings of the IEEE/CVF Conference on Computer Vision and Pattern Recognition (CVPR)}, 2023{\natexlab{a}}.

\bibitem[Wu et~al.(2024{\natexlab{a}})Wu, Karumuri, Zou, Bang, Li, Samaras, and Hadap]{wu_direct_2024}
Haoyu Wu, Meher~Gitika Karumuri, Chuhang Zou, Seungbae Bang, Yuelong Li, Dimitris Samaras, and Sunil Hadap.
\newblock Direct and {Explicit} {3D} {Generation} from a {Single} {Image}, 2024{\natexlab{a}}.

\bibitem[Wu et~al.(2024{\natexlab{b}})Wu, Liu, Cai, Yan, Wang, Hu, Duan, and Ma]{wu_unique3d_2024}
Kailu Wu, Fangfu Liu, Zhihan Cai, Runjie Yan, Hanyang Wang, Yating Hu, Yueqi Duan, and Kaisheng Ma.
\newblock {Unique3D}: {High}-{Quality} and {Efficient} {3D} {Mesh} {Generation} from a {Single} {Image}.
\newblock In \emph{Advances in Neural Information Processing Systems 37}, 2024{\natexlab{b}}.

\bibitem[Wu et~al.(2022)Wu, Liu, Chen, Li, Zheng, Cai, and Zheng]{wu_object-compositional_2022}
Qianyi Wu, Xian Liu, Yuedong Chen, Kejie Li, Chuanxia Zheng, Jianfei Cai, and Jianmin Zheng.
\newblock Object-{Compositional} {Neural} {Implicit} {Surfaces}.
\newblock In \emph{Computer {Vision} – {ECCV} 2022}, 2022.

\bibitem[Wu et~al.(2024{\natexlab{c}})Wu, Mildenhall, Henzler, Park, Gao, Watson, Srinivasan, Verbin, Barron, Poole, and Ho?y?ski]{wu_reconfusion_2024}
Rundi Wu, Ben Mildenhall, Philipp Henzler, Keunhong Park, Ruiqi Gao, Daniel Watson, Pratul~P. Srinivasan, Dor Verbin, Jonathan~T. Barron, Ben Poole, and Aleksander Ho?y?ski.
\newblock {ReconFusion}: {3D} {Reconstruction} with {Diffusion} {Priors}.
\newblock In \emph{Proceedings of the IEEE/CVF Conference on Computer Vision and Pattern Recognition (CVPR)}, 2024{\natexlab{c}}.

\bibitem[Wu et~al.(2025{\natexlab{a}})Wu, Zheng, Guan, Vedaldi, and Cham]{wu_amodal3r_2025}
Tianhao Wu, Chuanxia Zheng, Frank Guan, Andrea Vedaldi, and Tat-Jen Cham.
\newblock {Amodal3R}: {Amodal} {3D} {Reconstruction} from {Occluded} {2D} {Images}, 2025{\natexlab{a}}.

\bibitem[Wu et~al.(2025{\natexlab{b}})Wu, Zheng, Wu, and Cham]{wu_clusteringsdf_2025}
Tianhao Wu, Chuanxia Zheng, Qianyi Wu, and Tat-Jen Cham.
\newblock {ClusteringSDF}: {Self}-{Organized} {Neural} {Implicit} {Surfaces} for {3D} {Decomposition}.
\newblock In \emph{Computer {Vision} – {ECCV} 2024}, 2025{\natexlab{b}}.

\bibitem[Wu et~al.(2023{\natexlab{b}})Wu, Zhang, Fu, Wang, Ren, Pan, Wu, Yang, Wang, Qian, Lin, and Liu]{wu_omniobject3d_2023}
Tong Wu, Jiarui Zhang, Xiao Fu, Yuxin Wang, Jiawei Ren, Liang Pan, Wayne Wu, Lei Yang, Jiaqi Wang, Chen Qian, Dahua Lin, and Ziwei Liu.
\newblock {OmniObject3D}: {Large}-{Vocabulary} {3D} {Object} {Dataset} for {Realistic} {Perception}, {Reconstruction} and {Generation}.
\newblock In \emph{Proceedings of the IEEE/CVF Conference on Computer Vision and Pattern Recognition (CVPR)}, 2023{\natexlab{b}}.

\bibitem[Wu et~al.(2024{\natexlab{d}})Wu, Pan, Wu, Wang, Miao, and Wang]{wu_rl-gsbridge_2024}
Yuxuan Wu, Lei Pan, Wenhua Wu, Guangming Wang, Yanzi Miao, and Hesheng Wang.
\newblock {RL}-{GSBridge}: {3D} {Gaussian} {Splatting} {Based} {Real2Sim2Real} {Method} for {Robotic} {Manipulation} {Learning}, 2024{\natexlab{d}}.

\bibitem[Xiang et~al.(2025{\natexlab{a}})Xiang, Chen, Xu, Wang, Lv, Deng, Zhu, Dong, Zhao, Yuan, and Yang]{xiang2025trellis2}
Jianfeng Xiang, Xiaoxue Chen, Sicheng Xu, Ruicheng Wang, Zelong Lv, Yu~Deng, Hongyuan Zhu, Yue Dong, Hao Zhao, Nicholas~Jing Yuan, and Jiaolong Yang.
\newblock Native and compact structured latents for 3d generation.
\newblock \emph{Tech report}, 2025{\natexlab{a}}.

\bibitem[Xiang et~al.(2025{\natexlab{b}})Xiang, Lv, Xu, Deng, Wang, Zhang, Chen, Tong, and Yang]{xiang_structured_2025}
Jianfeng Xiang, Zelong Lv, Sicheng Xu, Yu~Deng, Ruicheng Wang, Bowen Zhang, Dong Chen, Xin Tong, and Jiaolong Yang.
\newblock Structured {3D} {Latents} for {Scalable} and {Versatile} {3D} {Generation}, 2025{\natexlab{b}}.

\bibitem[Xiang et~al.(2018)Xiang, Schmidt, Narayanan, and Fox]{xiang_posecnn_2018}
Yu~Xiang, Tanner Schmidt, Venkatraman Narayanan, and Dieter Fox.
\newblock {PoseCNN}: {A} {Convolutional} {Neural} {Network} for {6D} {Object} {Pose} {Estimation} in {Cluttered} {Scenes}.
\newblock In \emph{RSS 2018}, 2018.

\bibitem[Xie et~al.(2019)Xie, Yao, Sun, Zhou, and Zhang]{xie_pix2vox_2019}
Haozhe Xie, Hongxun Yao, Xiaoshuai Sun, Shangchen Zhou, and Shengping Zhang.
\newblock {Pix2Vox}: {Context}-{Aware} {3D} {Reconstruction} {From} {Single} and {Multi}-{View} {Images}.
\newblock In \emph{Proceedings of the IEEE/CVF International Conference on Computer Vision (ICCV)}, 2019.

\bibitem[Xie et~al.(2024)Xie, Zong, Qiu, Li, Feng, Yang, and Jiang]{xie_physgaussian_2024}
Tianyi Xie, Zeshun Zong, Yuxing Qiu, Xuan Li, Yutao Feng, Yin Yang, and Chenfanfu Jiang.
\newblock {PhysGaussian}: {Physics}-{Integrated} {3D} {Gaussians} for {Generative} {Dynamics}, 2024.

\bibitem[Xu et~al.(2024{\natexlab{a}})Xu, Yuan, Mardani, Liu, Song, Wang, and Vahdat]{xu_agg_2024}
Dejia Xu, Ye~Yuan, Morteza Mardani, Sifei Liu, Jiaming Song, Zhangyang Wang, and Arash Vahdat.
\newblock {AGG}: {Amortized} {Generative} {3D} {Gaussians} for {Single} {Image} to {3D}.
\newblock \emph{Transactions on Machine Learning Research}, 2024{\natexlab{a}}.

\bibitem[Xu et~al.(2024{\natexlab{b}})Xu, Cheng, Gao, Wang, Gao, and Shan]{xu_instantmesh_2024}
Jiale Xu, Weihao Cheng, Yiming Gao, Xintao Wang, Shenghua Gao, and Ying Shan.
\newblock {InstantMesh}: {Efficient} {3D} {Mesh} {Generation} from a {Single} {Image} with {Sparse}-view {Large} {Reconstruction} {Models}, 2024{\natexlab{b}}.

\bibitem[Xu et~al.(2023)Xu, Tan, Luan, Bi, Wang, Li, Shi, Sunkavalli, Wetzstein, Xu, and Zhang]{xu_dmv3d_2023}
Yinghao Xu, Hao Tan, Fujun Luan, Sai Bi, Peng Wang, Jiahao Li, Zifan Shi, Kalyan Sunkavalli, Gordon Wetzstein, Zexiang Xu, and Kai Zhang.
\newblock {DMV3D}: {Denoising} {Multi}-view {Diffusion} {Using} {3D} {Large} {Reconstruction} {Model}.
\newblock In \emph{International Conference on Learning Representations 2024}, 2023.

\bibitem[Xu et~al.(2025)Xu, Shi, Yifan, Chen, Yang, Peng, Shen, and Wetzstein]{xu_grm_2025}
Yinghao Xu, Zifan Shi, Wang Yifan, Hansheng Chen, Ceyuan Yang, Sida Peng, Yujun Shen, and Gordon Wetzstein.
\newblock {GRM}: {Large} {Gaussian} {Reconstruction} {Model} for {Efficient} {3D} {Reconstruction} and {Generation}.
\newblock In \emph{Computer {Vision} – {ECCV} 2024}, 2025.

\bibitem[Yan et~al.(2024{\natexlab{a}})Yan, Qu, Xu, Zhao, Wang, Wang, and Li]{yan_gs-slam_2024}
Chi Yan, Delin Qu, Dan Xu, Bin Zhao, Zhigang Wang, Dong Wang, and Xuelong Li.
\newblock {GS}-{SLAM}: {Dense} {Visual} {SLAM} with {3D} {Gaussian} {Splatting}.
\newblock In \emph{Proceedings of the IEEE/CVF Conference on Computer Vision and Pattern Recognition (CVPR)}, 2024{\natexlab{a}}.

\bibitem[Yan et~al.(2024{\natexlab{b}})Yan, Zhang, Li, Ma, and Ji]{yan_phycage_2024}
Han Yan, Mingrui Zhang, Yang Li, Chao Ma, and Pan Ji.
\newblock {PhyCAGE}: {Physically} {Plausible} {Compositional} {3D} {Asset} {Generation} from a {Single} {Image}, 2024{\natexlab{b}}.

\bibitem[Yao et~al.(2025)Yao, Zhang, Yan, Zeng, Zhang, Xu, Yang, Gu, and Yu]{yao_cast_2025}
Kaixin Yao, Longwen Zhang, Xinhao Yan, Yan Zeng, Qixuan Zhang, Lan Xu, Wei Yang, Jiayuan Gu, and Jingyi Yu.
\newblock {CAST}: {Component}-{Aligned} {3D} {Scene} {Reconstruction} from an {RGB} {Image}, 2025.

\bibitem[Yu et~al.(2022)Yu, Peng, Niemeyer, Sattler, and Geiger]{yu_monosdf_2022}
Zehao Yu, Songyou Peng, Michael Niemeyer, Torsten Sattler, and Andreas Geiger.
\newblock {MonoSDF}: {Exploring} {Monocular} {Geometric} {Cues} for {Neural} {Implicit} {Surface} {Reconstruction}.
\newblock In \emph{Advances in {Neural} {Information} {Processing} {Systems}}, 2022.

\bibitem[Yu et~al.(2024)Yu, Sattler, and Geiger]{yu_gaussian_2024}
Zehao Yu, Torsten Sattler, and Andreas Geiger.
\newblock Gaussian {Opacity} {Fields}: {Efficient} {Adaptive} {Surface} {Reconstruction} in {Unbounded} {Scenes}.
\newblock \emph{ACM Trans. Graph.}, 2024.

\bibitem[Yue et~al.(2024)Yue, Zhang, Chen, Chen, Zhou, and He]{yue_lidar-based_2024}
Xiangdi Yue, Yihuan Zhang, Jiawei Chen, Junxin Chen, Xuanyi Zhou, and Miaolei He.
\newblock {LiDAR}-based {SLAM} for robotic mapping: state of the art and new frontiers.
\newblock \emph{Industrial Robot: the international journal of robotics research and application}, 2024.

\bibitem[Yuksekgonul et~al.(2022)Yuksekgonul, Bianchi, Kalluri, Jurafsky, and Zou]{yuksekgonul_when_2022}
Mert Yuksekgonul, Federico Bianchi, Pratyusha Kalluri, Dan Jurafsky, and James Zou.
\newblock When and {Why} {Vision}-{Language} {Models} {Behave} like {Bags}-{Of}-{Words}, and {What} to {Do} {About} {It}?
\newblock In \emph{The Eleventh International Conference on Learning Representations}, 2022.

\bibitem[Zha et~al.(2020)Zha, Fu, Wang, Guo, Li, Wang, and Cai]{zha_semantic_2020}
Fusheng Zha, Yu~Fu, Pengfei Wang, Wei Guo, Mantian Li, Xin Wang, and Hegao Cai.
\newblock Semantic {3D} {Reconstruction} for {Robotic} {Manipulators} with an {Eye}-{In}-{Hand} {Vision} {System}.
\newblock \emph{Applied Sciences}, 2020.

\bibitem[Zhang et~al.(2021)Zhang, Cui, Zhang, Zeng, Pollefeys, and Liu]{zhang_holistic_2021}
Cheng Zhang, Zhaopeng Cui, Yinda Zhang, Bing Zeng, Marc Pollefeys, and Shuaicheng Liu.
\newblock Holistic {3D} {Scene} {Understanding} {From} a {Single} {Image} {With} {Implicit} {Representation}.
\newblock In \emph{Proceedings of the IEEE/CVF Conference on Computer Vision and Pattern Recognition (CVPR)}, 2021.

\bibitem[Zhang et~al.(2024)Zhang, Wang, Zhang, Qiu, Pang, Jiang, Yang, Xu, and Yu]{zhang_clay_2024}
Longwen Zhang, Ziyu Wang, Qixuan Zhang, Qiwei Qiu, Anqi Pang, Haoran Jiang, Wei Yang, Lan Xu, and Jingyi Yu.
\newblock {CLAY}: {A} {Controllable} {Large}-scale {Generative} {Model} for {Creating} {High}-quality {3D} {Assets}.
\newblock \emph{ACM Trans. Graph.}, 2024.

\bibitem[Zhang and Funkhouser(2018)]{zhang_deep_2018}
Yinda Zhang and Thomas Funkhouser.
\newblock Deep {Depth} {Completion} of a {Single} {RGB}-{D} {Image}.
\newblock In \emph{Proceedings of the IEEE Conference on Computer Vision and Pattern Recognition (CVPR)}, 2018.

\bibitem[Zhang et~al.(2020)Zhang, Chen, Ling, Gao, Zhang, Torralba, and Fidler]{zhang_image_2020}
Yuxuan Zhang, Wenzheng Chen, Huan Ling, Jun Gao, Yinan Zhang, Antonio Torralba, and Sanja Fidler.
\newblock Image {GANs} meet {Differentiable} {Rendering} for {Inverse} {Graphics} and {Interpretable} {3D} {Neural} {Rendering}.
\newblock In \emph{International Conference on Learning Representations}, 2020.

\bibitem[Zhao et~al.(2023)Zhao, Liu, Chen, Zeng, Wang, Cheng, Fu, Chen, Yu, and Gao]{zhao_michelangelo_2023}
Zibo Zhao, Wen Liu, Xin Chen, Xianfang Zeng, Rui Wang, Pei Cheng, Bin Fu, Tao Chen, Gang Yu, and Shenghua Gao.
\newblock Michelangelo: {Conditional} {3D} {Shape} {Generation} based on {Shape}-{Image}-{Text} {Aligned} {Latent} {Representation}.
\newblock In \emph{Advances in {Neural} {Information} {Processing} {Systems}}, 2023.

\bibitem[Zheng et~al.(2013)Zheng, Zhao, Yu, Ikeuchi, and Zhu]{zheng_beyond_2013}
Bo~Zheng, Yibiao Zhao, Joey~C. Yu, Katsushi Ikeuchi, and Song-Chun Zhu.
\newblock Beyond {Point} {Clouds}: {Scene} {Understanding} by {Reasoning} {Geometry} and {Physics}.
\newblock In \emph{2013 {IEEE} {Conference} on {Computer} {Vision} and {Pattern} {Recognition}}, 2013.

\bibitem[Zhu et~al.(2025)Zhu, Mou, Li, Ye, Huang, and Zhao]{zhu_vr-robo_2025}
Shaoting Zhu, Linzhan Mou, Derun Li, Baijun Ye, Runhan Huang, and Hang Zhao.
\newblock {VR}-{Robo}: {A} {Real}-to-{Sim}-to-{Real} {Framework} for {Visual} {Robot} {Navigation} and {Locomotion}, 2025.

\bibitem[Zhu et~al.(2024)Zhu, Wang, Kong, Kong, and Wang]{zhu_3d_2024}
Siting Zhu, Guangming Wang, Xin Kong, Dezhi Kong, and Hesheng Wang.
\newblock {3D} {Gaussian} {Splatting} in {Robotics}: {A} {Survey}, 2024.

\bibitem[Zou et~al.(2024)Zou, Yu, Guo, Li, Liang, Cao, and Zhang]{zou_triplane_2024}
Zi-Xin Zou, Zhipeng Yu, Yuan-Chen Guo, Yangguang Li, Ding Liang, Yan-Pei Cao, and Song-Hai Zhang.
\newblock Triplane {Meets} {Gaussian} {Splatting}: {Fast} and {Generalizable} {Single}-{View} {3D} {Reconstruction} with {Transformers}.
\newblock In \emph{Proceedings of the IEEE/CVF Conference on Computer Vision and Pattern Recognition (CVPR)}, 2024.

\bibitem[Šlapak et~al.(2024)Šlapak, Pardo, Dopiriak, Maksymyuk, and Gazda]{slapak_neural_2024}
Eugen Šlapak, Enric Pardo, Matúš Dopiriak, Taras Maksymyuk, and Juraj Gazda.
\newblock Neural radiance fields in the industrial and robotics domain: {Applications}, research opportunities and use cases.
\newblock \emph{Robotics and Computer-Integrated Manufacturing}, 2024.

\end{thebibliography}

\vfill\eject

\appendix

\section{Further Results}

\begin{figure*}[h!]
    \centering
    \begin{minipage}[h!]{0.49\textwidth}
        \resizebox{\textwidth}{!}{\input{plots/aria_chamfer.pgf}}
        \vspace{-10pt}
        \caption{Chamfer distances on Aria Digital Twin \cite{pan_aria_2023} dataset. Reconstruction errors on this dataset are noticeably higher than those shown in Fig. \ref{fig:ycb_chamfer}. Reconstructed surfaces tend to be 1cm distant from the closest target mesh surface.}
        \label{fig:aria_chamfer}
        \vspace{-10pt}
    \end{minipage}
    \hfill
    \begin{minipage}[h!]{0.49\textwidth}
        \vspace{-11pt}
        \resizebox{\textwidth}{!}{\input{plots/aria_stability.pgf}}
        \vspace{-10pt}
        \caption{Object stability on Aria Digital Twin \cite{pan_aria_2023} dataset. Most reconstructions lack stability within 5$^\circ$ of ground truth poses. In physics simulation, unstable geometries would cause scene reconstructions to collapse.}
        \label{fig:aria_stability}
        \vspace{-10pt}
    \end{minipage}
\end{figure*}

\begin{figure*}[h!]
    \centering
    \begin{minipage}[h!]{0.49\textwidth}
        \resizebox{\textwidth}{!}{\input{plots/ycb_chamfer_max.pgf}}
        \vspace{-20pt}
        \caption{Chamfer distances on YCB-Video \cite{xiang_posecnn_2018} aggregated by maximum distance per reconstructed object.}
        \label{fig:ycb_chamfer_max}
        \vspace{-10pt}
    \end{minipage}
    \hfill
    \begin{minipage}[h!]{0.49\textwidth}
        \resizebox{\textwidth}{!}{\input{plots/aria_chamfer_max.pgf}}
        \vspace{-20pt}
        \caption{Chamfer distances on the Aria Digital Twin \cite{pan_aria_2023} dataset aggregated by maximum distance per reconstructed object.}
        \label{fig:aria_chamfer_max}
        \vspace{-10pt}
    \end{minipage}
\end{figure*}

\begin{figure*}
    \centering
    \begin{minipage}[t]{0.49\textwidth}
        \resizebox{\textwidth}{!}{\input{plots/scene_ycb_chamfer.pgf}}
        \vspace{-20pt}
        \caption{Reconstruction error of scene reconstruction models on the YCB-Video \cite{xiang_posecnn_2018} dataset as measured by the Chamfer distance (CD). CD is averaged across 10,000 sampled surface points from the entire scene reconstruction and target. Indicators for min, max, and median are shown. Reconstructed surfaces tend to be over 1cm distant from the closest target surface.}
        \label{fig:scene_ycb_chamfer}
        \vspace{-10pt}
    \end{minipage}
    \hfill
    \begin{minipage}[t]{0.49\textwidth}
        \resizebox{\textwidth}{!}{\input{plots/scene_ycb_collision_pie.pgf}}
        \vspace{-20pt}
        \caption{Collision frequency in scene reconstruction on YCB-Video \cite{xiang_posecnn_2018}. Even though scene reconstruction models have access to information about the entire scene, mesh collisions in 3D reconstruction are common. PhysGen3D \cite{chen_physgen3d_2025} shows fewer collisions, potentially due to overestimating object distances (see Fig. \ref{fig:scene_object_distances}).}
        \label{fig:scene_collision_pie}
        \vspace{-10pt}
    \end{minipage}
\end{figure*}

\begin{figure*}
    \centering
    \begin{minipage}[t]{0.49\textwidth}
        \resizebox{\textwidth}{!}{\input{plots/scene_ycb_stability.pgf}}
        \vspace{-20pt}
        \caption{Total number of objects with physically stable poses within 5$^\circ$ from the ground truth scene pose versus total number of objects without such stable poses. Evaluated on YCB-Video using scene reconstruction models. Only a small fraction of the objects show physically stable poses near those object poses that have been estimated by the scene reconstruction models.}
        \label{fig:scene_ycb_stability}
        \vspace{-10pt}
    \end{minipage}
    \hfill
    \begin{minipage}[t]{0.49\textwidth}
        \resizebox{\textwidth}{!}{\input{plots/scene_ycb_stability_with_fit.pgf}}
        \vspace{-20pt}
        \caption{Object stability as in Fig. \ref{fig:scene_ycb_stability} but where each reconstructed object mesh is \textit{separately} aligned with the target mesh, instead of relying on the (relative) object poses from the scene reconstruction model. Individual alignment results in a greater number of stable poses, indicating that the scene reconstruction models produce suboptimal object poses. SAM3D displays especially strong stability results.}
        \label{fig:scene_ycb_stability_with_fit}
        \vspace{-10pt}
    \end{minipage}
\end{figure*}

\begin{figure*}
    \centering
    \begin{minipage}[t]{0.49\textwidth}
        \resizebox{\textwidth}{!}{\input{plots/scene_ycb_occlusion_chamfer.pgf}}
        \vspace{-10pt}
        \caption{Occlusion handling by scene-level models. Objects are individually aligned using the masked ICP procedure described in Section \ref{sec:methodology} to avoid pose estimation errors. Indicators for min, max, and median are shown. Gen3DSR \cite{dogaru_generalizable_2024} and PhysGen3D \cite{chen_physgen3d_2025} rely on image in-painting to resolve occlusions, MIDI \cite{huang_midi_2024} uses a custom shared denoising process to condition objects on one another, SAM3D \cite{sam3dteam2025sam3d3dfyimages} uses artificial occlusions during the training process. All approaches show noticeable improvements over single-object reconstruction models without any such procedures (see Fig. \ref{fig:ycb_occlusion_chamfer}). We conclude that scene-level information can be highly beneficial for resolving object occlusions.}
        \label{fig:scene_ycb_occlusion}
        \vspace{-10pt}
    \end{minipage}
    \hfill
    \begin{minipage}[t]{0.49\textwidth}
        \resizebox{\textwidth}{!}{\input{plots/scene_ycb_computation.pgf}}
        \vspace{-10pt}
        \caption{Reconstruction time in seconds and memory utilisation for reconstructing entire scenes with multiple objects using different scene reconstruction models. While VRAM utilisation is roughly in line with what is reported for single-object models in Fig. \ref{fig:compute}, the reconstruction time is an order of magnitude greater when operating on entire scenes. This effectively prohibits real-time execution when utilising such models.}
        \label{fig:scene_computation}
        \vspace{-10pt}
    \end{minipage}
\end{figure*}

\begin{figure*}
    \centering
    \begin{minipage}[t]{0.49\textwidth}
        \resizebox{\textwidth}{!}{\input{plots/ycb_chamfer_sampling.pgf}}
        \vspace{-10pt}
        \caption{The scatter plot compares Chamfer distances (CD) computed using area-weighted sampling and farthest-point sampling across all single-object reconstruction models on the YCB dataset. Each point represents an object reconstructed by one of the models, with coordinates corresponding to the Chamfer distance under the two sampling strategies. The strong correlation (near-unity slope) and tight clustering around the diagonal indicate that both methods yield similar results, suggesting that the choice of sampling strategy does not significantly affect the results.}
        \vspace{-10pt}
    \end{minipage}
    \hfill
    \begin{minipage}[t]{0.49\textwidth}
        \centering
        \resizebox{0.39\textwidth}{!}{\input{plots/ycb_stability_dso.pgf}}
        \vspace{2pt}
        \caption{DSO \cite{li_dso_2025} stability evaluation with a 20$^\circ$ tolerance to evaluate whether poor performance is a threshold artifact, given that DSO was specifically trained for object stability. This tolerance results in a majority of objects being stable. Note, this reflects threshold-sensitivity, as opposed to the 20$^\circ$ settling tolerance proposed in \cite{li_dso_2025}. Our criterion measures agreement with the observed scene pose, and a 20$^\circ$ pose difference is too large a discrepancy for simulators to predict contact behaviour with neighbouring objects.}
        \vspace{-10pt}
    \end{minipage}
\end{figure*}

\clearpage
\newpage

\begin{figure*}[hp!]
\centering
\resizebox{0.9\textwidth}{!}{\graphicspath{{plots/mesh_renders_joint}} \input{plots/reconstruction_examples_joint.pgf}}
\caption{
Sample reconstructions on the YCB-Video \cite{xiang_posecnn_2018} dataset (rows 1, 2, 3) and the Aria Digital Twin \cite{pan_aria_2023} dataset (rows 4, 5, 6) using different 3D reconstruction methods. Reconstruction target objects are contoured and coloured. The scene renderings are solely for context as the reconstruction models are only given the segmented target object on transparent background. We observe that models tend to perform better on simple object shapes from highly informative viewpoints. Non-standard objects and more extreme viewpoints result in markedly lower reconstruction accuracy (compare rows 1, 2). Reconstruction quality on the Aria dataset is arguably lower than on YCB-Video. Objects in the Aria dataset tend to occupy less space on the image plane, resulting in less visual information in the model inputs.}
\label{fig:joint_reconstruction_examples}
\end{figure*}

\begin{figure*}[hp!]
\centering
\resizebox{0.8\textwidth}{!}{\graphicspath{{plots/mesh_renders_scene_ycb_sample/}} \input{plots/scene_reconstruction_examples_ycb_sample.pgf}}
\caption{YCB-Video \cite{xiang_posecnn_2018} scene reconstructions using different 3D reconstruction methods that process multi-object scenes. The objects that are reconstructed are shown contoured and in colour. The scene renderings are solely for context as the reconstruction models are only given the segmented target objects on transparent background. Scene reconstruction models struggle not only with object reconstruction quality but also accurate object pose estimation, resulting in mesh collisions and levitating objects.}
\label{fig:scene_reconstruction_examples}
\end{figure*}

\clearpage
\newpage

\section{Additional Desiderata Discussion}
\label{app:desiderata_details}

\begin{table}[t]
\centering
\caption{Overview of desiderata alongside and based on tolerances reported in the literature, where available.}
\small
\begin{tabular}{lll}
\hline
 & \textbf{Reported tolerances} & \textbf{Our operating point} \\ \hline
D1 Accuracy & sub-mm (assembly) to 1\,mm (grasping) & 5\,mm, deliberately more permissive \\
D2 Collision & binary; no tolerance defined & any detected collision (strictest) \\
D3 Stability & not established for scene-pose agreement & 5$^\circ$ \\
D4 Occlusion & attainable level, measured & +10\% CD over visible regions \\
D5 Latency & deployment constraint, not quality & $<$2s per scene \\ \hline
\end{tabular}
\label{tab:desiderata_tab}
\end{table}

\textbf{Domain Gap in Reconstruction Inputs.}
Real-world robots are often equipped with RGB-D cameras which offer both RGB and depth perception at relatively low cost \cite{tadic_perspectives_2022}. Yet these cameras typically do not match the resolution and colour profile of images commonly found in 3D reconstruction training datasets \cite{wu_omniobject3d_2023, deitke_objaverse-xl_2023}. The difference is especially noticeable when an object occupies only a small part of the image plane, as is common when observing an entire scene. Consequently, 3D reconstruction training data distributions \cite{tochilkin_triposr_2024, liu_syncdreamer_2023, liu_zero-1--3_2023} differ substantially from inputs encountered during real-world deployment, potentially resulting in less accurate reconstructions and artifacts \cite{barcellona_dream_2024}. While some approaches like \cite{jiang_real3d_2024} co-train on both real-world and synthetic data to improve accuracy, it remains an open question whether this can yield satisfactory results for the precision demands of robotics applications. For the bootstrapping scenario, looser tolerances may suffice for the initial reconstruction, with accuracy improving through active refinement.

Prior work often uses high-fidelity synthetic or smartphone-captured imagery for training and evaluation, which can mask performance degradation under typical robotic sensing conditions. In contrast, our benchmark employs camera data that is closer to robotic deployment standards, to reflect realistic input quality. Ablation studies reconstructing YCB objects from higher-quality iPhone images show marked improvements for some tested models, confirming sensitivity to sensor characteristics. This divergence underscores that the observed limitations are not inherent to the models alone, but arise from the mismatch between training data assumptions and real deployment conditions. Thus, our evaluation captures a critical challenge for practical uptake in robotics and highlights the need for training strategies robust to sensor domain shifts.

\textbf{Occlusion completion approaches.}
Some 3D reconstruction models run a separate occlusion completion model in their pipeline, which can lead to computational overhead and accumulation of reconstruction error \cite{agarwal_scenecomplete_2024}. Other works assume that objects tend to be symmetrical or follow other manual shape priors \cite{kong_vmap_2023, li_rico_2023}. Object stability, as described in Desideratum 3, can give valuable clues about the structure of occluded parts. Our threshold of no more than 10\% increase in Chamfer distance at occluded regions serves as a practical benchmark for household objects and tabletop scenarios. In settings with simpler geometry or more forgiving manipulation strategies, larger degradation may be acceptable. Tasks requiring precise contact at occluded surfaces may demand tighter tolerances.

\textbf{On-device computational constraints.}
Robots generally have limited on-board compute due to power and payload restrictions \cite{swaminathan_benchmarking_2024}. Many 3D reconstruction models utilise computationally expensive operations such as diffusion \cite{xu_instantmesh_2024, liu_syncdreamer_2023} or iterative optimisation \cite{liu_zero-1--3_2023, wang_score_2023, qian_magic123_2023}, rendering them unable to run on-device without significantly increased latency. Running reconstruction models on dedicated servers and streaming results to the robotic platform is often not feasible due to latency, network dependence, and potential security or privacy concerns \cite{abdulsalam_security_2022}. While one could argue that pre-computing reconstructions and retrieving them from a database might address computational concerns, this is generally impractical in online environments like households where object positions, orientations, and even the objects themselves change frequently.
 
\textbf{Accuracy/efficiency trade-off.}
Another consideration is a possible accuracy/efficiency trade-off, as some computationally more expensive models produce lower-error reconstructions. This would indicate that the practitioner has to make a judgement call as to what reconstruction error and computational cost can be afforded. As we show in Section \ref{sec:experiments}, however, we do not observe a clear trend suggesting that computationally more expensive models automatically produce more accurate reconstructions.

\clearpage
\newpage

\section{3D Reconstruction $-$ A Brief Taxonomy}
\label{app:taxonomy}
\noindent In this section, we provide a focused literature overview of 3D reconstruction approaches to motivate our approach and position our findings. Given the rapidly evolving nature of this research area with new methods appearing frequently, we cannot comprehensively cover all existing approaches. Instead, we focus on representative methods that illustrate key capabilities and limitations relevant to robotic manipulation requirements. This framework highlights how current approaches align with or fall short of the practical needs of robotic systems operating in real-world environments. Specifically, we look at different 3D representations that are used for reconstruction (Section \ref{app:representation_types}), various input data requirements (Section \ref{app:input_data}), different prediction paradigms (Section \ref{app:optimisation_types}), different representation abstraction levels (Section \ref{app:representation_abstractions}), the consideration of physical properties and parameters during reconstruction (Section \ref{app:physics}), and training data properties (Section \ref{app:training_data}). Tables \ref{tab:reconstruction_overview_single_1} and \ref{tab:reconstruction_overview_single_2} give an overview of a selection of existing reconstruction models and their properties.

\subsection{Types of Representations}
\label{app:representation_types}
\noindent \textbf{Explicit representations} store each data point individually. \textbf{Point clouds}, for example, define surfaces with sets of 3D coordinates in Euclidean space. Due to the proliferation of RGB-D cameras \cite{tadic_perspectives_2022}, point clouds are often readily available in robotics domains and provide comparably more accurate information about scene geometry than 2D RGB projections. 

For a detailed overview of 3D reconstruction with point clouds, see \cite{huang_surface_2024}. \textbf{Meshes} are among the most widely used 3D representations for simulation as they allow efficient rendering and accurate rigid body collision computation. As a result, simulators are primarily built around mesh representations \cite{coumans_pybullet_2016,makoviychuk_isaac_2021,mittal_orbit_2023}. Meshes consist of sets of vertices and faces, surfaces that span subsets of vertices and define the object surface. Due to this very specific structure, directly predicting meshes is non-trivial and existing methods mostly resort to predicting deformation fields on mesh initialisations (e.g. \cite{pontes_image2mesh_2019,jack_learning_2019,wang_pixel2mesh_2018}), or predicting other representations that are subsequently converted into meshes through techniques like ray marching or marching cubes (e.g. \cite{liu_zero-1--3_2023,tochilkin_triposr_2024,boss_sf3d_2024,liu_one-2-3-45_2023}). Exceptions include \cite{siddiqui_meshgpt_2024}, who define a sequential order on the mesh vertices and faces to predict mesh structures with a transformer architecture \cite{vaswani_attention_2017}. Mesh reconstruction from visual observations has received significant attention from the computer vision community \cite{wang_pixel2mesh_2018,groueix_papier-mache_2018,mezghanni_physically-aware_2021,liu_zero-1--3_2023,qian_magic123_2023,chen_fantasia3d_2023,liu_one-2-3-45_2023,boss_sf3d_2024,tang_lgm_2025,xu_instantmesh_2024} and is the main focus of our evaluations in this work due to the widespread support for meshes in robotic simulators. 
\textbf{3DGS} \cite{kerbl_3d_2023} is a relatively recent addition to explicit representations. Each Gaussian is defined by its centre position in Euclidean space, a covariance matrix defining orientation and shape of the Gaussian blob, an opacity value, and its colour. Together, a set of Gaussian blobs defines object surfaces or volumes. 3DGS representations are very efficient to render but come with memory overhead as each Gaussian needs to be stored individually. Due to its high visual fidelity, 3DGS has recently become a focus for novel view synthesis \cite{kerbl_3d_2023,dalal_gaussian_2024} and 3D reconstruction \cite{melas-kyriazi_im-3d_2024,szymanowicz_splatter_2024,yan_phycage_2024,xu_grm_2025,xu_agg_2024,szymanowicz_flash3d_2024,liu_gaussian_2024,huang_2d_2024,yu_gaussian_2024}.

\textbf{Implicit representations} instead define a parametric function where 3D coordinates are mapped to various quantities of interest, such as occupancy, colour, or signed distance. The number of free parameters can be set strictly smaller than the number of Euclidean coordinates, resulting in a (potentially lossy) compression of the geometric and visual information. At the same time, however, rendering is very inefficient as the function has to be queried many times to accumulate geometry and visual information along the camera rays. Specific models include DeepSDF \cite{park_deepsdf_2019} and Occupancy Networks \cite{mescheder_occupancy_2019}.

\subsection{Input Data}
\label{app:input_data}
\noindent We focus on reconstruction from image or depth data in robotics contexts. Such 3D reconstruction approaches can be categorised based on the density of input viewpoints required: dense-view, sparse-view, and single-view methods, each with distinct characteristics and applications in robotics scenarios. 

\textbf{Dense-view} 3D reconstruction, that is reconstructing only what is observed, has a rich history. If camera poses, projection parameters, and depth data are available, it is straightforward to convert depth information to dense point clouds in the world coordinate system. Without camera extrinsics, reconstruction becomes an optimisation problem. One major line of work, Simultaneous Localisation and Mapping (SLAM), concerns reconstruction of a map of the environment and estimation of the camera trajectory in it from real-time sensor data \cite{aulinas_slam_2008}. Another line of work, Structure-from-Motion, does not operate on online sequential data but instead reconstructs 3D structures and estimates camera viewpoints from unordered sets of offline images \cite{schonberger_structure--motion_2016}. 3DGS \cite{kerbl_3d_2023}, a more recent form of dense-view reconstruction, typically relies on unordered sets of dense views with camera extrinsics available; extensions to simultaneous localisation and mapping have also been proposed \cite{yan_gs-slam_2024,keetha_splatam_2024,matsuki_gaussian_2024}. 

\textbf{Sparse-view} reconstruction alleviates the need for complete scene coverage in the reconstruction inputs by leveraging explicit priors \cite{zheng_beyond_2013,li_rico_2023}, exploiting artifacts in scene representations \cite{kong_vmap_2023}, or relying on learnt parametric models \cite{yu_monosdf_2022,choy_3d-r2n2_2016,kar_learning_2017,xie_pix2vox_2019} to complete unobserved regions. \cite{zheng_beyond_2013} utilise explicit priors for volumetric completion while \cite{li_rico_2023} makes assumptions on object smoothness and object-background relations. \cite{kong_vmap_2023} exploits artifacts resulting from implicit neural representations: as neural networks tend towards smooth function approximation, they observe that implicit neural representations naturally perform object closure and completion to some degree. \cite{yu_monosdf_2022} use parametric models to predict depth and surface normal maps for input images and report increased detail in reconstruction when using these cues. Other works learn parametric models that individually embed multiple views in latent space, fuse these embeddings, and decode towards complete 3D reconstructions \cite{choy_3d-r2n2_2016,kar_learning_2017,xie_pix2vox_2019}. \cite{liu_gaussian_2024} cluster 3D Gaussians into individual object and complete occluded surfaces using a variational autoencoder \cite{kingma_auto-encoding_2014}. \cite{charatan_pixelsplat_2024} learn a feed-forward model to reconstruct radiance fields from two views and \cite{chen_single-stage_2023} learn a density function over NeRFs that allows image-conditioned sampling with arbitrary numbers of views.

\textbf{Single-view} reconstruction is arguably more difficult for producing faithful reconstructions as substantial portions of a scene are unobserved. Due to this ill-position, proposed methods are trained on datasets to capture general geometric properties of the objects in our world. Single-view reconstruction is particularly central to our work, as it represents the most practical and generalisable scenario for robotic deployment. Unlike multi-view methods that require substantial scene coverage or specialised equipment, single-view approaches align with real-world constraints where robots must rapidly assess and interact with novel environments from limited viewpoints. While these methods traditionally face greater challenges in producing accurate reconstructions, their successful development would enable transformative capabilities for autonomous systems operating in unstructured environments. Broadly speaking, methods can be categorised into two distinct flavours. Firstly, and historically more prominently, there are methods that only rely on the single input view during 3D reconstruction by either directly regressing 3D shapes (for example \cite{fan_point_2017,szymanowicz_splatter_2024,chen_transformers_2022,li_instant3d_2023}) or via test-time optimisation (for example \cite{pavllo_shape_2023,lee_understanding_2022,jain_zero-shot_2022}).

The second popular flavour of single-view 3D reconstruction is leveraging multi-view generative models in the reconstruction process. One way of doing so is explicitly generating additional views based on the input image using multi-view diffusion models, predicting 3D Gaussians for each individual view using a U-Net \cite{tang_lgm_2025} or transformer \cite{xu_grm_2025}, and finally fusing the predicted Gaussians. \cite{melas-kyriazi_im-3d_2024} instead use generated views as rendering targets in the original 3DGS formulation \cite{kerbl_3d_2023}. \cite{xu_instantmesh_2024} choose to predict triplane features from multiple artificial views and extract meshes while \cite{liu_one-2-3-45_2023} use a dense set of artificially generated views to predict signed distance values. Multi-view consistency is generally an issue in these approaches as any inconsistencies directly translate to artifacts in the reconstruction. \cite{liu_one-2-3-45_2023} address this by using intermediate cost-volumes that capture inconsistencies. \cite{liu_one-2-3-45_2024} instead fine-tune a multi-view generative model to be more consistent across views, and \cite{liu_syncdreamer_2023} and \cite{shi_mvdream_2023} propose to condition on a joint feature space derived from multiple views. Some approaches utilise additional information such as albedo or surface normal maps to increase reconstruction fidelity \cite{chen_2l3_2024,wu_unique3d_2024}. A separate line of work does not use explicit multi-view generation but instead uses multi-view diffusion models to guide optimisation of an underlying 3D representation \cite{liu_zero-1--3_2023,qian_magic123_2023}. This is explained in more detail in Appendix \ref{app:optimisation_types}.

\subsection{Optimisation Paradigms}
\label{app:optimisation_types}
\noindent In this section, we highlight two primary approaches 3D reconstruction leverage to arrive at their outputs: feed-forward prediction and inference-time optimisation. 

\textbf{Feed-forward prediction} is most in line with the standard use-case of deep learning architectures. For example, \cite{wang_pixel2mesh_2018} propose a series of convolutional layers to gradually deform ellipsoid mesh initialisations, \cite{fan_point_2017} propose an encoder-decoder structure to predict point cloud coordinates.
For a detailed overview of such reconstruction methods up to the year 2021, please refer to \cite{han_image-based_2021}. More recently, other models use feed-forward prediction to regress 3D Gaussians: \cite{szymanowicz_splatter_2024} use a U-Net \cite{ronneberger_u-net_2015} to predict one Gaussian blob per input image pixel while allowing arbitrary position offsets for 3D Gaussians. The simplicity of this architecture enables very fast prediction at up to 38 Hz. \cite{xu_agg_2024} instead use transformers to predict 3D Gaussians based on DINOv2 \cite{oquab_dinov2_2023} image features and \cite{szymanowicz_flash3d_2024} rely on a ResNet-type \cite{he_deep_2016} backbone. Implicit representations can also be produced in a feed-forward manner: \cite{chen_transformers_2022} learn to regress the parameters of NeRFs using an image-conditioned transformer network while \cite{jun_shap-e_2023} train a diffusion model on NeRF MLP weight matrices. \cite{hong_lrm_2023} propose a model that maps from image input to triplane features which, when interpolated, can be used to query radiance information at spatial locations. \cite{tochilkin_triposr_2024} and \cite{boss_sf3d_2024} propose several improvements including data curation, masked losses for higher reconstruction fidelity, material and illumination estimation, and efficient mesh extraction.

A major limitation of feed-forward shape prediction is that 3D datasets have historically been orders of magnitude smaller than their 2D counterparts. This has often limited feed-forward reconstructions to few specific object categories \cite{choy_3d-r2n2_2016,wang_pixel2mesh_2018,rematas_sharf_2021}, since the limited data volume precludes effective training for category-agnostic reconstruction. In response to this, several works proposed leveraging pre-trained 2D vision models in a GAN-style \cite{goodfellow_generative_2014} setup $-$ training a 3D generative model with guidance through 2D renderings \cite{nguyen-phuoc_hologan_2019,zhang_image_2020,rajeswar_pix2scene_2018}. Simultaneously, significant efforts have been made towards scaling up 3D datasets \cite{dai_scannet_2017,deitke_objaverse-xl_2023} which has lead to impressive feed-forward reconstruction performances \cite{huang_zeroshape_2024,xu_dmv3d_2023,li_instant3d_2023,wang_pf-lrm_2023,zou_triplane_2024}.

\textbf{Inference-time optimisation} of an underlying 3D representation with guidance signals from pre-trained 2D vision models is an alternative to feed-forward prediction. Some works propose using CLIP \cite{radford_learning_2021} to produce gradients on 2D renderings of the underlying 3D asset \cite{hong_avatarclip_2022,khalid_clip-mesh_2022,lee_understanding_2022,jain_zero-shot_2022}. If the rendering process itself is differentiable, the chain-rule can be used to obtain gradient signals w.r.t. the 3D representation. However, the bag-of-words behaviour of CLIP models \cite{yuksekgonul_when_2022} usually limits the visual fidelity. More recently, powerful 2D diffusion models have replaced CLIP in guiding shape optimisation. \cite{poole_dreamfusion_2022} and \cite{wang_score_2023} both propose a process that repeatedly adds noise to 2D renders of an underlying 3D asset, then uses a pre-trained 2D diffusion model to obtain the 2D score (the gradient of the log-density over the noised 2D image space w.r.t. the 2D image), and apply the chain rule from 2D image to 3D asset to update the asset. An issue with these early diffusion approaches is that the diffusion score cannot be sufficiently guided for multi-view usage $-$ a requirement for making meaningful updates to all parts of the 3D asset. In response, \cite{liu_zero-1--3_2023} train a view-conditioned 2D diffusion model and increase the 3D reconstruction quality significantly. \cite{qian_magic123_2023} propose several extensions, including using both 2D and 3D diffusion models and a multi-stage setup. Generally, runtime and memory consumption is a major concern with these approaches, with object generations often requiring multiple hours of optimisation. 

\subsection{Scene-level vs. Object-level Reconstruction}
\label{app:representation_abstractions}
\noindent 3D reconstruction approaches can be broadly categorised based on their representational abstraction level. \textbf{Wholistic approaches}, including many SLAM systems, typically focus on generating unified representations of entire scenes (e.g. \cite{wang_neuris_2022,yu_monosdf_2022,szymanowicz_flash3d_2024,turkulainen_dn-splatter_2025,szymanowicz_bolt3d_2025}). These methods generally rely on depth information or estimation to reconstruct observed elements, resulting in representations that essentially project 2.5D information into 3D space. 

In contrast, \textbf{object-centric approaches} focus on reconstructing individual objects as distinct entities \cite{wang_pixel2mesh_2018,melas-kyriazi_im-3d_2024,xu_instantmesh_2024,tang_lgm_2025,li_rico_2023}. These methods prioritise resolving self- and viewpoint-occlusion through heuristics or parametric models trained to capture typical object features from datasets.

Between these two extremes lies an emerging middle ground: generating digital twins of entire scenes by fully reconstructing constituent objects while maintaining their individuality \cite{yao_cast_2025,dogaru_generalizable_2024,huang_midi_2024,han_reparo_2024}. This hybrid approach is particularly valuable for robotics applications as it enables both scene-level understanding and object-level interaction. Additionally, it allows reconstruction algorithms to leverage contextual information about an object's surroundings, providing valuable cues about potential occlusions and spatial relationships. For example, proximity between objects suggests possible unseen object parts due to occlusion, rather than assuming these parts simply do not exist because they are not visible in the observation.

\subsection{Physics}
\label{app:physics}
\noindent Physical plausibility is essential for physics simulation. For rigid bodies, this entails aspects of the overall scene composition, such as object collision and stability, as well as physical properties of individual objects themselves. 

On the scene level, many approaches formulate loss terms that are incorporated into gradient-based optimisation. For instance, SDF-based representations allow querying whether a particular point in space belongs to the volume of more than one object. As the SDF representation itself is typically optimised with gradient descent, adding a penalty term for doubly occupied space into the overall optimisation is straightforward \cite{wu_object-compositional_2022,zhang_holistic_2021,gao_graphdreamer_2024}. \cite{li_rico_2023} and \cite{chatterjee_3d-scene-former_2024} also regularise collision between objects and scene boundaries and background walls by relying on SDFs and bounding boxes, respectively. \cite{wu_clusteringsdf_2025} and \cite{hassena_objectcarver_2024} take a slightly different approach by not discouraging existing collisions between objects, but instead clustering and decomposing a scene representation into individual objects which naturally addresses object overlap. \cite{yao_cast_2025} propose a complex setup that first queries a vision-language model to produce a scene graph containing object-object relations, such as one object supporting another or objects generally being in close contact with one another. Based on these pairwise graph relations, the authors select loss terms from a set of handcrafted loss functions that are then used to optimise physical scene coherence via gradient descent. \cite{huang_midi_2024} do not formulate explicit loss terms but instead resolve collisions by coupling the denoising process of multiple objects in their proposed 3D diffusion model, leading to fewer violations.

Instead of handcrafted loss-terms, some approaches resort to differentiable physics simulation. This allows penalising anything that would result in undesirable behaviour during physics simulation, including object collision and object stability $-$ even involving multiple objects at once. \cite{yan_phycage_2024} dissect single objects into several components, such as a frog wearing a scarf, and optimise component shapes to resolve collision and object stability. \cite{ni_phyrecon_2024} perform a similar process on the scene level by reconstructing individual objects and adapting their position and shape such that the resulting scene is physically stable.

Some works do not consider entire scenes or compositional objects but instead concentrate on individual objects only. \cite{mezghanni_physically-aware_2021} learn a stability predictor that is used for guiding object reconstruction. \cite{guo_physically_2024} formulate differentiable loss terms to optimise stability of single objects. \cite{chen_atlas3d_2024} use differentiable simulation during text-to-3D generation for generating stable objects, and \cite{mezghanni_physical_2022} augment an SDF decoder with a differentiable simulator for object stability.

Beyond stability and collision, researchers have also looked at estimating object articulations \cite{chen_urdformer_2024,le_articulate-anything_2024,li_dragapart_2025} and material properties \cite{boss_sf3d_2024,feng_pie-nerf_2024,guo_physically_2024}. While certainly useful for digital twinning of more complex environments, these aspects go beyond the scope of our study of whether reconstructing physically coherent scenes in simulation for robotics is feasible with current category-agnostic 3D reconstruction methods.

\subsection{Training Data}
\label{app:training_data}
\noindent 3D datasets have historically been smaller than text or image datasets and most existing large-scale 3D datasets are primarily object-centric rather than scene-centric. This focus on isolated objects fails to capture important contextual relationships between objects in realistic environments. For robotics applications seeking to create accurate digital twins of entire scenes, having training data that represents realistic object arrangements, support relationships, and partial occlusions is crucial. Scene-level data would allow reconstruction models to learn typical object configurations and physical constraints that govern real-world environments. Without such contextual learning, models struggle to produce physically plausible reconstructions of complex scenes, leading to the stability and collision issues we observe in our evaluations.

\noindent\textbf{Object Datasets.}
ShapeNet \cite{chang_shapenet_2015} has been very influential ever since its introduction, offering around 51k 3D object models in its curated form. Still, it does not offer enough breadth and detail to allow for training category-agnostic reconstruction models and as a result, most proposed reconstruction methods focused on few specific categories, such as cars or chairs \cite{choy_3d-r2n2_2016,wang_pixel2mesh_2018,rematas_sharf_2021}. More recently, OmniObjects3D \cite{wu_omniobject3d_2023} and GSO \cite{downs_google_2022} proposed highly-detailed 3D object datasets, albeit at a smaller scale than ShapeNet. In 2023, Objaverse-XL \cite{deitke_objaverse-xl_2023} presented an inflection point for 3D datasets, offering over 10 million highly-diverse 3D models. For the first time, 3D reconstruction models trained on this dataset made the value proposition of being able to reconstruct 'any' object \cite{liu_zero-1--3_2023}. A recurring theme among large-scale 3D datasets is the reliance on synthetic object rendering. With robotics being an area that is heavily concerned with the Sim2Real gap, including subtle differences between renderings and real-world camera footage \cite{jaunet_sim2realviz_2021}, relying solely on simulated training data might not translate to good downstream performance. 

\noindent\textbf{Scene Datasets.}
Regarding scene-level datasets, the focus is not on object diversity but rather on diversity of scene layout and composition. Similar to object-centric datasets, scene-level datasets can be synthetic and potentially procedurally generated \cite{khanna_habitat_2024,kolve_ai2-thor_2022,deitke_procthor_2022}. Procedural generation allows theoretically infinite scene layouts from a finite set of objects. Other datasets focus on manually obtained 3D scans of real scenes \cite{chang_matterport3d_2017,baruch_arkitscenes_2021}. Due to the scanning process that is used for obtaining these datasets, one could say they are not truly 3D but only a collection of 2.5D information. As a result, the collected scenes frequently contain holes and do not feature complete object meshes. Instead, they only reconstruct what has been scanned. \cite{xiang_posecnn_2018} provide a collection of 3D scenes featuring YCB objects \cite{calli_ycb_2015}, a collection of common household objects that are frequently featured in robot experiments \cite{deng_self-supervised_2020,lu_systematic_2021,lerher_robotic_2023}. Recently, \cite{pan_aria_2023} proposed a richly annotated digital twin dataset comprised of two indoor scenes.

\clearpage
\newpage

\begin{sidewaystable*}[hp!]
    \tiny
    \centering
    \caption{Relevant single-view 3D reconstruction models and their properties.
    }
    \label{tab:reconstruction_overview_single_1}
    \begin{tabular}{@{}p{0.4cm} p{1.2cm} p{0.8cm} p{1.2cm} p{2.0cm} p{3.6cm} p{1.3cm} p{3.5cm} p{0.6cm}@{}}
    \toprule
    \textbf{Paper} & \textbf{Input Viewpoints} & \textbf{Depth Input} & \textbf{Reconstruction Level} & \textbf{Asset Type} & \textbf{Optimisation Type} & \textbf{Object Decomposition} & \textbf{Physics} & \textbf{Open Source} \\ 
    \midrule

\multicolumn{1}{l|}{\cite{chatterjee_3d-scene-former_2024}} & Single & Pretrained & Scene & Mesh & Feed-Forward & Yes & Object Collision & No  \\ 
\multicolumn{1}{l|}{\cite{xu_agg_2024}} & Single & No & Object & 3D Gaussians & Feed-Forward & No & No & No  \\ 
\multicolumn{1}{l|}{\cite{wu_amodal3r_2025}} & Single & No & Object, Scene & 3D Gaussians, Radiance Field, Mesh & Feed-Forward & Segments single foreground object & No & No  \\ 
\multicolumn{1}{l|}{\cite{szymanowicz_bolt3d_2025}} & Single, Multi & No & Scene & 3D Gaussians & Feed-Forward & No & No & No  \\ 
\multicolumn{1}{l|}{\cite{chu_buol_2023}} & Single & Pretrained & Scene & Voxel & Feed-Forward & Yes & Objects cannot collide as scene is represented by voxels & Yes  \\ 
\multicolumn{1}{l|}{\cite{yao_cast_2025}} & Single & Pretrained & Scene & Mesh & Iterative Feed-Forward (Object Reconstruction), GPT-4v (Scene Graph), Gradient-Descent (Physics Optimisation) & Yes & Collision and Stability & No  \\ 
\multicolumn{1}{l|}{\cite{zhang_clay_2024}} & Single & Yes & Object & Mesh & Feed-Forward & No & No & No  \\ 
\multicolumn{1}{l|}{\cite{wu_direct_2024}} & Single & Custom & Object & 3D Gaussians, Mesh & Feed-Forward & No & No & No  \\ 
\multicolumn{1}{l|}{\cite{xu_dmv3d_2023}} & Single & No & Object & NeRF, Mesh & Feed-Forward & No & No & No  \\ 
\multicolumn{1}{l|}{\cite{sun_dreamcraft3d_2023}} & Single & Pretrained & Object & Implicit Surface, Mesh & Gradient-Descent (Score Distillation) & No & No & Yes  \\ 
\multicolumn{1}{l|}{\cite{tang_dreamgaussian_2023}} & Single & No & Object & 3D Gaussians, Mesh & Gradient-Descent (Score Distillation \& Texture Refinement) & No & No & Yes  \\ 
\multicolumn{1}{l|}{\cite{li_dso_2025}} & Single & No & Object & Mesh & Feed-Forward & No & Fine-tuned to predict stable object & Yes  \\ 
\multicolumn{1}{l|}{\cite{szymanowicz_flash3d_2024}} & Single & Pretrained & Scene & 3D Gaussians & Feed-Forward & No & No & Yes  \\ 
\multicolumn{1}{l|}{\cite{shen_gamba_2024}} & Single & No & Object & 3D Gaussians & Feed-Forward & No & No & Yes  \\ 
\multicolumn{1}{l|}{\cite{dogaru_generalizable_2024}} & Single & Pretrained & Scene & Mesh & Gradient-Descent (Score Distillation) by default; can be combined with other models & Yes & No & Yes  \\ 
\multicolumn{1}{l|}{\cite{xu_grm_2025}} & Single & No & Object & 3D Gaussians, Mesh & Feed-Forward & No & No & No  \\ 
\multicolumn{1}{l|}{\cite{zhang_holistic_2021}} & Single & No & Scene & SDF, Mesh & Feed-Forward, Iterative Refinement & Yes & Collision & Yes  \\ 
\multicolumn{1}{l|}{\cite{melas-kyriazi_im-3d_2024}} & Single & No & Object & 3D Gaussians & Feed-Forward (Video Gen), Gradient-Descent (3D Gaussians) & No & No & No  \\ 
\multicolumn{1}{l|}{\cite{li_instant3d_2023}} & Text, Single, Multi & No & Object & Triplane, NeRF, Mesh & Feed-Forward & No & No & No  \\ 
\multicolumn{1}{l|}{\cite{xu_instantmesh_2024}} & Single & No & Object & Mesh & Feed-Forward & No & No & Yes  \\ 
\multicolumn{1}{l|}{\cite{lu_large_2024}} & Single & No & Object & 3D Gaussians & Feed-Forward & No & No & No  \\ 
\multicolumn{1}{l|}{\cite{tang_lgm_2025}} & Single & No & Object & 3D Gaussians, Mesh & Feed-Forward & No & No & Yes  \\ 
\multicolumn{1}{l|}{\cite{hong_lrm_2023}} & Single & No & Object & NeRF, Mesh & Feed-Forward & No & No & No  \\ 
\multicolumn{1}{l|}{\cite{qian_magic123_2023}} & Single & No & Object & Mesh & Gradient-Descent (Score Distillation) & No & No & Yes  \\ 
\multicolumn{1}{l|}{\cite{wei_meshlrm_2025}} & Single & No & Object & Mesh & Feed-Forward & No & No & No  \\ 
\multicolumn{1}{l|}{\cite{zhao_michelangelo_2023}} & Single & No & Object & Mesh & Feed-Forward & No & No & Yes  \\ 
\multicolumn{1}{l|}{\cite{liu_one-2-3-45_2023}} & Single & No & Object & SDF, Mesh & Feed-Forward & No & No & Yes  \\ 
\multicolumn{1}{l|}{\cite{liu_one-2-3-45_2024}} & Single & No & Object & SDF, Mesh & Feed-Forward & No & No & No  \\ 
\multicolumn{1}{l|}{\cite{yan_phycage_2024}} & Single & No & Object & 3D Gaussians & Feed-Forward (Multi-View Gen), Gradient-Descent (SDF, Score Distillation, Diff'ble Physics Simulation) & Single object into 2 components & Full Simulation & Yes  \\ 
\multicolumn{1}{l|}{\cite{chen_physgen3d_2025}} & Single & Pretrained & Scene & Mesh & Feed-Forward (Reconstruction), GPT-4o, Gradient-Descent (Diff'ble Renderer) & Yes & Reconstructions can be used in particle simulator but stability and collision play no role during reconstruction process. & Yes  \\ 
\multicolumn{1}{l|}{\cite{guo_physically_2024}} & Single & No & Object & Mesh & Gradient-Descent & No & Stability, Mechanical Material Properties & Yes  \\ 
    \bottomrule
    \end{tabular}
\end{sidewaystable*}

\begin{sidewaystable*}[hp!]
    \tiny
    \centering
    \caption{(Continued.) Relevant single-view 3D reconstruction models and their properties.
    }
    \label{tab:reconstruction_overview_single_2}
    \begin{tabular}{@{}p{0.4cm} p{1.2cm} p{0.8cm} p{1.2cm} p{2.0cm} p{3.6cm} p{1.3cm} p{0.6cm} p{0.6cm}@{}}
    \toprule
    \textbf{Paper} & \textbf{Input Viewpoints} & \textbf{Depth Input} & \textbf{Reconstruction Level} & \textbf{Asset Type} & \textbf{Optimisation Type} & \textbf{Object Decomposition} & \textbf{Physics} & \textbf{Open Source} \\ 
    \midrule

\multicolumn{1}{l|}{\cite{xiang2025trellis2}} & Single & No & Object & Mesh & Feed-Forward & No & No & Yes  \\ 
\multicolumn{1}{l|}{\cite{sam3dteam2025sam3d3dfyimages}} & Single & Yes & Object, Scene & 3D Gaussians, Mesh & Feed-Forward & No & No & Yes  \\ 
\multicolumn{1}{l|}{\cite{huang_midi_2024}} & Single & No & Scene & SDF, Mesh & Feed-Forward & Yes & Soft Collision & Yes  \\ 
\multicolumn{1}{l|}{\cite{wu_multiview_2023}} & Single & Yes & Object, Scene & Point Cloud & Feed-Forward & No & No & Yes  \\ 
\multicolumn{1}{l|}{\cite{shi_mvdream_2023}} & Single & No & Object & NeRF & Gradient-Descent (Score Distillation) & No & No & Yes  \\ 
\multicolumn{1}{l|}{\cite{wen_object-aware_2023}} & Single & Pretrained & Scene & Mesh & Feed-Forward & Yes & Collision & No  \\ 
\multicolumn{1}{l|}{\cite{mezghanni_physical_2022}} & Single & No & Object & SDF & Feed-Forward (SDF), Gradient-Descent (Diff'ble Physics Sim) & No & Stability & No  \\ 
\multicolumn{1}{l|}{\cite{jiang_real3d_2024}} & Single & No & Object & Mesh & Feed-Forward & No & No & Yes  \\ 
\multicolumn{1}{l|}{\cite{han_reparo_2024}} & Single & Pretrained & Scene & Mesh & Gradient-Descent (Score Distillation \& Diff'ble Rendering for Poses) & Yes & No & Not functional  \\
\multicolumn{1}{l|}{\cite{boss_sf3d_2024}} & Single & No & Object & Mesh & Feed-Forward & No & No & Yes  \\ 
\multicolumn{1}{l|}{\cite{jun_shap-e_2023}} & Single & No & Object & NeRF, Mesh & Feed-Forward & No & No & Yes  \\ 
\multicolumn{1}{l|}{\cite{chen_single-view_2024}} & Single & No & Scene & Mesh & Feed-Forward & Yes & No & Yes  \\ 
\multicolumn{1}{l|}{\cite{szymanowicz_splatter_2024}} & Single & No & Object & 3D Gaussians & Feed-Forward & No & No & Yes  \\ 
\multicolumn{1}{l|}{\cite{voleti_sv3d_2025}} & Single & No & Object & Mesh & Feed-Forward (Multi-View Gen), Gradient-Descent (Score Distillation) & No & No & Yes  \\ 
\multicolumn{1}{l|}{\cite{liu_syncdreamer_2023}} & Single & No & Object & Mesh & Gradient-Descent (Score Distillation) & No & No & Yes  \\ 
\multicolumn{1}{l|}{\cite{chen_transformers_2022}} & Single & No & Object & NeRF & Feed-Forward & No & No & Yes  \\ 
\multicolumn{1}{l|}{\cite{zou_triplane_2024}} & Single & No & Object & 3D Gaussians & Feed-Forward & No & No & Yes  \\ 
\multicolumn{1}{l|}{\cite{wu_unique3d_2024}} & Single & Custom & Object & Mesh & Feed-Forward & No & No & Yes  \\ 
\multicolumn{1}{l|}{\cite{qi_vpp_2023}} & Single & No & Object & Point Cloud & Feed-Forward & No & No & Yes  \\ 
\multicolumn{1}{l|}{\cite{long_wonder3d_2024}} & Single & No & Object & Mesh & Feed-Forward (Multi-View Gen), Gradient-Descent (SDF) & No & No & Yes  \\ 
\multicolumn{1}{l|}{\cite{liu_zero-1--3_2023}} & Single & No & Object & Mesh & Gradient-Descent (Score Distillation) & No & No & Yes  \\ 
\multicolumn{1}{l|}{\cite{huang_zeroshape_2024}} & Single & Custom & Object & Mesh & Feed-Forward & No & No & Yes  \\ 
 
    \bottomrule
    \end{tabular}
\end{sidewaystable*}

\clearpage
\newpage

\section{Model Descriptions}
\label{app:model_descriptions}
\noindent\textbf{Object-Level Reconstruction Models.}
\textbf{SF3D} \cite{boss_sf3d_2024} extends TripoSR \cite{tochilkin_triposr_2024} by adding material and lighting estimation. Uses a transformer to predict a Triplane \cite{chan_efficient_2022}, which is meshed via differentiable Marching Tetrahedron \cite{shen_deep_2021}.
\textbf{InstantMesh} \cite{xu_instantmesh_2024} uses a multi-view diffusion model to generate views, which are encoded into Triplane \cite{chan_efficient_2022} features. These features are then converted into meshes.
\textbf{One2345} \cite{liu_one-2-3-45_2023} produces dense multi-view images via diffusion, then aggregates them into a discretised volume. This is converted into a signed distance function and finally meshed.
\textbf{LGM} \cite{tang_lgm_2025} generates multi-view images with diffusion and processes them via U-Net \cite{ronneberger_u-net_2015} into 3D Gaussians. These are rendered to train a NeRF \cite{mildenhall_nerf_2022}, which is used for mesh extraction.
\textbf{Michelangelo} \cite{zhao_michelangelo_2023} combines image, shape, and text in a latent autoencoder, followed by diffusion-based shape sampling. A conditioned occupancy network then extracts meshes.
\textbf{ZeroShape} \cite{huang_zeroshape_2024} estimates depth and camera intrinsics to project the image into 3D space. A conditional occupancy network processes this for mesh generation.
\textbf{Real3D} \cite{jiang_real3d_2024} builds on TripoSR \cite{tochilkin_triposr_2024} and incorporates real-world data using an unsupervised loss. This co-training reduces domain shift and improves performance on real data.
\textbf{DSO} \cite{li_dso_2025} fine-tunes TRELLIS \cite{xiang_structured_2025} with a physics-informed reward system for stable 3D shapes. TRELLIS encodes images into structured latents, decoded into SDFs for mesh extraction.
\textbf{TRELLIS.2} \cite{xiang2025trellis2} proposes a custom voxelisation that allows efficient conversion between mesh- and voxel-representation. This voxel representation is compressed into a latent space by a VAE, on which a large generative model operates.

\noindent\textbf{Scene-Level Reconstruction Models.}
\textbf{MIDI} \cite{huang_midi_2024} segments all objects in the scene using an image segmentation model. Then denoises latent embeddings jointly across objects using cross-instance self-attention, improving spatial reasoning. This enhances the model’s understanding of object relationships in the scene.
\textbf{Gen3DSR} \cite{dogaru_generalizable_2024} segments and inpaints occluded objects before applying single-object 3D reconstruction. Aligns meshes to the scene using RANSAC \cite{fischler_and_random_1981} between object meshes and estimated depth. Background reconstruction is omitted for fair comparison in experiments.
\textbf{PhysGen3D} \cite{chen_physgen3d_2025} segments, inpaints, and reconstructs objects with InstantMesh \cite{xu_instantmesh_2024}. Aligns them using keypoint descriptors and a differentiable renderer for refined pose. Adds language-model-inferred physical properties but does not handle collisions or stability during reconstruction.
\textbf{SAM3D} \cite{sam3dteam2025sam3d3dfyimages} uses a mixture of transformers to predict poses and voxel representations of objects in the image. The voxel representation is subsequently refined with a flow transformer on whose output prediction heads produce either mesh or 3D Gaussian reconstructions.

\noindent\textbf{Justification for Model Exclusions.}
Since we are interested in estimating the complete scene geometry from visual observation, we do not consider methods that generate 3D structures only from text input (e.g. \cite{chen_fantasia3d_2023,poole_dreamfusion_2022,chen_atlas3d_2024}). Furthermore, given our focus on single-view reconstruction, we do not evaluate models that require multiple viewpoints (e.g. \cite{chen_2l3_2024,wang_pf-lrm_2023,xu_grm_2025,yu_monosdf_2022,wu_object-compositional_2022,li_rico_2023,kong_vmap_2023}) or even dense observations (including \cite{chabal_online_2025,ni_phyrecon_2024,turkulainen_dn-splatter_2025,jiang_vr-gs_2024,patel_real--sim--real_2024,jia_discoverse_2024,barcellona_dream_2024,abou-chakra_physically_2024,han_re3sim_2025,torne_reconciling_2024,wu_rl-gsbridge_2024,lou_robo-gs_2024,li_robogsim_2024,pfaff_scalable_2025,qureshi_splatsim_2024,zhu_vr-robo_2025,jiang_vr-gs_2024}). Most established simulation engines in robotics require mesh representations of objects. At the same time, converting 3D Gaussians to mesh surfaces is not a trivial operation \cite{huang_2d_2024}. To fairly assess reconstruction performance, we opt to only consider methods that either produce meshes themselves or come with pre-defined mesh extraction routines. Further, optimisation-based routines such as score distillation are slow by design. Since we are concerned with computational requirements and responsiveness during reconstruction (Section \ref{sec:desiderata}), we do not consider such methods in our experiments (including \cite{liu_zero-1--3_2023,liu_syncdreamer_2023,qian_magic123_2023,sun_dreamcraft3d_2023,tang_dreamgaussian_2023,chen_atlas3d_2024,shi_mvdream_2023,voleti_sv3d_2025,kim_complete_2025,wu_reconfusion_2024}). Finally, while we generally outline potential issues with offloading compute during deployment from the robot platform to cloud services in Appendix \ref{app:desiderata_details}, our empirical evaluation is practically constrained to methods with publicly available open-source implementations.

\section{Limitations of the Evaluation}
\noindent While we have shown that the reconstructions produced by the evaluated models are too inaccurate for transferring grasps from simulation to the real world, the effect of reconstruction errors on the performance of robot manipulation tasks in general and on physical simulation in particular is highly context-dependent. The task, target object, its surroundings, the type of robot gripper, and even the simulation engine all affect manipulation and simulation outcomes. As such, the question of precisely what error magnitude is permissible in a certain situation and whether any systematic relationship can be drawn up is an entire research question of itself and left for future work.

Our per-object analysis of the relationship between Chamfer distance and grasp transfer success (Fig. \ref{fig:grasp_vs_chamfer}) provides a first quantitative characterisation of how reconstruction quality translates to downstream task performance, but it also reveals the limitations of relying on global geometric metrics alone. The moderate correlation ($r=-0.55$) and the diminishing-returns regime below 1cm suggest that additional metrics may be needed to fully predict manipulation outcomes. Candidate metrics that our evaluation does not currently include are local surface normal accuracy at contact regions, mesh watertightness (required for correct collision volume computation in many physics engines), centre-of-mass estimation error (which directly affects stability and dynamics prediction), and the accuracy of fine geometry features such as edges and rims that are often critical for grasping thin or complex objects. Developing task-grounded metrics that capture these local properties $-$ for instance, by evaluating reconstruction fidelity specifically at regions identified as likely grasp or contact points $-$ is an important direction for future work that would complement the global metrics presented here. Furthermore, the relationship between reconstruction quality and task performance is itself task-dependent: the error tolerance for a robust parallel-jaw grasp of a cylinder differs substantially from that of a precision pinch grasp on a pen, and systematically characterising these task-specific tolerances remains an open research question.

\section{Discussion Details}
\label{app:discussion_details}
\noindent\textbf{Depth integration.}
We suggest that future models should incorporate depth information more explicitly in the reconstruction pipeline. Depth information is heavily commoditised in robotics due to the prevalence of RGB-D cameras \cite{tadic_perspectives_2022} and consequently widely used in applications \cite{wei_d3roma_2024, li_visual_2024,kent_leveraging_2020}. Depth can resolve location/scale ambiguities and provide near-ground-truth supervision about object shapes. However, as most low-cost RGB-D cameras rely on structured light projection, depth is often incomplete or noisy at locations occluded from the infrared sensor, at highly reflective surfaces, low-texture regions, or those affected by external light sources. With depth completion being non-trivial \cite{zhang_deep_2018, khan_comprehensive_2022}, reconstruction models should be able to cope with incomplete depth, low resolution, and depth misalignment.
 
Among the evaluated models, only SAM3D allows utilising 2.5D information from RGB-D cameras by design, though some others could be adapted. For example, One2345 \cite{liu_one-2-3-45_2023} could initialise its dense 3D volume with depth information; Michelangelo \cite{zhao_michelangelo_2023} could initialise its latent space with observed point clouds before running reverse diffusion. Gen3DSR \cite{dogaru_generalizable_2024} appears most readily adaptable since it relies on a separate pre-trained model to predict depth maps for mesh alignment. However, during exploratory experiments where we replaced the depth prediction model with the observed RGB-D depth map, we found that Gen3DSR is unable to handle missing depth values $-$ a common issue with RGB-D cameras. Depth information offers a way of making reconstruction less ambiguous at almost no additional cost, and its utilisation is a promising avenue for future work.

While the few-view setting presents a promising compromise between reconstruction and practicality, our evaluation focuses strictly on single-view methods, many of which cannot natively process additional views. Extending them to multi-view inputs would require architectural changes that go beyond the scope of a fair comparison. We acknowledge this gap and discuss the trade-offs in Appendix \ref{app:input_data}.
 
\noindent\textbf{Scene-level training data limitations.}
A reason that scene-level reconstruction approaches struggle is that some \cite{dogaru_generalizable_2024,yao_cast_2025,chen_physgen3d_2025,sam3dteam2025sam3d3dfyimages} rely on single-object reconstruction models under the hood. MIDI \cite{huang_midi_2024} is an exception but does not show reduced object collisions either, potentially because its training data only contains larger objects such as room-level furniture, and applying models trained on these datasets to tabletop manipulation tasks is likely to result in poor performance. This highlights a common issue: scene-level datasets either are real-world datasets without true 3D annotations (only 2.5D) \cite{silberman_indoor_2012,dai_scannet_2017,chang_matterport3d_2017} or synthetic datasets with a large domain gap \cite{chang_shapenet_2015,deitke_objaverse-xl_2023}. The ideal training dataset for robotics-oriented reconstruction would include both synthetic data for scale and diversity, and carefully curated real-world scene data with accurate ground-truth meshes.
 
\noindent\textbf{Object stability training.}
While DSO \cite{li_dso_2025} specifically trains for object stability, the authors set a \textit{settling} threshold of 20$^\circ$, i.e. a pose is considered stable if the equilibrium pose the object finds after being initialised at that pose is within 20$^\circ$ of said pose. This is different from our metric, as we restrict maximum pose difference between the observed scene pose and any found stable object pose. Our evaluation shows that even the best-performing model, Michelangelo \cite{zhao_michelangelo_2023}, only produces such stable poses for about 65\% of objects on YCB-Video \cite{xiang_posecnn_2018} and 40\% on Aria Digital Twin \cite{pan_aria_2023}. This is not sufficient for physical simulation where object stability is crucial for scene integrity. Future models should train for scene-pose agreement at practically relevant levels, not only self-support.

\end{document}